\documentclass[aos]{arxiv-imsart} 

\RequirePackage{amsthm,amsmath,amsfonts,amssymb,color,xcolor}
\RequirePackage[numbers]{natbib}
\RequirePackage{graphicx}

\usepackage[colorlinks=True,citecolor=blue,urlcolor=blue]{hyperref}
\usepackage[postexp]{optional} 
\startlocaldefs

\usepackage{upgreek} 
\usepackage{xspace} 
\usepackage{enumitem} 
\usepackage{etoc} 
\usepackage{algorithm}
\usepackage[noend]{algpseudocode}
\usepackage[linesnumbered,ruled,algo2e]{algorithm2e}%

\usepackage{multirow}

\usepackage{booktabs}
\usepackage{dsfont}
\usepackage{amsmath}
\usepackage{thmtools}
\usepackage{nicefrac}
\usepackage{subfig}
\usepackage{graphicx}
\usepackage{tikz} 
\usetikzlibrary{positioning,fit,shapes.geometric,backgrounds}

\usepackage{thmtools,thm-restate}
\usepackage[capitalize]{cleveref}  

\usepackage{autonum}

\theoremstyle{plain}

\newtheorem{theorem}{Theorem}
\newtheorem{lemma}{Lemma}
\newtheorem{definition}{Definition}
\newtheorem{example}{Example}
\newtheorem{remark}{Remark}

\newtheorem{assumption}{Assumption}
\newtheorem{corollary}{Corollary}
\newtheorem{proposition}{Proposition}

\newtheorem{myexample}{Example}

\newtheorem{myproposition}{Proposition}

\newtheorem{mycorollary}{Corollary}

\newtheorem{mylemma}{Lemma}

\newtheorem{mydefinition}{Definition}

\newtheorem{myremark}{Remark}

\newtheorem{myassumption}{Assumption}

\crefname{appendix}{App.}{Apps.}
\crefname{subsubsubappendix}{App.}{Apps.}
\crefname{equation}{}{}
\crefname{lemma}{Lem.}{Lems.}
\crefname{theorem}{Thm.}{Thms.}
\Crefname{theorem}{THM.}{THMS.}
\crefname{Corollary}{Cor.}{Cors.}
\crefname{algorithm}{Alg.}{Algs.}

\crefname{section}{Sec.}{Secs.}
\crefname{table}{Tab.}{Tabs.}
\crefname{remark}{Rem.}{Rems.}
\crefname{definition}{Def.}{Defs.}
\crefname{Proposition}{Prop.}{Props.}
\crefname{myremark}{Rem.}{Rems.}
\crefname{mylemma}{Lem.}{Lems.}
\Crefname{mylemma}{LEM.}{LEMS.}
\crefname{mydefinition}{Def.}{Defs.}
\crefname{myproposition}{Prop.}{Props.}
\Crefname{myproposition}{PROP.}{PROPS.}
\crefname{mycorollary}{Cor.}{Cors.}
\Crefname{mycorollary}{COR.}{CORS.}
\crefname{myassumption}{Assum.}{Assums.}
\crefname{figure}{Fig.}{Figs.}
\crefname{myexample}{Ex.}{Exs.}
\crefname{enumi}{}{}
\crefname{name}{}{} %

\usepackage{amsmath}

\newcommand{\indicator}{\mbf 1}

\newcommand{\snorm}[1]{\Vert #1 \Vert}
\newcommand{\sinfnorm}[1]{\snorm{#1}_\infty}

\newcommand{\sless}[1]{\stackrel{#1}{\leq}}

\newcommand{\seq}[1]{\stackrel{#1}{=}}

\newcommand{\x}{x}

\newcommand{\axi}[1][i]{\x_{#1}}

\newcommand{\dirac}{\mbi{\delta}}

\newcommand{\kernel}{\mbf{k}}

\newcommand{\knorm}[1]{\Vert{#1}\Vert_{\kernel}}

\DeclareMathOperator{\mmd}{MMD}

\renewcommand{\l}{\ell}

\newcommand{\vareps}{\varepsilon}

\newcommand{\pseqxn}[1][n]{(\axi[i])_{i\geq 1}} %
\newcommand{\pseqxnn}[1][n]{(\axi[i])_{i=1}^n} %

\newcommand{\brackets}[1]{\left[ #1 \right]}
\newcommand{\bigbrackets}[1]{\big[ #1 \big]}
\newcommand{\biggbrackets}[1]{\bigg[ #1 \bigg]}

\newcommand{\parenth}[1]{\left( #1 \right)}
\newcommand{\bigparenth}[1]{\big( #1 \big)}

\newcommand{\sbraces}[1]{\{ #1  \}}
\newcommand{\braces}[1]{\left\{ #1 \right \}}
\newcommand{\bigbraces}[1]{\big\{ #1 \big \}}
\newcommand{\biggbraces}[1]{\bigg\{ #1 \bigg \}}
\newcommand{\abss}[1]{\left| #1 \right |}

\newcommand{\angles}[1]{\left\langle #1 \right \rangle}

\newcommand{\ceils}[1]{\left\lceil #1 \right \rceil}

\newcommand{\tp}{^\top}

\newcommand{\real}{\ensuremath{\mathbb{R}}}

\newcommand{\Prob}{\ensuremath{{\mathbb{P}}}}

\def\balign#1\ealign{\begin{align}#1\end{align}}
\def\baligns#1\ealigns{\begin{align*}#1\end{align*}}
\def\balignat#1\ealign{\begin{alignat}#1\end{alignat}}
\def\balignats#1\ealigns{\begin{alignat*}#1\end{alignat*}}
\def\bitemize#1\eitemize{\begin{itemize}#1\end{itemize}}
\def\benumerate#1\eenumerate{\begin{enumerate}#1\end{enumerate}}

\newenvironment{talign*}
 {\let\displaystyle\textstyle\csname align*\endcsname}
 {\endalign}
\newenvironment{talign}
 {\let\displaystyle\textstyle\csname align\endcsname}
 {\endalign}

\def\balignst#1\ealignst{\begin{talign*}#1\end{talign*}}
\def\balignt#1\ealignt{\begin{talign}#1\end{talign}}

\newcommand{\qtext}[1]{\quad\text{#1}\quad} 
\newcommand{\stext}[1]{\ \text{#1}\ } 

\let\originalleft\left
\let\originalright\right
\renewcommand{\left}{\mathopen{}\mathclose\bgroup\originalleft}
\renewcommand{\right}{\aftergroup\egroup\originalright}

\def\tinycitep*#1{{\tiny\citep*{#1}}}
\def\tinycitealt*#1{{\tiny\citealt*{#1}}}
\def\tinycite*#1{{\tiny\cite*{#1}}}
\def\smallcitep*#1{{\scriptsize\citep*{#1}}}
\def\smallcitealt*#1{{\scriptsize\citealt*{#1}}}
\def\smallcite*#1{{\scriptsize\cite*{#1}}}

\def\mbi#1{\boldsymbol{#1}} %
\def\mbf#1{\mathbf{#1}}
\def\mbb#1{\mathbb{#1}}
\def\mc#1{\mathcal{#1}}
\def\mrm#1{\mathrm{#1}}
\def\trm#1{\textrm{#1}}
\def\tbf#1{\textbf{#1}}

\def\<{\left\langle} %
\def\>{\right\rangle}

\def\implies{\quad\Longrightarrow\quad}

\def\defeq{\triangleq} %
\def\norm#1{\left\|{#1}\right\|} %
\newcommand{\twonorm}[1]{\norm{#1}_2} %

\def\what#1{\widehat{#1}}

\def\indic#1{\indicator({#1})} %
\def\bindic#1{\indicator\!\Big(\!{#1}\!\Big)} %

\def\E{\mbb{E}} %

\def\P{\mbb{P}} %

\newcommand{\Bern}{\textnormal{Bern}}

\def\indep{\perp\!\!\!\perp} %
\newcommand{\iid}{\textrm{i.i.d.}\xspace}

\providecommand{\argmin}{\mathop\mathrm{arg min}}

\ifdefined\nonewproofenvironments\else
\ifdefined\ispres\else

\newenvironment{proof-sketch}{\noindent\textbf{Proof Sketch}
  \hspace*{1em}}{\qed\bigskip\\}
\newenvironment{proof-idea}{\noindent\textbf{Proof Idea}
  \hspace*{1em}}{\qed\bigskip\\}
\newenvironment{proof-of-lemma}[1][{}]{\noindent\textbf{Proof of Lemma {#1}}
  \hspace*{1em}}{\qed\\}
\newenvironment{proof-of-theorem}[1][{}]{\noindent\textbf{Proof of Theorem {#1}}
  \hspace*{1em}}{\qed\\}
\newenvironment{proof-attempt}{\noindent\textbf{Proof Attempt}
  \hspace*{1em}}{\qed\bigskip\\}

\newcommand{\kerNN}{\textsc{kernel-NN}\xspace}

\newcommand{\gio}[1][j,1]{\Delta_{#1}}

\newcommand{\unbhdstarA}[1][1]{\overline{\mbf N}_{#1, \eta, \mc A}^\star}
\newcommand{\lnbhdstarA}[1][1]{\underline{\mbf N}_{#1, \eta, \mc A}^\star}

\newcommand{\nbhdstar}[1][1, \eta]{{\mbf N}_{#1}^{\trm{never-ad}}}
\newcommand{\lnbhdstar}[1][1, \eta]{{\underline{\mbf N}}_{#1}^{\trm{never-ad}}}

\newcommand{\overoj}{\sum_{s \neq 1} A_{1, s}A_{j, s}}

\newcommand{\unbhdstarp}{\overline{\mbf N}_{1, \eta, p}^\star}
\newcommand{\lnbhdstarp}{\underline{\mbf N}_{1, \eta, p}^\star}

\newcommand{\empdistnotag}{\mu}

\newcommand{\empdistZ}[1]{\empdistnotag_{#1}^{(Z)}}

\newcommand{\AN}{\mc A_{1}}
\newcommand{\Arho}{\mc A_{-1}}

\newcommand{\alldist}{\mc P}

\newcommand{\wholerand}{{ \mc V_{-1}, \mc D_{-1},  \mc U, \mc A}}
\newcommand{\uarand}{{ \mc U ,  \mc A}}

\newcommand{\overlapconc}{\mc E_{\trm{dist-conc}}}
\newcommand{\overlapconcp}{\mc E_{\trm{ov-conc}}}
\newcommand{\nbhdconc}{\mc E_{\trm{nhbd-conc}}}

\newcommand{\dte}{\mrm{\texttt{i}DTE}}

\newcommand{\etadirect}{\what{\eta}_{\mrm{dir}}}
\newcommand{\etacv}{\what{\eta}_{\mrm{cv}}}

\graphicspath{{figs/},{../figs/}}

\newcommand{\papertitle}{Learning Counterfactual Distributions via Kernel Nearest Neighbors}

\endlocaldefs

\begin{document}
\etocdepthtag.toc{mtchapter}
\etocsettagdepth{mtchapter}{section}

\makeatletter
\patchcmd{\@algocf@start}
  {-1.5em}
  {0pt}
  {}{}
\makeatother

\begin{frontmatter}
\title{\papertitle}
\runtitle{Learning Counterfactual Distributions via Kernel Nearest Neighbors}




\begin{aug}
\author[A,C]{\fnms{Kyuseong}~\snm{Choi}\ead[label=e1]{kc728@cornell.edu}}, 
\author[B]{\fnms{Jacob}~\snm{Feitelberg}\ead[label=e2]{jef2182@columbia.edu}},
\author[C]
{\fnms{Caleb}~\snm{Chin}\ead[label=e3]{ctc92@cornell.edu}},
\author[B]
{\fnms{Anish}~\snm{Agarwal}\ead[label=e4]{aa5194@columbia.edu}},
\and
\author[A,C]{\fnms{Raaz}~\snm{Dwivedi}\ead[label=e5]{dwivedi@cornell.edu}}
\address[A]{Cornell Tech\printead[presep={,\ }]{e1,e5}}
\address[B]{Columbia\printead[presep={,\ }]{e2,e4}}
\address[C]{Cornell University\printead[presep={,\ }]{e3}}

\end{aug}
\begin{abstract}
Consider a setting with multiple units (e.g., individuals, cohorts, geographic locations) and outcomes (e.g., treatments, times, items), where the goal is to learn a multivariate distribution for each unit-outcome entry, such as the distribution of a user's weekly spend and engagement under a specific mobile app version. 
A common challenge is the prevalence of missing not at random data---observations are available only for certain unit-outcome combinations---where the missingness can be correlated with properties of distributions themselves, i.e., there is unobserved confounding. 
An additional challenge is that for any observed unit-outcome entry, we only have a finite number of samples from the underlying distribution.
We tackle these two challenges by casting the problem into a novel distributional matrix completion framework and introduce a kernel-based distributional generalization of nearest neighbors to estimate the underlying distributions. 
By leveraging maximum mean discrepancies and a suitable factor model on the kernel mean embeddings of the underlying distributions, we establish consistent recovery of the underlying distributions even when data is missing not at random and positivity constraints are violated. 
Our nearest neighbors approach is robust to heteroscedastic noise given two or more measurements for the observed unit-outcome entries, unlike prior nearest neighbors methods with only single measurements.
Finally, when data is missing completely at random, we identify a regime in which our method is minimax optimal up to a logarithmic factor.
\end{abstract}



\end{frontmatter}



\section{Introduction}
\label{sec : Intro}

Developments of sensors and database capacities have enriched modern data sets, meaning multiple measurements of heterogeneous outcomes are collected from different units. Rich data sets arise across modern applications, ranging from online digital platforms to healthcare or clinical settings. Consider an internet retail company that is testing $T$ different pricing strategies across $N$ different geographical regions to test how they impact sales. Often, the company can only test a subset of strategies in certain geographic locations but is interested in knowing the distribution of sales under each strategy for all regions. To formalize this, we denote $i \in [N]$ as the region, $t \in [T]$ as the strategy, $A_{i, t}$ as the indicator of whether strategy $t$ is tested in region $i$, and $\mu_{i, t}$ as the corresponding sales revenue distribution. When strategy $t$ is tested in region $i$, let $X_{1:n}(i, t) \defeq \sbraces{X_{1}(i, t), \ldots, X_n(i, t)}$ denote the revenue from $n$ sales. This example can be cast as a \emph{distributional matrix completion} problem where the observations are given by the following:
\begin{align}
\qtext{for} i \in [N], t\in[T]:\quad Z_{i, t}\defeq
     \begin{cases}
        X_{1}(i, t), \ldots, X_{n}(i, t) \ \sim \ \mu_{i, t}&\qtext{if} A_{i, t}=1, \\
        \mrm{unknown}&\qtext{if} A_{i, t}=0.
    \end{cases}
    \label{model : dist matrix completion}
\end{align}
Given this data with missing observations, the practitioner is interested in estimating the whole collection of distributions $\alldist \defeq \sbraces{\mu_{i, t}}_{(i, t) \in [N]\times [T]}$. When $A_{i, t} = 0$, we have no accessible information from $\mu_{i, t}$, and when $A_{i, t} = 1$, we do not have access to the exact distribution $\mu_{i, t}$, rather only $n$ measurements from $\mu_{i, t}$ are available. 

In some settings, $A_{i,t}$ does not denote whether we have measurements, but rather a different intervention or condition for those measurements. 
Consider a mobile health app trying to learn a recommendation strategy between two exercise routines. To start, suppose the app is provided with an observational dataset where $N$ different users alternate between these two routines repeatedly over $T$ weeks, and their health activities~(say physical step counts) throughout each week are available. For each user $i \in [N]$ in week $t\in[T]$ and exercise routine $a\in \sbraces{0, 1}$, we associate a potential outcome~(e.g. health activity by recommendation) distribution $\mu_{i, t}^{(a)}$. The goal of the practitioner is to learn distributions $\mu_{i, t}^{(1)}$ and $\mu_{i, t}^{(0)}$ under the \emph{potential outcome distributional matrix completion} problem:
\begin{align}
    \qtext{for} i \in [N], t\in[T]:\quad Z_{i, t}\defeq
    \begin{cases}
        X_{1}^{(1)}(i, t), \ldots, X_{n}^{(1)}(i, t) \ \sim \ \mu_{i, t}^{(1)} &\qtext{if} A_{i, t}=1, \\
        X_{1}^{(0)}(i, t), \ldots, X_{n}^{(0)}(i, t)\ \sim \ \mu_{i, t}^{(0)}  &\qtext{if} A_{i, t}=0,
    \end{cases}
    \label{mod:potential-outcome}
\end{align}
where $X_{1:n}^{(a)}(i, t) \defeq \sbraces{X_1^{(a)}(i, t) , ..., X_n^{(a)}(i, t)}$ denote $n$ measurements from the distribution $\mu_{i, t}^{(a)}$ for both $a = 0, 1$. Problem \cref{mod:potential-outcome} is an instance of Neyman-Rubin causal model~\cite{rubin1976inference}, following conventional assumptions, such as consistency with no delayed spillover effect.

An additional challenge in these two distributional matrix completion problems is that the missingness pattern, given by $\mc A \defeq \{A_{i, t}\}_{(i, t) \in [N]\times [T]}$, is commonly not random. In other words, (i) the missing mechanism might be correlated with latent characteristics of the distributions $\alldist$, and (ii) the measurements from some unit-outcome entry might never be observed. The first condition is called missing not at random~(MNAR) and the second condition is termed violation of positivity~(or non-positivity) in the matrix completion and causal inference literature. MNAR missingness and non-positivity occur commonly in modern applications. For example, the internet retail company from above can select a fixed subset of strategies depending on the characteristics of each region or their goal of interest. In the other example, the healthcare app's recommendation strategy will likely be tailored to each user's characteristics, and some recommendations may be scheduled beforehand so as to minimize interference of the user's daily routine. 

\subsection{Our contributions and related work}
Prior strategies in matrix completion and causal inference on panel data have not considered distributional settings and often ignore MNAR settings. 
These gaps motivate our work, which builds on and contributes to three research threads: (i) generalizing matrix completion to the distributional setting, (ii) introducing distributional counterfactual inference for panel data settings with a rich set of missingness mechanisms, and (iii) leveraging kernel mean embeddings for treatment effect estimation with panel data. Overall, our contributions can be summarized as follows:

\begin{itemize}[leftmargin=*]
\item We propose a formal model for a distributional version of the matrix completion problem, where multiple measurements are available for each unit-outcome entry for observed entries and the estimand is the unit-outcome specific distribution $\mu_{i, t}$. 

\item We introduce an estimation procedure, \kerNN, which generalizes the popular nearest neighbor algorithm to the distributional setting using reproducing kernels and maximum mean discrepancies.

\item We introduce a latent factor model~(LFM) on the kernel mean embeddings~(KME) of the underlying distributions. This LFM is a key modeling assumption which allows us to provide an instance dependent bound of $\kerNN$, with a MNAR missingness pattern where what is observed can depend on the latent factors and there can  exist entries with zero probability of being observed. Under further structural assumptions, guarantees of $\kerNN$ are optimized by balancing the bias variance trade-off.

\item We identify a regime with MCAR missingness where $\kerNN$ is minimax optimal up to a logarithmic factor.

\item We apply these theoretical guarantees to establish bounds for learning a distributional level causal effect, termed an {individual distributional treatment effect, or in short $\dte$}. 

\item Lastly, we show that when only one sample per entry is available, the model and algorithm introduced here recover the scalar counterparts (for learning mean parameters) from prior works \cite{li2019nearest,dwivedi2022counterfactual} as a special case.
\end{itemize}

We now contextualize our contributions in the context of three main research threads.

\paragraph{Matrix completion}
Matrix completion methods are widely used practical tools in settings such as panel data and image denoising. Penalized empirical risk minimization and spectral methods are well established with rigorous guarantees~\cite{candes2010power, candes2012exact, hastie2015matrix, chatterjee2015matrix}. Another set of approaches are nearest neighbor methods~\cite{chen2018explaining, li2019nearest}, which are simple and scalable, making them popular in practice. These methods have generally been analyzed for matrix completion with \iid missingness, a setting known as missing completely at random~(MCAR). Matrix completion has also been recently connected to the causal inference literature, specifically with respect to  panel data, where a latent factor structure is assumed on the expected potential outcomes, and with time-dependent missingness such as staggered adoption~\cite{xu2017generalized, athey2021matrix, bai2021matrix}. Other missing-not-at-random mechanisms have also been studied in~\cite{ma2019missing, bhattacharya2022matrix, dwivedi2022counterfactual, agarwal2023causal}. In this context, our work extends the reach of matrix completion methods with the various missingness patterns stated above to the multivariate distributional settings and provides a new instance-based analysis for (kernel) nearest neighbors.

\paragraph{Kernel methods and causal inference}
Kernel methods are ubiquitous in statistics and machine learning, especially for non-parametric problems, due to their model expressivity and theoretical tractability~\cite{scholkopf2002learning, hofmann2008kernel, sejdinovic2013equivalence}. In causal inference, kernels have been extensively used in causal discovery via conditional independence testing~\cite{gretton2007kernel, laumann2023kernel} and have also been used to model mean embeddings to encode distributional information~\cite{wenliang2023distributional, szabo2016learning} and model counterfactual distributions~\cite{muandet2021counterfactual}. More recently, it has been employed in semi-parametric inference for estimating treatment effects in observational settings~\cite{chernozhukov2022automatic} and to model causal estimands that can be expressed as functions~\cite{singh2023kernel}. 
Our work extend kernel methods to model and estimate distributional causal estimands for multiple units and outcomes, a setup common in causal panel data settings. 

\paragraph{Factor models and nearest neighbors}
Causal panel data settings typically denote causal inference settings where we have multiple units and multiple measurements for a single type of outcome over time/space. A classical approach for inference in such settings is factor modeling~\cite{shah2022counterfactual,dwivedi2022counterfactual,agarwal2023causal,abadie2024doubly} which has been effective for estimating entry-wise inference guarantees. In these works, the estimand is typically a mean parameter and the estimation procedure is commonly nearest neighbors due to its interpretability in practice and theoretical traceability with non-linear factor models~\cite{dwivedi2022counterfactual,dwivedi2022doubly}. Here we extend this line of work to distributional causal panel data in a few ways: (i) our estimand is the multivariate counterfactual distribution (and not just a functional), (ii) we introduce a non-linear factor model on kernel mean embeddings of the underlying distributions, and (iii) we generalize nearest neighbors to estimate distributions rather than scalars. 

\subsection{Organization}

\cref{sec : Problem set-up and Algorithm} introduces and discusses a novel kernel based factor model. \cref{sec:algo} outlines our proposed kernel nearest neighbors~(\kerNN) algorithm and \cref{sec : Results} states guarantees for this algorithm under a variety of settings, together with its minimax optimality in an MCAR regime. \cref{sec:sim} contain empirical performance of $\kerNN$ for simulated data, and our method is applied to a real world dataset in \cref{sec:application}. The appendix contains proofs of the theoretical results, as well as some specifics on practical implementations of $\kerNN$. 

\paragraph{Notation}
We set $\mu f = \int f (x) d\mu(x)$ and let $[n] = \{ 1, 2, ..., n \}$ for any positive integer $n$. For a point $x \in \mc X$, define $\dirac_x(y) = \indic{ y = x }$ as the indicator function, so that $\dirac_X$ for any random $X$ is the Dirac measure. For a vector $v\in\real^d$, its $j$th coordinate is $v(j)$, and a vector of ones in $\real^d$ is $\mbf 1_d$. For scalars or vectors $a_i$ with index $ i \in \mc I$, $\sbraces{a_i}_{i \in \mc I}$ denotes the set $\{ a_i : i \in \mc I \}$. 
If $\mc I = [N] \times [T]$ then $[a_{i, j}]_{(i, j)\in [N]\times [T]}$ denotes an $N\times T$ matrix with $a_{i, j}$ as entries. For a vector $x$ or matrix $A$, we denote their transpose as $x\tp$ and $A\tp$, respectively. For a function $g$ of two parameters $n$ and $m$, we write $g(n,m)=O(h(n,m))$ if there exists positive constants $c$, $n_0$, and $m_0$ such that $g(n,m) \leq c h(n,m)$ for all $n \geq n_0$ and $m \geq m_0$ \citep{cormen2022introduction}. We write $\Tilde{O}$ to hide any logarithmic factors of the function parameters.

\section{Background and problem set-up}
\label{sec : Problem set-up and Algorithm}

In this section, we give a brief summary on reproducing kernels and related concepts. We then state the target parameter of interest for the distributional matrix completion problem, along with key modeling assumptions and the data-generating process. Specific examples that are compatible with our modeling assumptions are described.

\subsection{Background on reproducing kernels}
\label{sec : background on kernels}
Our distributional learning set-up utilizes kernels throughout, and hence we provide a brief review here; we refer the readers to \cite{muandet2017kernel} for a detailed exposition. For $\mc X \subset \real^d$, a reproducing kernel $\kernel : \mc X \times \mc X \to \real$ is a symmetric and positive semi-definite function, i.e., $\kernel(x_1, x_2) = \kernel(x_2, x_1)$ and the Gram matrix $[\kernel(x_i, x_j)]_{i, j \in [n]}$ is positive semi-definite for any selection of a finite set $\{ x_1, ..., x_n \} \subset \mc X$. For any such kernel $\kernel$, there exists a unique reproducing kernel Hilbert space $(\mc H, \langle \cdot, \cdot \rangle_{\kernel})$ and a feature map $\Phi : \mc X \to \mc H$ such that $\kernel(x, y) = \langle \Phi(x), \Phi(y) \rangle_{\kernel}$ and $\langle f, \kernel(\cdot, x)\rangle_{\kernel} = f(x)$ for all $x, y \in \mc X$ and $ f\in \mc H$. Hilbert norm induced by kernel $\kernel$ is denoted here as $\| \cdot \|_{\kernel}.$ We use $T_{\kernel}$ to denote the operator that takes a distribution $\mu$ to its kernel mean embedding $\mu\kernel \in \mc H$ as follows:
\begin{align}
T_\kernel : \mu \mapsto \mu\kernel(\cdot) \defeq \int \kernel(x, \cdot) d\mu(x).
\end{align}
When $\kernel$ is characteristic, the mapping $T_\kernel$ is one-to-one~\cite{muandet2017kernel}, and under this condition we occasionally write $\mu$ to both refer to the distribution and its embedding $\mu\kernel$ when there is sufficient context to differentiate between the two. Finally, for a reproducing kernel $\kernel$ and two distributions $\mu$ and $\nu$, the maximum mean discrepancy~($\mmd$) is defined as
\begin{align}\label{eq:def_mmd}
    \mmd_\kernel(\mu, \nu) \defeq \sup_{f : \|f \|_{\kernel}\leq 1} \left| \int f d \mu(x) - \int f d\nu(x) \right| 
    {=} \| \mu - \nu \|_{\kernel},
\end{align}
where notably the last equality is known to follow from Cauchy-Schwarz inequality. A few common examples of kernels include polynomial kernels $\kernel(x, y) = (x\tp y + 1)^q$ and exponential kernels $\kernel(x,y) = \exp(-\norm{x-y}_2^2/\sigma^2)$.

Depending on $\kernel$, $\mmd$ effectively measures the weighted distance between the moments of the two distributions, e.g. for two probability measures $\mu, \nu$ on $\real$ and the square polynomial kernel $\kernel(x, y) = (xy + 1)^2$,
the kernel norm expression of MMD in \cref{eq:def_mmd} and the linearity of inner product implies $\mmd_\kernel^2(\mu, \nu) = (\E[X^2] - \E[Y^2])^2 + 2(\E[X] - \E[Y])^2$ where $X \sim \mu, Y \sim \nu$. An analogous argument holds for any polynomial kernels $\kernel(x, y) = (xy + 1)^q$ on any two measures $\mu, \nu$ on $\real$, where in this case $\mmd_\kernel^2(\mu, \nu)$ effectively measures the weighted distance between the $q$th order moments of the two distributions. 

It is well-known that when $\int \kernel(x, x) d\mu(x) <\infty$ (known as Mercer's condition), the pair $(\kernel, \mu)$ has an eigen-expansion of the form $ \kernel(x, y) = \sum_{j = 1}^{\infty}\lambda_j \phi_j(x) \phi_j(y)$, where $\lambda_1 \geq \lambda_2 \geq \ldots$ denote the eigenvalues and $\sbraces{\phi_j}_{j\in\mbb N}$ taken to be an orthonormal basis of $L^2(\mu)$, denote the eigenfunctions. Note $\{\sqrt{\lambda_j}\phi_j\}_{j \in \mbb N}$ is an orthonormal basis of $\mc H$ as well. 

\subsection{Estimand}
\label{sec : estimand}

For the problem formalized in observational model \cref{model : dist matrix completion}, our goal is to estimate the distribution $\mu_{i, t}$ for each $i \in [N]$ and $t \in[T]$. For $(i, t)$th entries with $A_{i, t} = 0$, this means estimating the distribution without any directly observed data, and for entries with $A_{i, t} = 1$, our goal is to provide a better estimate of $\mu_{i, t}$ than the empirical distribution $\frac{1}{n} \sum_{k=1}^n\dirac_{X_{k}(i, t)}$\footnote{Empirical distribution assigns uniform measure $1/n$ to all $n$ measurements.}. 

For an output of some algorithm $\what \mu_{i, t}$ that aims to learn the estimand $\mu_{i, t}$, we evaluate its performance via the $\mmd$ metric, 
\begin{align}\label{eq:mmd}
    \mmd_{\kernel}(\mu_{i, t}, \what{\mu}_{i,t}) = \norm{\mu_{i, t}- \what{\mu}_{i,t}}_{\kernel}.
\end{align}
Notably, the choice of the kernel determines what the metric \cref{eq:mmd} evaluates. Depending on the application, some may only be interested on how the mean of the estimator $\hat{\mu}_{i, t}$ approximates that of $\mu_{i, t}$, while others may care about the performance of $\hat{\mu}_{i, t}$ in approximating $\mu_{i, t}$ beyond the mean~(e.g. variance, skewness, quantile etc).

To be specific, when one is interested in the mean performance of $\hat \mu_{i, t}$, a linear kernel $\kernel(x, y) = xy$ yields\footnote{We set $d = 1$ in this subsection, which is without loss of generality.} the squared $\mmd$ metric
\begin{align}
    \mmd_\kernel^2(\mu_{i, t}, \hat{\mu}_{i,t}) = \bigparenth{\E[X] - \E[Y]}^2 \qtext{where $X \sim \hat{\mu}_{i, t}, Y \sim \mu_{i, t}$,}
\end{align}
which measures the mean difference of the distributions $\hat{\mu}_{i, t}$ and $\mu_{i, t}$.
When one is further interested on how $\what{\mu}_{i, t}$ can approximate the target up to the second moment, choosing second order polynomial $\kernel(x, y) = (xy + 1)^2$ yields squared $\mmd$ metric
\begin{align}
    \mmd_\kernel^2 (\mu_{i, t}, \what\mu_{i, t}) = \bigparenth{\E[X^2] - \E[Y^2]}^2 + 2 \bigparenth{\E[X] - \E[Y]}^2 \qtext{where $X \sim \hat{\mu}_{i, t}, Y \sim \mu_{i, t}$,}
\end{align}
hence measuring the weighted distance between the first and the second moments of the two distributions. One can extrapolate the discussion to an arbitrary $q$th order polynomial kernels $\kernel(x, y) = (xy + 1)^q$, and the metric becomes the weighted distance up to the $q$th order moments of the estimator $\what{\mu}_{i, t}$ and the estimand $\mu_{i, t}$. When the application requires evaluation of $\what{\mu}_{i, t}$ up to arbitrarily higher moments or multiple quantiles~(i.e. overall shape of the density), then the exponential kernel $\kernel(x, y) = \exp(-(x - y)^2/\sigma^2)$ yields an appropriate metric. 

Regarding the estimand and the evaluation metric of interest, additional remarks are in order.
\paragraph{Choice of kernel for the proposed algorithm} 

Our proposed method $\kerNN$~(see \cref{sec:algo} for a formal discussion) is an extension of the nearest neighbors algorithm~\cite{li2019nearest}, and the kernel is one of the inputs that characterizes the method's behavior. Roughly speaking, whenever $\kerNN$ is employed with a $q$th order polynomial kernel $\kernel(x,y) = (xy + 1)^q$, $\kerNN$ identify entries $(j, s)$ as neighbors whose distribution $\mu_{j, s}$ are close to the target distribution $\mu_{i, t}$ up to the $q$th order moments. We assume here on and after that the kernel chosen for $\kerNN$ matches the kernel used in the evaluation metric \cref{eq:mmd}. Our assumption simply states that the algorithm is to be evaluated according to its purpose and design.

\begin{figure}[t]
    \centering
    \includegraphics[width=0.95\linewidth]{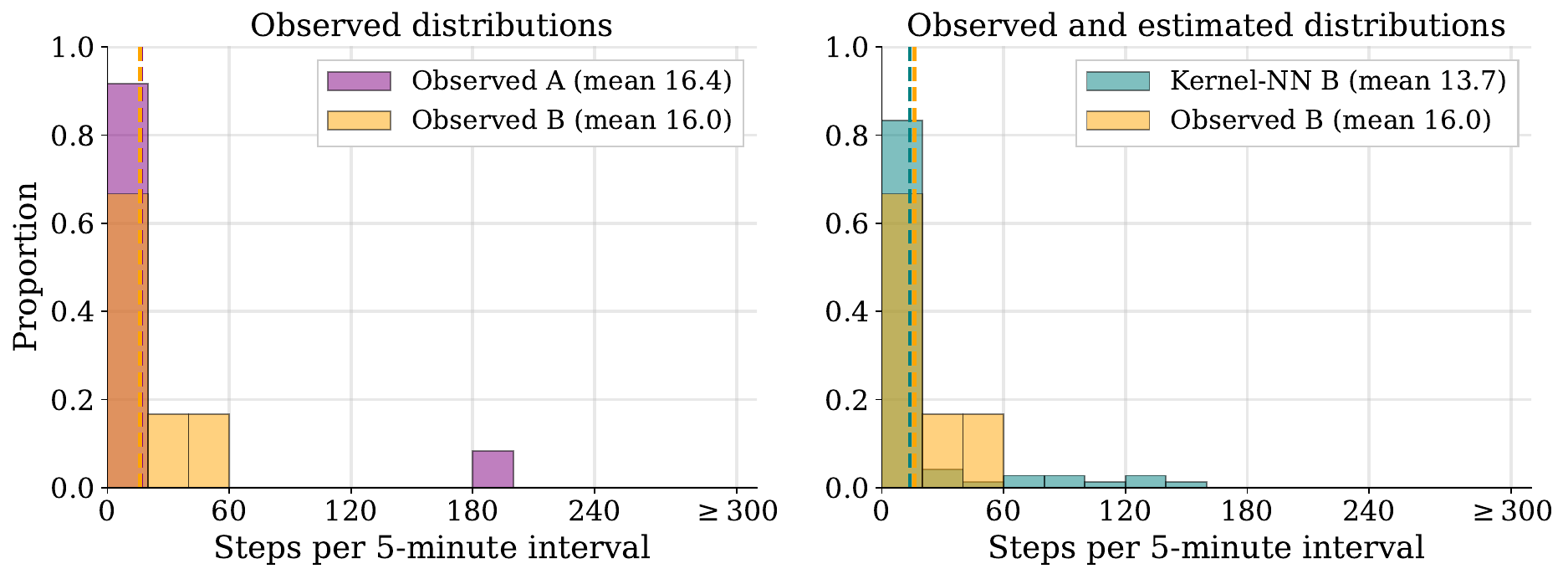}
    \caption{\tbf{HeartSteps app user's five-minute step count distribution} Each figures contain information of the step counts for different participants in the HeartSteps study~\cite{klasnja2019efficacy}~(see \cref{sec:application} for details). 
    Left panel contains their five-minute step count distribution for two different participants who received notification, using twelve measurements from the following hour.
    The right panel compares the observed five-minute step count distribution for one of the participants from the left panel following a notification with the estimated (using $\kerNN$) counterfactual step count distribution had no notification been sent at the same decision point. The dashed lines are the averages of the histograms with corresponding colors.
    } 
    \label{fig : justify kte}%
\end{figure}

\paragraph{Why a distribution as an estimand} We briefly motivate the significance of setting the distribution $\mu_{i, t}$ as the estimand through an example from the HeartSteps study~\cite{klasnja2019efficacy}. Here we give a summary of the study, and we refer the reader to \cref{sec:application} for more details. In the study, a health app sends out notifications, with some probability, to encourage physical activity to available participants at five scheduled decision points per day, and their physical step counts are recorded per minute. At available decision points, the treatment is sending a notification. For the figure, each observed outcome comprises twelve five-minute step counts during the hour following a decision point. The left panel in \cref{fig : justify kte} illustrates the step count distributions for two different participants that were notified by the app at certain time points in the study. We cannot observe the counterfactual step counts of these participants for the hypothetical scenario when notifications were not sent, hence they need to be estimated. The right panel contains the estimated (via $\kerNN$) histogram of the counterfactual step count for one of the participants in the left panel.

The distribution of the observed step counts for the two participants in the left panel of \cref{fig : justify kte} are distinct in shape despite their averages~(dashed lines in yellow and purple) being similar. A comparison of averages, as in mean estimation using scalar matrix completion~\cite{chatterjee2015matrix,mazumder2010softimpute,agarwal2023causal,dwivedi2022counterfactual}, 
would not capture these differences in distributional shape (e.g., variance, skewness). In a similar manner, when it comes to studying the treatment effect, a comparison of average outcomes~\cite{imbens2015causal} estimated by causal matrix completion techniques~\cite{agarwal2023causal,dwivedi2022counterfactual,athey2021matrix} would compare the two means shown by the dashed lines in the right panel of \cref{fig : justify kte} but would not capture differences in the concentration near zero.
However, we can see these differences when we compare the histograms as a whole. For the participant on the right panel of \cref{fig : justify kte}, the estimated step-count distribution without a notification has a higher proportion of zero-step intervals than the observed distribution following notification---this could be an actionable insight utilized for designing notification schedules. 

\paragraph{Individual distribution treatment effect} 
Motivated by the previous discussion, we formalize the treatment effect that compare distributions as a whole. For that end, we borrow the potential outcome model \cref{mod:potential-outcome}. In this case, our distributional causal estimand is defined as 
\begin{align}
    \dte_{i, t} \defeq \snorm{\mu^{(1)}_{i, t} - \mu^{(0)}_{i, t}}_{\kernel},
    \label{eq:dte}
\end{align}
which is simply the $\mmd$ distance between the potential outcome distributions. From the previous discussion, given the choice of a kernel, we observe that $\dte_{i, t}$ measures a weighted distance of the moments of the two distributions. Such distributional treatment effects have been studied in some prior works under a non-matrix setting~\cite{muandet2021counterfactual}---our discussion focuses on estimating the distributional treatment effect under the panel data setting with MNAR patterns. 

Whenever estimators $\what\mu^{(1)}_{i, t}$ and $\what{\mu}_{i, t}^{(0)}$ are available for the distributions $\mu_{i, t}^{(1)}$ and $\mu_{i, t}^{(0)}$ respectively, we propose a meta-estimator that simply takes the RKHS norm of the difference of the estimators
\begin{align}
    \what{\dte}_{i, t}\defeq \snorm{\what\mu^{(1)}_{i, t} - \what\mu^{(0)}_{i, t}}_{\kernel}.
\end{align}
We note that for this case, guarantees on the individual estimators, $\what\mu^{(1)}_{i, t}$ and $\what{\mu}_{i, t}^{(0)}$, directly translate to a guarantee of $\what{\dte}_{i,t}$ via the triangle inequality:
\begin{align}\label{eq:KTE-triangle}
    |\dte_{i, t} - \what{\dte}_{i, t}| \leq  \snorm{\mu^{(1)}_{i, t} - \what\mu^{(1)}_{i, t}}_{\kernel}
    +  \snorm{\mu^{(0)}_{i, t} - \what\mu^{(0)}_{i, t}}_{\kernel}.
\end{align}
As is the case for \cref{eq:KTE-triangle}, analysis on the observation model \cref{model : dist matrix completion} can be applied without much modification to the potential outcome model \cref{mod:potential-outcome}.

\subsection{Modeling assumptions}
\label{sec : model}
We introduce structural assumptions made on the model \cref{model : dist matrix completion} which we use for a rigorous analysis of our method. First we discuss a factor structure on the collection of distributions that reduces the number of unknowns in our problem. Next, we describe the assumptions on the missing mechanism of $A_{i, t}$. Finally, we introduce a natural data generating process that is consistent with these assumptions.

\begin{assumption}[Latent factor model on kernel mean embeddings]
\label{assump:factorization}
    There exists a set of row latent factors $\mc U \defeq \sbraces{u_i}_{i \in [N]} \subset \real^r$, column latent factors $\mc V \defeq \sbraces{v_t}_{t\in [T]} \subset \real^r$ and an operator $g: \real^r \times \real^r \to \mc H$, such that the kernel mean embeddings of the distributions $\alldist$ satisfy a factor model as follows: {for the kernel $\kernel$ used in the metric  \cref{eq:mmd},}
    \begin{align}
        \mu_{i, t}\kernel = g(u_i, v_t).
        \label{eq:factor_model}
    \end{align}
\end{assumption}

We briefly discuss the implications of \cref{assump:factorization}. When a practitioner settles on a evaluation metric \cref{eq:mmd} by specifying a kernel $\kernel$, \cref{assump:factorization} hypothesizes the existence of a factor model on the distribution embeddings embedded by the \textit{same} kernel. Loosely speaking, when the first $q$ moments are what one cares about~(i.e. $\kernel(x, y) = (x^Ty + 1)^q$ is used for the metric \cref{eq:mmd}), then \cref{assump:factorization} implies that the moments of $\mu_{i, t}$ up to the $q$th order are factored via some latent factors $u_i, v_t$. 

\cref{assump:factorization} assumes only the existence of an operator $g$. We can make the model \cref{eq:factor_model} more interpretable by specifying the form of $g$. For instance, suppose one is interested only on the mean approximation of $\mu_{i, t}$, thereby fixing a linear kernel $\kernel(x, y)= xy$ for the evaluation metric \cref{eq:mmd}. Then the existence of latent factors $u_i, v_t$ and a real valued mapping $m_1(u_i, v_t)$ satisfying $g(u_i, v_t)(y) = m_1(u_i, v_t)\cdot y$ for \cref{eq:factor_model} implies a factor model on the mean of distribution, which is clear by observing $\int x d\mu_{i, t}(x) \cdot y = m_1 (u_i, v_t) \cdot y$ for any $y$. So the standard mean factorization assumption made in the matrix completion~\cite{chatterjee2015matrix,li2019nearest,agarwal2023causal} and the panel data setting~\cite{abadie2010synthetic,athey2021matrix,dwivedi2022counterfactual} can be recovered via \cref{assump:factorization}.

\begin{assumption}[Independence across latent factors]
\label{assump:latent-independence}
    The latent factors ${u_1, ..., u_N}$ are drawn \iid from a distribution $\Prob_u$ on $\real^r$ and independently of ${v_1, ..., v_T}$, which in turn are drawn \iid from $\Prob_v$ defined over $\real^r$ as well.
\end{assumption}
Independence across row factors in \cref{assump:latent-independence} is a mild condition. For instance, participants in the healthcare app experiment can be independently chosen from a homogeneous super-population. Independence across column factors in \cref{assump:latent-independence} is a more stringent condition as different outcomes for the same unit might have dependence over each other. In Appendix~\ref{sec:dependent-column-appendix}, we relax the independence across column factors and derive error bounds for $\kerNN$ when the latent column process is stationary and $\beta$-mixing\footnote{In \cref{subsec:dependent-columns}, we compare the error of $\kerNN$ under independent column factors with its error under AR$(1)$ and AR$(2)$ column factors, under both staggered adoption and MCAR.}
The next assumption models non-positivity and unobserved confounding in the missingness patterns, by allowing missingness to depend on the latent row factors.

\begin{assumption}[Selection on row latent factors] 
\label{assump : unobs confounding}
Conditioned on the row factors $\mc U$, the missingness $ \mc A \defeq \sbraces{A_{i, t}}_{(i, t)\in[N] \times [T]}$ are independent to the column latent factors $\mc V$. As a result, potential outcomes of interest are independent of the treatment, conditioned on $\mc U$. 

\end{assumption}
\cref{assump : unobs confounding} implies that row latent factors $\mc U$ can explain the unobserved confounding between the missingness $\mc A$ and the potential outcomes. For instance, in a mobile health app setting~(specifically the HeartSteps study~\cite{klasnja2019efficacy}), the interventions are given at times only when the users are available, and the available times for each units are scheduled ahead of time. The driving factor for the  available times are likely to be the daily routines or personal schedules unique to each individuals, that are otherwise not observed. 

\begin{assumption}[\iid measurements]
\label{assump : measurement generation}
    Conditioned on the latent factors $u_i, v_t,$ and $A_{i, t}=1$, the repeated measurements  $X_1(i, t), ..., X_n(i, t)$ are sampled \iid from $\mu_{i, t}$ and independently of all other randomness.
\end{assumption}
The \iid measurements are assumed in \cref{assump : measurement generation} for convenience of analysis. Our analysis under \iid measurements can be extended without much modification to account for dependent measurements, as long as some type of concentration is allowed~(e.g. Martingales with bounded differences~\cite[Thm.~2.19]{wainwright2019high}). We do not provide a formal discussion regarding dependent measurements to keep the discussion concise.

\paragraph{A data generating process } We outline an example of a data generating process for the observational setting in ~\cref{model : dist matrix completion}, which is consistent with \cref{assump:factorization,assump:latent-independence,assump : unobs confounding,assump : measurement generation}~(see \cref{fig : dgp dag} for graphical representation), 

\begin{enumerate}[leftmargin=*]
\item \emph{Latent factors} :
Row latent factors $\mc U$ and column latent factors $\mc V$ are generated through the mechanism of \cref{assump:latent-independence}. {For some fixed kernel $\kernel$ and the RKHS $\mc H$ generated by $\kernel$,} the distribution $\mu_{i, t}$ is determined by an unknown mapping $g: \real^r \times \real^r \to \mc H$ and latent factors $u_i, v_t$, via $\mu_{i, t}\kernel = g(u_i, v_t)$, so that \cref{assump:factorization} holds~(i.e. $\mu_{i, t} = T_\kernel^{-1} g( u_i, v_t )$ if $\kernel$ is characteristic).

\item \emph{Missing mechanism} : Given latent factors $\mc U$, missing indicators $A_{i, t}$ are generated by some mechanism that respects \cref{assump : unobs confounding}.
\item \emph{Repeated measurements} : If $A_{i, t}=1$, then the vectors $X_{k}(i, t) \in \mc X \subset \real^d$ for $k \in [n]$ are sampled from the distribution $\mu_{i, t}$, as in \cref{assump : measurement generation}.

\definecolor{offwhite}{HTML}{F2EDED}

\tikzset{
    > = stealth,
    every node/.append style = {
        text = black
    },
    every path/.append style = {
        arrows = ->,
        draw = black,
        fill = offwhite
    },
    hidden/.style = {
        draw = black,
        shape = circle,
        inner sep = 1pt
    }
}

\begin{figure}[tb]
\centering
\tikz{
    \node (a) at (0,1) {$\mc A$};
    \node[hidden] (b) at (1, 2) {$\mc U$};
    \node[hidden] (c) at (2,1) {$\alldist$};
    \node (e) at (1,0) {$\sbraces{ X_{1:n}(i, t) }$};
    \node[hidden] (d) at (2, 2) {$\mc V$};
    \path (b) edge (a);
    \path (c) edge (e);
    \path (b) edge (c);
    \path (a) edge (e);
    \path (d) edge (c);
}

\caption{Data generating process of observational model \cref{model : dist matrix completion}. Circled $\mc U$, $\mc V$, and $\alldist$ are the unobserved, $\mc U$ is the common cause~(confounder) for the observed missingness $\mc A$ and measurements $\sbraces{X_{1:n}(i, t)}$.}
\label{fig : dgp dag}
\end{figure}

\end{enumerate}

\subsection{Distribution families satisfying \cref{assump:factorization}}
\label{sec:examples}

Here we present two examples for families of distributions that satisfy the kernel mean embedding factorization of \cref{assump:factorization}. The examples specify the explicit form of the operator $g$.

 \begin{example}[Location-scale family]
 \label{exam:gaussian_family}
     Suppose $\alldist$ is a location-scale family on $[-1,1]^d$. Assume a second order polynomial kernel $\kernel(x, y) = (x^Ty + 1)^2$ for both the metric \cref{eq:mmd} and the factor model \cref{eq:factor_model}. Let $u_i,v_t\in[-1,1]^2$, and write $a_{i,t}=u_{i,1}v_{t,1}$ and $b_{i,t}=u_{i,2}v_{t,2}$. Observations sampled from $ \mu_{i,t}$ have independent coordinates, and coordinate $j$ of the observation takes either of the two values $(-1)^j a_{i,t}/2 \pm 2^{-j/2}b_{i,t}/2$ each with probability $1/2$. So $a_{i,t}$ controls the location and $b_{i,t}$ the scale. Its kernel mean embedding has the form
     \begin{align}\label{eq:moment_factor}
         g(u_i, v_t)(y) = 1+a_{i,t}\sum_{j=1}^d(-1)^j y_j+\frac{b_{i,t}^2}{4}\sum_{j=1}^d2^{-j}y_j^2+\frac{a_{i,t}^2}{4}\bigg(\sum_{j=1}^d(-1)^j y_j\bigg)^2,
     \end{align}
     satisfying \cref{assump:factorization}.
 \end{example}
Notably the kernel mean embedding of distribution $\mu_{i, t}$ for square polynomial kernel is $\mu_{i, t}\kernel (y) = \int \kernel(x, y) d\mu_{i, t}(x) = y^T\big(\int xx^T d\mu_{i, t}(x)\big)y + 2 y^T\int x d\mu_{i, t}(x) + 1$, hence \cref{exam:gaussian_family} implies that the first and second moments are factorized via $\E[X_j]=(-1)^j a_{i,t}/2$ and $\E[X_j^2]=(a_{i,t}^2+2^{-j}b_{i,t}^2)/4$ for $X\sim\mu_{i,t}$. Independence of the coordinates also gives $\E[X_jX_\ell]=(-1)^{j+\ell}a_{i,t}^2/4$ for $j\neq\ell$. Thus the quadratic term includes both the covariance and the outer product of the mean. While the prior example covers a finite-dimensional class of distributions where only first and second moments are considered, our next example shows that the factor model assumption also covers a wide-range of infinite-dimensional class of distributions. Recall that $\psi_j = \sqrt{\lambda_j}\phi_j$ serves as the orthonormal basis of $\mc H$ constructed from the Mercer kernel $\kernel$, which we assume for the following examples.

\begin{example}[Infinite-dimensional family]
    \label{exam:infinite_family}
    Suppose the distributions in $\alldist$ are non-parametric on $\real^d$, meaning that each $\mu_{i, t}$ is characterized not only by its mean and covariance, but by all the higher order moments. Assume an exponential kernel $\kernel(x, y) = \exp(-\|x - y\|_2^2/2)$ for both the metric \cref{eq:mmd} and the factor model \cref{eq:factor_model}. Let $\sbraces{\psi_j}_{j \in \mbb N}$ be the $\mc H$-basis of $\mu_{i, t}\kernel$. Assume there exist latent factors $u_i, v_t \in \real^r$ and an operator $g$ of the form
    \begin{align}\label{eq:coeff_factor}
        g(u_i, v_t)(y) = \sum_{j = 1}^\infty \alpha_j(u_i, v_t) \psi_j(y)
    \end{align}
    satisfying \cref{assump:factorization}, where $\alpha_j: \real^r \times \real^r \to \real$ are $L_j$ lipschitz functions, i.e. $|\alpha_j(u, v) - \alpha_j(u', v')| \leq L_j\cdot ( \|u - u'\|\vee \|v - v'\| )$. 
    
\end{example}
Exponential kernel satisfies the Mercer condition and the corresponding RKHS has an orthonormal eigen-basis $\psi_j$~(see \cite[Thm. 4.38]{steinwart2008support} for closed form of basis), allowing an expansion of the kernel mean embedding $\mu_{i, t}\kernel = \sum_{j = 1}^{\infty} \langle \mu_{i, t}\kernel, \psi_j\rangle_\kernel \psi_j$. So we observe that the $j$th basis coefficients of the embedding $\mu_{i, t}\kernel = g(u_i, v_t)$ are factored by $\alpha_j$.

\section{\kerNN\ Algorithm}
\label{sec:algo}

We next describe the primary algorithmic contribution of this work: kernel nearest neighbors, or $\kerNN$ in short, for estimating the distribution $\mu_{i, t}$. We briefly review the nearest neighbors presented in \cite{li2019nearest} used for scalar matrix completion, when at most a single measurement of dimension $d = 1$ is available per matrix cell from the observational model \cref{model : dist matrix completion}, say $X_1(j, s) \in \real$.
The following procedure adapted from \cite{li2019nearest}, and typical of how nearest neighbor methods are used in collaborative filtering applications, aims to learn the first moment of the distribution $m_{i, t}=\int x d\mu_{i, t}(x)$. 
\paragraph{Row-wise scalar nearest neighbors:}
\begin{itemize}[leftmargin=0.7cm]
    \item[(1)] (Distance between rows) For any row $j \neq i$, calculate an averaged squared Euclidean distance across overlapping columns,
    \begin{align}\label{eq:snn-distance}
        \varrho_{i, j} = \frac{\sum_{s \neq t}A_{i, s}A_{j, s}(X_1(i, s) - X_1(j, s))^2}{\sum_{s \neq t} A_{i, s}A_{j, s}}.
    \end{align}
    \item[(2)] (Average across observed neighbors) For row-wise neighbors $\sbraces{ j \neq i : \varrho_{i, j} \leq \eta}$ within $\eta$ radius, average across observed neighbors within $t$th column,
    \begin{align}\label{eq:snn-average}
        \what{m}_{i, t, \eta} = \frac{\sum_{j \neq i : \varrho_{i, j}\leq \eta}A_{j, t}X_1(j, t)}{\sum_{j \neq i : \varrho_{i, j}\leq \eta}A_{j, t}}.
    \end{align}
\end{itemize}
The fact that nearest neighbors target a single entry at a time via matching makes it effective against various types of missing patterns---the algorithm was extended and generalized since, to account for a wide range of applications~\cite{dwivedi2022counterfactual, dwivedi2022doubly, agarwal2023causal}, with a focus on inference for personalized treatment effects in the causal inference literature. 

We extend the scalar nearest neighbors algorithm to handle distribution imputation, and we do so by extending the notion of distance in \cref{eq:snn-distance} and average in \cref{eq:snn-average} so that it is suitable for handling multi-dimensional distributions. In essence, the squared Euclidean distance of single measurements in \cref{eq:snn-distance} is substituted by the $\mmd$ distance between the empirical distributions of multiple measurements, and the Euclidean average of single measurements in \cref{eq:snn-average} is substituted by the barycenter of the empirical distribution of multiple measurements within a given neighborhood. 

Now we formalize our proposed distributional nearest neighbors algorithm $\kerNN$. We refer the reader to \cref{sec : fully general kerNN} for the most general version of $\kerNN$, which is applicable to both models \cref{model : dist matrix completion} and \cref{mod:potential-outcome}, but here we present a version of $\kerNN$ that is specifically applied on model \cref{model : dist matrix completion}. The observed outcome of multiple measurements $Z_{j, s} = \sbraces{X_1(j, s), ..., X_n(j, s)}$ in model \cref{model : dist matrix completion} is equivalently denoted as the empirical distribution\footnote{We emphasize that the collection of measurements, and the empirical distribution of the same measurements contain exactly the same information.}
\begin{align}
    \empdistZ{j, s} \defeq \frac{1}{n} \sum_{\ell = 1}^{n} \dirac_{X_{\ell}(j, s)} , \quad \text{for} \quad A_{j, s} = 1.
\end{align}
Set the input of $\kerNN$ as the kernel $\kernel$, observed outcomes $\mc Z \defeq \sbraces{ Z_{i, t} : A_{i, t} = 1}$, the missingness $\mc A$, hyper-parameter $\eta > 0$ and the index $(i, t)$ of the target distribution $\mu_{i, t}$. Then, \kerNN, with $n\geq 2$ measurements for each observed outcome, is described in the following two steps:
\paragraph{$\kerNN (\kernel, \mc Z, \mc A, \eta, i, t)$:}
\begin{enumerate}[leftmargin=0.6cm]
    \item[(1)] \tbf{Distance between rows via unbiased-MMD estimator}: First we estimate the row-wise distance $\rho_{i, j}$, as the averaged squared estimated MMD between the empirical distributions corresponding to unit $i$ and $j (\neq i)$ across the indices $[T]\backslash\{t\}$:
   \begin{align}
    \rho_{i,j} &\defeq \frac{\sum_{s \neq t} A_{i, s}A_{j,s}\what{\mmd}^2_{\kernel}(\empdistZ{i, s},\empdistZ{j, s})  }{\sum_{s \neq t} A_{i,s}A_{j,s}}, \ \text{where} 
\label{eq:row-metric} \\
\what{\mmd}^2_{\kernel}(\empdistZ{i, s},\empdistZ{j, s}) &\defeq \frac{1}{n(n-1)} \sum_{\l\neq\l'} \mbf{h}(X_{\l}(i,\! s), X_{\l'}(i, \!s), X_{\l}(j, \!s), X_{\l'}(j,\! s)),\label{def : ustat_mmd} \\ 
 \qtext{and}
 \mbf{h}(x, x', y, y') &\defeq \kernel(x, x') + \kernel(y, y') -\kernel(x, y') - \kernel(x', y).
    \end{align}
 Notably, $\what{\mmd}_{\kernel}^2$ above is the standard U-statistics estimator of $\mmd_{\kernel}^2(\mu_{i,s}, \mu_{j, s})$ (see \cite[Lem.~6]{JMLR:v13:gretton12a}).
    We set $\rho_{i, j} = \infty$ whenever the denominator on the RHS of \cref{eq:row-metric} is zero. 

\item[(2)] \tbf{MMD barycenter over observed neighbors}: Next, we define the units that are $\eta$-close to unit $i$, as its neighbors $\mbf N_{i, \eta}$, where we exclude the unit from being its own neighbor:
\begin{align}
    \mbf N_{i, \eta} &\defeq \braces{ j \in [N]\setminus\{i\} : \rho_{i,j} \leq \eta}.
    \label{eq:actual_nbhd}
\end{align}
Finally, the \kerNN-estimate $\what{\mu}_{i, t, \eta}$ is given by the $\mmd$-barycenter across the row neighbors that are observed at time $t$, namely
\begin{align}
\label{eq:mmd-barycenter}
\what{\mu}_{i, t, \eta} &\defeq \argmin_{\mu } \frac{\sum_{j \in \mbf N_{i, \eta}} A_{j,t}\mmd_{\kernel}^2( \empdistZ{j, t} , \mu)}{\sum_{j \in \mbf N_{i, \eta}} A_{j,t}}\\   
&\seq{(*)} \frac{\sum_{j \in \mbf N_{i, \eta}} A_{j, t} \empdistZ{j,t}}{\sum_{j \in \mbf N_{i, \eta}} A_{j,t} } = \frac{1}{n\sum_{j \in \mbf N_{i, \eta}} A_{j,t} } \sum_{j \in \mbf N_{i, \eta}}  \sum_{\l=1}^n A_{j,t} \cdot \dirac_{X_\l(j, t)},
\end{align}
where step~$(*)$ follows directly from \cite[Prop.~2]{cohen2020estimating}. If $\sum_{j\in\mbf N_{i,\eta}}A_{j,t}=0$, we use the zero measure or a default probability measure, such as a mixture of the empirical distributions observed at time $t$, when available.
\end{enumerate}
In the above calculations, we do not use $t$-th column's data in estimating distances step~(1); such a sample-split is for ease in theoretical analysis. Moreover, for brevity in notation, we omit the dependence of $\rho_{i, j}$ and $\mbf N_{i,\eta}$ on $t$. 

\begin{remark}
\label{rem:truncatv_nbd}
    In practice, when estimating ${\mu}_{i, t}$ we can restrict the search space for nearest neighbors only over the units $j \in [N]$ such that $\sum_{s\neq t}A_{i, s}A_{j, s} \geq \underline{c}$ for some large choice of $\underline{c}$ to ensure that the distance $\rho_{i, j}$, is estimated reliably. We can further restrict the computations solely to units $j$ with $A_{j, t}=1$ to further reduce computational overhead. 
\end{remark}

\paragraph{Choice of hyper-parameter $\eta$}
Our theory shows that naturally the hyper-parameter $\eta$ characterizes the bias-variance of the \kerNN\ estimate and needs to be tuned. Our theoretical results (\cref{prop:most-raw-bound,thm:stagger-bound,thm:prop-bound}) characterize the error guarantees as a function of any fixed $\eta$, {and in practice we propose two different strategies to choose $\eta$: the first strategy is principled as it relies on the theoretical guarantees of $\kerNN$~(see discussion after \cref{prop:most-raw-bound} and \cref{sec:sim}), and the second strategy is based on the generic cross-validation~(see \cref{app:sim}).}

\paragraph{Computational and storage complexity}
For any fixed $\eta$, computing $\rho_{i, j}$ takes $O(n^2T)$ kernel evaluations, where a kernel evaluation takes typically $O(d)$ time when the measurements are in $\real^d$. Moreover, querying the kernel mean embedding for any small value at any point in the outcome space requires $O(Nn)$ kernel evaluations. Saving the distances requires $O(N^2)$ memory and saving the distribution support points requires $O(Nn)$ memory. Thus overall computational complexity of the \kerNN\ algorithm is $O(NT n^2d)$ operations and $O(N^2)$ storage.

\paragraph{Generalization of prior work}

We elaborate how our work generalizes the prior work of scalar nearest neighbors to a distributional setting. Specifically we show how our work recovers prior factor models and algorithms for scalar matrix completion as a special case. For example, the set-up of \cite{li2019nearest,dwivedi2022counterfactual} can be cast in our framework under model \cref{model : dist matrix completion} with one measurement per entry, i.e., $n = 1$ so that only $X_1(i, t)$ is available when $A_{i, t} = 1$. In this case, since the U-statistics are not defined, using V-statistics~\cite{muandet2017kernel} as the MMD measure in \cref{eq:row-metric}, the following dissimilarity measure can be used,
\begin{align}
    \rho^{\textsc{V}}_{i, j} \defeq \frac{\sum_{s \neq t} A_{i, s} A_{j, s} \mmd_\kernel^2(\empdistZ{i, s}, \empdistZ{j, s}) }{\sum_{s \neq t}A_{i, s}A_{j, s}}.
    \label{def : modified dissim}
\end{align}
When $n = 1$, we observe that $\empdistZ{i, s} = \dirac_{X_1(i, s)}$ and $\empdistZ{j, s} = \dirac_{X_1(j, s)}$. We show in \cref{prop : generalization of prior work} in \cref{sec : generalization and hetero} that by instantiating our data generating process with single measurements~($n=1$) and using a linear kernel recovers the previously studied non-linear factor models used in scalar-valued matrix completion. Further, by choosing a linear kernel along with the biased estimate \cref{def : modified dissim} for the row metrics, $\kerNN$ recovers the nearest neighbor algorithm studied in \cite{li2019nearest,dwivedi2022counterfactual,dwivedi2022doubly}.

\section{Main Results}
\label{sec : Results}

This section presents the main results regarding the performance of $\kerNN$. We first present an instance dependent guarantee of $\kerNN$ which holds for nearly any types of missingness pattern that depend on unobserved latent variables. This bound serves as the theoretical basis to analyze $\kerNN$ under a range of important MNAR missing mechanisms studied previously in the literature. From a practical standpoint, the instance dependent bound motivates a principled and computationally efficient way of training~(i.e. choosing hyper-parameter $\eta$) the algorithm $\kerNN$. 

To show the flexibility of the instance dependent guarantee, we specify our results for different missingness models. First, we provide a result for when $\kerNN$ is applied on the widely encountered staggered adoption observation pattern seen in practice. Second, we present a guarantee of $\kerNN$ when the missingness pattern is modeled via propensities, which provides an understanding of how the proposed algorithm performs as a function of the probability of various entries being observed, and its robustness to how much positivity can be violated. We then relax the independence of the column latent factors and establish an oracle minimax lower bound for the target-column smoothing rate.

\subsection{An instance-based guarantee for \kerNN}
\label{sec : main instance}
\newcommand{\anbd}{{\mbf N}^{\star}_{1, \eta, \mc A}}
\newcommand{\errja}[1][j, \mc A]{e_{#1}}
\newcommand{\errjp}[1][j, p]{e_{#1}}
\newcommand{\errjt}[1][j, T]{e_{#1}}

Unless otherwise stated, we state our results for estimating the distribution $\mu_{1, 1}$ corresponding to $(1, 1)$-th entry, which is without loss of generality. To state our result, we introduce some additional notation. Define the squared MMD distance between the mean embeddings marginalized over the column latent factors:
\begin{gather}
    \gio \defeq \int \| g(u_j,v) - g(u_1, v) \|_\kernel ^2 d\P_v.
    \label{def : Delta and etai}
\end{gather}
For any $\delta>0$, define the two population neighborhoods as
\begin{align}
    \unbhdstarA &\defeq \braces{ j \neq 1 : \gio
    < \eta  + \errja }
    \qtext{and}
     \lnbhdstarA \defeq \braces{ j \neq 1 : \gio < \eta - \errja},
    \label{def : pop nbhd data centered}
\end{align}
where 
\begin{align}
     \errja &\defeq \frac{c_0 \| \kernel\|_\infty \sqrt{\log(2N /\delta)}}{ \sqrt{\sum_{s \neq 1} A_{1,s}A_{j, s} }} 
     \qtext{and}
     c_0 \defeq \frac{8e^{1/e}}{\sqrt{2e\log 2}};
     \label{eq:err_defn}
\end{align}
and we omit the dependence on $\delta$ in our notation for brevity. Note that $( \unbhdstarA ,\lnbhdstarA)$ depend solely on $\sbraces{\uarand}$, and in our guarantees serve as a sandwich for the neighbor set $\mbf N_{1, \eta}$ used to define the $\kerNN$ estimate, that is 
\begin{align}\label{eq:sandwich}
    \lnbhdstarA \subseteq \mbf N_{1, \eta} \subseteq \unbhdstarA.
\end{align}

We are now ready to state our first main guarantee---an instance dependent error bound on the $\kerNN$ estimate, which does not require any pre-specification of the missingness pattern, but only the confoundedness condition stated in \cref{assump : unobs confounding}. Refer to \cref{sec:proof_of_instance_based} for the proof of the following result.

\begin{proposition}[\tbf{Instance dependent guarantee}]\label{prop:most-raw-bound}
    Suppose the observed measurements and missingness from model \cref{model : dist matrix completion} respect \cref{assump:factorization,assump:latent-independence,assump : unobs confounding,assump : measurement generation}. Then for any values of $\eta,\delta > 0$, the estimator $\what{\mu}_{1, 1, \eta}$ of $\kerNN$ satisfies
    \begin{align}
        \E \brackets{\| \what{\mu}_{1, 1, \eta} \!-\! \mu_{1, 1}\|_\kernel^2 \vert \mc U, \mc A }
        \!\leq\! \eta +\!  \sinfnorm{\kernel}\biggbrackets{\!\!\max_{j \in \unbhdstarA}\!\!\! 
        A_{j, 1} \frac{c_0 \sqrt{\log(2N/\delta)}}{ \sqrt{\sum_{s \neq 1} A_{1,s}A_{j, s} }}  
        \!+\!  \frac{4 (\log n \!+\! 1.5)}{ n \sum_{j \in \lnbhdstarA} A_{j, 1} } + 4\delta}.
        \label{eq:most-raw-guarantee}
    \end{align}
\end{proposition}

Notably, the instance dependent guarantee of nearest neighbors algorithm is valid under unobserved confounding in the missingness. The terms appearing in the guarantee \cref{eq:most-raw-guarantee} warrant interpretation. The first two terms of the RHS in display \cref{eq:most-raw-guarantee} are akin to the bias. Construction of the neighborhood $\mbf N_{1, \eta}$~(defined in \cref{eq:actual_nbhd}) as the first step of $\kerNN$ is the source of the bias. The hyper-parameter $\eta$ in the bias term determines the number of heterogeneous neighbors that are averaged upon, so larger $\eta$ induces bias. The second term in the bias measures the precision of the data-driven metric $\rho_{j, 1}$ in approximating the true row-wise metric $\gio$. The definition \cref{eq:row-metric} implies that even under $n = \infty$~(i.e. access to the distribution $\mu_{i, t}$ whenever $A_{i, t} = 1$), the true distance $\gio$ can only be recovered when we have many overlapping columns---the second term reflects this observation as it does not vanish as $n$ tends to $\infty$. The third term of \cref{eq:most-raw-guarantee} is akin to the variance of the $\mmd$ barycenter over the neighborhood, which vanishes as the total number measurements increase. Larger $\eta$ would increase the number of measurements averaged upon, hence inducing smaller variance. Overall, the MMD error above expresses a bias-variance tradeoff as a function of the hyper-parameter $\eta$. 
We further make several remarks on the instance dependent bound below.

First, the instance dependent bound motivates a principled and fast optimization procedure for choosing the hyper-parameter $\eta$, which does not rely on cross-validation. Normally, cross validation on kernel-based algorithms demand high computation overload when the problem scales with the number of data points~\cite{dwivedi2021kernel,gong2024supervised,rahimi2007random,williams2000using}. The idea is to optimize $\eta$ over a \textit{computable error bound};
that is, we first substitute the non-computable components $(\lnbhdstarA, \unbhdstarA)$ in the upper bound of the $\kerNN$ guarantee \cref{eq:most-raw-guarantee} by their data-driven~(hence computable) counterpart $\mbf N_{1, \eta}$, and then we optimize the computable upper bound by $\eta$. 
The substitution of the non-computable components by the computable neighbor $\mbf N_{1, \eta}$ is justified by the sandwich relationship \cref{eq:sandwich}, which is shown to hold with high probability~(see \cref{eq:n_ubnd,eq:n_lbnd}).

Second, \cref{prop:most-raw-bound} also serves as an instance dependent error bound for the scalar nearest neighbors when estimating the mean parameters in a non-parametric factor model with unobserved confounding, when there are $n\geq 2$ samples in each entry~(see \cref{sec : generalization and hetero} for a discussion when $n=1$); this is because we can recover both the canonical scalar matrix completion setting and the scalar nearest neighbor algorithm with a linear kernel $\kernel$, see \cref{prop : generalization of prior work} for details. 

Third, many prior works on nearest neighbors for scalar matrix completion either require the noise variance to be identical across entry~\cite{li2019nearest,dwivedi2022counterfactual} or require a uniform upper bound on the noise variance~\cite{agarwal2023causal,agarwal2020synthetic}, in order to derive a non-vacuous error guarantee for the mean parameters. With $n\geq 2$ measurements per observed entry, \cref{prop:most-raw-bound} bounds the expected squared $\mmd$ between the $\kerNN$ estimate and the target distribution under the chosen kernel, while allowing variances and higher moments to vary across entries.
This flexibility with $n\geq 2$ samples in each observed entry arises from our choice to use U-statistics to construct unbiased estimates of distances in $\kerNN$\footnote{This claim can also be seen when comparing \cref{prop:most-raw-bound} with prior guarantees, e.g., \cite[Thm.~1]{dwivedi2022doubly} where the leading bias term is $\eta-2\sigma^2$ where $\sigma^2$ is the variance, and the corresponding term in \cref{prop:most-raw-bound} is simply $\eta$, independent of the noise variances.}.

\begin{figure}[tb]
    \centering
    \includegraphics[width=0.95\linewidth]{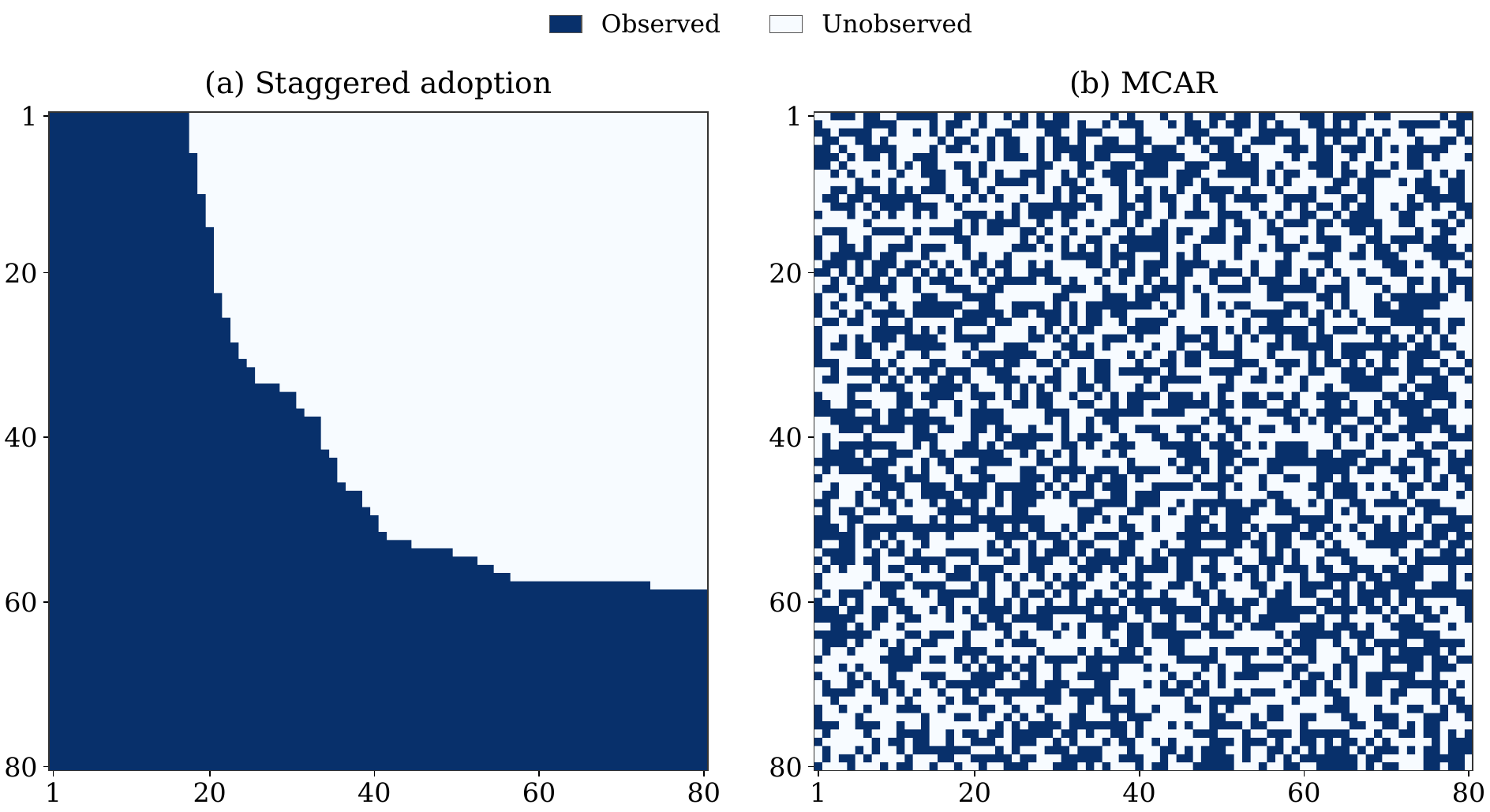}
    \caption{\tbf{Observation patterns under staggered adoption and MCAR.} Dark blue indicates observed entries and near-white indicates unobserved entries. Rows represent units and columns represent time points. (a) Observation of control potential outcomes under staggered adoption, with units ordered by adoption time; the bottom rows remain untreated throughout. (b) Entries are observed independently with probability $p=0.5$. See \cref{app:sim} for simulation details.}%
    \label{fig:missingness}%
\end{figure}

\subsection{Distributional recovery under staggered adoption}
\newcommand{\adtime}[1][i]{\tau_{#1}}
\newcommand{\adtimes}{\mc T_{\trm{adoption}}}
\newcommand{\neverad}{\mc I_{\trm{never-ad}}}
\label{sec : staggered adoption}
Staggered adoption is a recurring intervention assignment pattern in policy-evaluation applications~\cite{ben2022synthetic,athey2022design}---its key characteristic is that a unit remains treated throughout once it receives treatment at its adoption time. Previous works on staggered adoption setup aim to impute the mean outcome of non-treated units when adoption times are completely random~\cite{athey2022design} or fixed~\cite{ben2022synthetic}. The work of \cite{gunsilius2023distributional} consider distributional recovery of univariate outcomes in the synthetic control setup~(a special case of staggered adoption where a single unit is treated at a fixed adoption time).
We show here that $\kerNN$ recovers individual distribution treatment effect for multivariate outcomes, when the adoption times depend on unobserved variables. 

We introduce a special version of our observational model \cref{mod:potential-outcome} that exemplifies a staggered adoption scenario. We use integer adoption times $\adtime[j]\geq1$, with treatment beginning at time $\adtime[j]+1$ and continuing thereafter; values $\adtime[j] \geq T$ are allowed for never-adopters. For unit $j$ with adoption time $\adtime[j]$, set the missingness $A_{j, s} = \indic{s > \tau_j}$ and consider the following observational model
\begin{align}
     \qtext{for} j \in [N], s \in [T]:\quad Z_{j, s} \defeq \begin{cases}
        X_{1}^{(1)}(j, s), \ldots, X_{n}^{(1)}(j, s) \sim \mu_{i, t}^{(1)}&\qtext{if} s > \adtime[j], \\
        X_{1}^{(0)}(j, s), \ldots, X_{n}^{(0)}(j, s) \sim \mu_{i, t}^{(0)}&\qtext{if} s \leq \adtime[j].
    \end{cases}
    \label{mod:potential-outcome staggered}
\end{align}
We consider here a staggered adoption model with confounded adoption times~\cite{athey2022design}. Each adoption time $\adtime[j]$ determines the values of the row missingness $\sbraces{ A_{j, s} }_{s \in [T]}$, and we allow adoption times $\adtime[j]$ to be confounded by latent factors, which is specified below.
\begin{assumption}[Staggered adoption with unobserved confounding]
\label{assump : confounded stagger}
    The distribution of adoption times $\adtimes \defeq (\adtime[1], \ldots, \adtime[N])$ can depend on the latent factor $\mc U$, and $\adtimes$ is independent of $\mc V$ conditioned on $\mc U$.
\end{assumption}
\cref{assump : confounded stagger} can be thought of as analogous to \cref{assump : unobs confounding}. In staggered adoption settings, it is common~(and typically necessary) to assume that there exists a subset of units that are \emph{never adopters}, where $ j \in \neverad \defeq \sbraces{i \in [N]: \adtime[i] \geq T}$, i.e., $A_{j, s} = 0$ for all $s \in [T]$. See \cref{fig:missingness} for an example of the typical induced sparsity pattern in the staggered adoption setting (where \cref{assump : confounded stagger} holds). 

We now present an instance based error bound of $\kerNN$ estimate under the setup \cref{mod:potential-outcome staggered}. As discussed in \cref{sec:algo}, to prove this bound, we use the more general version of $\kerNN$ presented in \cref{sec : fully general kerNN}. Without loss of generality, we assume that the first unit is under treatment at time $T$~(i.e. $\tau_1 < T$) and state our result for estimating that unit's counterfactual outcome under control at that time. Refer to \cref{proof : staggered adoption} for a proof of the following result. 
\newcommand{\lnbhdstag}{\mbf N_{1, \eta, \mc T}^\star}
\begin{theorem}[\tbf{Staggered adoption guarantee}]\label{thm:stagger-bound}
    Suppose the controlled measurements and missingness of \cref{mod:potential-outcome staggered} respect \cref{assump:factorization,assump:latent-independence,assump : measurement generation,assump : confounded stagger}. Then for any $\eta,\delta>0$, estimator $\what{\mu}_{1, T, \eta}^{(0)}$ of $\kerNN$ satisfies
    \begin{align}
        \E \brackets{\| \what{\mu}^{(0)}_{1, T, \eta} - \mu_{1, T}^{(0)}\|_\kernel^2 \big| \mc U, \adtimes}
        \leq \eta+ \sinfnorm{\kernel}\biggbrackets{\frac{c_0 \log(2N/\delta) }{\sqrt{\adtime[1]  }} +  \frac{4 (\log n + 1.5)}{ n |\nbhdstar| } + 4\delta }   ,
        \label{eq:most-raw-guarantee stagger}
    \end{align}
        where $\nbhdstar \defeq \sbraces{j \in \neverad : \gio < \eta - \frac{c_0\| \kernel\|_\infty \log(2N/\delta) }{\sqrt{\adtime[1]  }}}$, and the constant $c_0$ and expectation are as in \cref{prop:most-raw-bound}. 
\end{theorem}

Inequality \cref{eq:most-raw-guarantee stagger} is an instance dependent guarantee on counterfactual distributional recovery under the staggered adoption set-up~\cite{abadie2010synthetic, athey2022design, ben2022synthetic, gunsilius2023distributional} where the adoption times are confounded by unobserved factors. We refer the reader to the discussion following \cref{prop:most-raw-bound} for a more detailed review on the role of the instance dependent bound for devising a fast optimization procedure for choosing $\eta$.
The first two terms in the RHS of \cref{eq:most-raw-guarantee stagger} are akin to the bias of $\kerNN$, where longer adoption time $\tau_1$ contributes to a more precise row metric estimate~(see \cref{prop:most-raw-bound}), yielding low bias. The third term in the RHS of \cref{eq:most-raw-guarantee stagger} is akin to the variance component, where more never adopters enlarges the neighborhood of $\kerNN$ that is averaged upon.

We now refine the guarantees when additional structural assumptions on the operator $g$ and the distribution of latent factors in \cref{assump:factorization} are given. The following result provides guarantees of $\kerNN$ with respect to the fully integrated $\mmd$ metric, so the guarantees are not data-dependent but they reveal how $\kerNN$ explicitly depend on the model parameters. We refer the reader to \cref{proof:stagger_complete} for the proof.

\begin{corollary}[\tbf{Guarantees for specific examples under staggered adoption}]
\label{cor:stagger-bound}
Consider the control measurements and missingness of \cref{mod:potential-outcome staggered} under the assumptions of \cref{thm:stagger-bound}, with measurement support $\mc X=[-1,1]^d$. Let $\neverad\subset[N]\setminus\{1\}$ be a fixed set of $N^\alpha$ indices, chosen before the unit latent factors are drawn. Suppose $\tau_j\in[T^\beta,T]$ for $j\notin\neverad$ and $\tau_1<T$, where $\alpha,\beta\in(0,1)$. Assume the parameter conditions \cref{eq:stagger_complete_parameters,eq:stagger_complete_coverage}, with the values of $r,L,\sinfnorm{\kernel}$ specified there for the corresponding example. The choice of $\eta^\star$ is given in \cref{eq:stagger_complete_radius}.
\begin{enumerate}[label=(\alph*), leftmargin=*]
\item Under \cref{exam:gaussian_family}, suppose the unit and outcome latent factors are sampled independently and uniformly from $[-1,1]^2$. Then
\begin{align}
\E\brackets{\knorm{\what\mu^{(0)}_{1,T,\eta^\star}-\mu^{(0)}_{1,T}}^2}
\leq \tilde O\biggbrackets{\frac{d^2}{\sqrt{nN^\alpha}}+\frac{d^2}{\sqrt{T^\beta}}}.
\label{eqn : ex1 mmd bound}
\end{align}
\item Under \cref{exam:infinite_family}, suppose the unit and outcome latent factors are sampled independently and uniformly from $[-1,1]^r$. Let $L=(\sum_{\ell\geq1}L_\ell^2)^{1/2}\in(0,\infty)$, where $L_\ell$ are the Lipschitz constants of the coefficient functions $\alpha_\ell$. Then
\begin{align}
\E\brackets{\knorm{\what\mu^{(0)}_{1,T,\eta^\star}-\mu^{(0)}_{1,T}}^2}
\leq \tilde O\biggbrackets{\left(\frac{L^r}{nN^\alpha}\right)^{\frac{2}{r+2}}+\frac{1}{\sqrt{T^\beta}}}.
\label{eqn : ex2 mmd bound}
\end{align}
\end{enumerate}
\end{corollary}

The guarantees \cref{eqn : ex1 mmd bound,eqn : ex2 mmd bound} in \cref{cor:stagger-bound} are on the fully integrated $\mmd$ metric; it loses granularity compared to the instance-based guarantees, but it present how the model parameters interact. We make several remarks regarding these parameters. 

Note that $\eta$ is a stand alone additive term~(hence a dominant one) that characterizes the error bound in the instance-based bounds \cref{prop:most-raw-bound,thm:stagger-bound}; this highlights the significance of choosing an appropriate value for $\eta$ as it is the dominant term that characterizes how fast $\kerNN$ recovers the distribution. The choice of $\eta^\star$ in \cref{cor:stagger-bound} balances the leading bias and variance terms while accounting for the distance estimation error.

Second, the dimension $r$ of latent factors~(see \cref{assump:factorization}) governs the rate of convergence of $\kerNN$, hence serving as the effective dimension, while the dimension $d$ of measurements appears only as a scaling constant\footnote{The lipschitz constant $L$ for item (b) of \cref{cor:stagger-bound} can be expressed as a function of $d$ when more assumptions are given on $g$, but we do not go into further details.}. In \cref{cor:stagger-bound}, $\alpha$ controls the never-adopter group size $N^\alpha$, while $\beta$ controls the minimum overlap $T^\beta$ with the target unit. Referring to the discussion following \cref{prop:most-raw-bound}, the precision of the row-metric \cref{eq:row-metric} is one source of bias and $\beta$ controls this degree of precision. The variance of $\kerNN$ depends on the number of effective sample size which amounts to the number of total measurements within the observed neighborhood $\mbf N_{1, \eta}$, and parameter $\alpha$ controls the size of the neighborhood. 

\paragraph{Distributional treatment effect for staggered adoption}

We leverage \cref{cor:stagger-bound} to provide guarantees of an estimator that learns the kernel treatment effect~(see \cref{sec : estimand}) in the staggered adoption scenario. The causal estimand here is $\dte_{1, T} = \snorm{ \mu_{1, T}^{(1)} - \mu_{1, T}^{(0)}  }_\kernel$. For hyper-parameters $\eta = (\eta_0, \eta_1)$, we propose an estimator $\what{\dte}_{1, T, \eta} = \snorm{ \what\mu_{1, T, \eta_1}^{(1)} -  \what{\mu}_{1, T, \eta_0}^{(0)} }_\kernel$, formally defined in \cref{app:dte-proof} --- it is the normed difference of the output of $\kerNN$ applied on two different set of outcomes, $X_{1:n}^{(1)}(i, t)$ and $X_{1:n}^{(0)}(i, t)$. The following result presents a guarantee of $\what{\dte}_{1, T, \eta}$ under the staggered adoption setting on the potential outcome model \cref{mod:potential-outcome staggered} that is specified in \cref{app:dte-proof}. A notable feature of the data generating process in \cref{app:dte-proof} is that (i) the embeddings $\mu_{i, t}^{(0)}\kernel$ and $\mu_{i, t}^{(1)}\kernel$ are factored according to \cref{assump:factorization} and they share the row latent factors $\mc U$ and (ii) assignment pattern is according to \cref{assump : confounded stagger}.

\begin{corollary}[$\dte$ guarantee under staggered adoption]\label{cor:dte-stagger}
Suppose the data generating process specified in \cref{app:dte-proof}, with both potential-outcome families satisfying \cref{exam:gaussian_family} and the unit and outcome latent factors sampled independently and uniformly from $[-1,1]^2$, sharing $\mc U$ across the two arms. Let $\neverad\subset[N]\setminus\{1\}$ be a fixed set of $N^\alpha$ indices, chosen before the unit latent factors are drawn. Suppose $\adtime[j]\in[T^\beta,T^{1-\beta}]$ for $j\notin\neverad$, where $\alpha\in(0,1)$ and $\beta\in(0,1/2)$ are fixed. Assume the parameter and coverage conditions in \cref{proof:dte-stagger}. Then, for the choices $\eta^\star=(\eta_0^\star,\eta_1^\star)$ specified there,
\begin{align}
    \E \brackets{ ( \what{\dte}_{1, T, \eta^\star} - \dte_{1, T} )^2 } \leq \tilde{O}\biggbrackets{ \frac{d^2}{\sqrt{n \cdot N^{(1 - \alpha) \wedge \alpha}}} + \frac{d^2}{\sqrt{ T^{(1 - \beta) \wedge \beta}}}}.
\end{align}
\end{corollary}

\subsection{Distributional recovery under a propensity model for missingness}\label{sec : main instance non-positivity}

In this section, we express our guarantee of $\kerNN$ using the propensities $p_{i, t} = \Prob(A_{i, t} = 1|\mc U)$, where the dependence of $p_{i, t}$ on the latent factors $\mc U$ indicates there exists unobserved confounding. The non-positivity of missingness $\mc A$ formally means there exist some entry $(i, t)$ such that its propensity assume value zero~($p_{i, t} = 0$), hence the entry is never observed. So a guarantee of $\kerNN$ expressed via propensities, unlike that of \cref{prop:most-raw-bound}, conveniently reflect how non-positivity of missingness plays a role on the performance of $\kerNN$. We introduce the following assumption regarding randomness of $\mc A$.

\begin{assumption}[Conditional independence in missingness] 
\label{assump : cond_ind_a}
Conditioned on the row factors $\mc U $, the $A_{i, t}$'s are drawn independently across $i$ and $t$ with mean $\Prob( A_{i, t} = 1 | \mc U ) = p_{i, t}$\footnote{We omit in notation the dependence of propensity to the latent factors.}.

\end{assumption}
The conditional independence of $\mc A$ assumed in \cref{assump : cond_ind_a} simplifies discussion and analysis, but we expect our analysis to be valid with slight modification even when \cref{assump : cond_ind_a} is generalized to some appropriate conditional mixing conditions~(so that it allows concentration of measurement's average). Further, we note that conditional independence does not necessarily imply marginal independence across missingness.

We introduce some shorthands
\begin{align}
    \unbhdstarp &\defeq \braces{ j \neq 1 : \gio
    < \eta  + \errjp }
    \qtext{and}
     \lnbhdstarp \defeq \braces{ j \neq 1 : \gio < \eta - \errjp},
     \label{eq:pop_neighbor_p}
\end{align}
where $\errjp \defeq \frac{\sqrt{2}c_0 \| \kernel\|_\infty \sqrt{\log(2N /\delta)}}{ \sqrt{\sum_{s \neq 1} p_{1,s}p_{j, s} }}$, with $\errjp=\infty$ when the denominator is zero; note that all the shorthands depend up to the latent factors $\mc U$.
The following result presents the guarantee of $\kerNN$ expressed via propensities, and its proof can be found in \cref{proof:thm_propensity}.

\begin{theorem}[\tbf{Propensity-based guarantee}]
\label{thm:prop-bound}Suppose observed measurements and missingness from model \cref{model : dist matrix completion} respect \cref{assump:factorization,assump:latent-independence,assump : unobs confounding,assump : measurement generation,assump : cond_ind_a}. Fix $\delta\in(0,1)$. For large enough $\eta > 0$ and for appropriate choices of $\mc U$ as specified by the overlap and neighbor-mass conditions in \cref{event:good3}, estimator $\what{\mu}_{1, 1, \eta}$ of $\kerNN$ satisfies
\begin{align}
        \E\brackets{ \|\what{\mu}_{1, 1, \eta} - \mu_{1, 1}\|_\kernel^2  | \mc U} 
        \leq \eta + \sinfnorm{\kernel}\biggbrackets{\max_{\substack{j \in \overline{\mbf N}_{1, \eta, p}^\star}} \frac{\sqrt{2}c_0 \sqrt{\log(2N /\delta)}}{ \sqrt{\sum_{s \neq 1} p_{1,s}p_{j, s} }}+ \frac{(8\log n + 12)}{n \sum_{j \in \underline{\mbf N}_{1, \eta, p}^\star } p_{j, 1} } + \mbf r_\delta }
        \label{eq:prop-bound}
\end{align}
where the term $\mbf r_\delta = 4\delta + 8N \exp\bigbraces{ -\frac{1}{8}\min_{\substack{j \neq 1\\p_{j,1}>0}}\sum_{s \neq 1}p_{1, s}p_{j, s} } + 8\exp\bigbraces{ -\frac{1}{8} \sum_{j \in \lnbhdstarp}p_{j, 1} }$. 
\end{theorem}

The non-positivity condition of various missingness patterns can be conveniently plugged into the propensity dependent bound \cref{eq:prop-bound} so as to derive guarantees of $\kerNN$. We illustrate our point through an example. For fixed latent factors $\mc U$ and given some constants $\alpha, \beta \in (0, 1], \underline c > 0$, suppose (i) at most $(1 - \beta)$ proportion of $T/2$ number of columns\footnote{To be precise, here we mean $(1 - \beta)$ proportion of the log-transform of $T/2$. So  at most $(1 - \beta)\log(T/2)$ entries out of $\log (T/2)$ entries are never observed.} are never observed for all rows $j\in [N]$, (ii) at most $(1 - \alpha)$ proportion of first column entries are never observed and (iii) all other entries have propensity that is lower bounded by some constant $\underline c > 0$. These three conditions, which collectively indicate potential violation of positivity, yield the following 
\begin{align}\label{eq:prop-lower-bound}
    \sum_{s \neq 1} p_{1, s}p_{j, s} \geq \underline c T^{\beta} \qtext{for all $j \neq 1$ and} \sum_{j \in \lnbhdstarp}p_{j, 1} \geq \underline c \cdot |\sbraces{j \neq 1: j \in \lnbhdstarp,\ p_{j, 1} \geq \underline c}|,
\end{align}
where the second inequality in the above display implicitly depends on the parameter $\alpha$. Plugging the condition \cref{eq:prop-lower-bound} into our propensity based guarantee \cref{eq:prop-bound} immediately induces
\begin{align}
    \E\brackets{ \|\what{\mu}_{1, 1, \eta} - \mu_{1, 1}\|_\kernel^2  | \mc U} 
        \leq \eta + \tilde O\biggbrackets{ \frac{1}{|\sbraces{j\neq 1:j \in \lnbhdstarp,\ p_{j, 1} \geq \underline{c}}|} + \frac{1}{\sqrt{T^\beta}} }.
\end{align}

Notably, with minor adjustments, the staggered adoption missingness specified in \cref{cor:stagger-bound} can be recovered from the condition \cref{eq:prop-lower-bound} by setting $\underline c = 1$. 
Hence $\kerNN$ can go beyond the confounded staggered adoption setting specified in \cref{assump : confounded stagger} and account for general MNAR missingness that also violates positivity. 

As a special case, we present a result on the missing completely at random~(MCAR) scenario, the most widely studied missingness pattern in the matrix completion literature~\cite{chatterjee2015matrix,li2019nearest}. MCAR is characterized as the missingness $\mc A$ that are exogenous~(i.e. independent to all other randomness), i.i.d generated with propensities $p_{i, t} = p$ for all $i\in [N]$ and $t\in [T]$. Observe that MCAR is a special case of \cref{eq:prop-lower-bound}, where $\alpha = 0, \beta = 0$ and $\underline c = p$.We refer to \cref{proof:positivity_complete} for a proof of the following result.

\begin{corollary}[\tbf{Guarantees for specific examples under MCAR}]
\label{cor:mcar-bound}
Consider measurements from \cref{model : dist matrix completion} satisfying \cref{assump : measurement generation}, with measurement support $\mc X=[-1,1]^d$. Let $A_{j,s}$ be independent Bernoulli$(p)$ variables, independent of the latent factors and measurements. Assume the parameter conditions \cref{eq:mcar_complete_parameters,eq:mcar_complete_coverage}, with the values of $r,L,\sinfnorm{\kernel}$ specified there for the corresponding example. The choice of $\eta^\star$ is given in \cref{eq:mcar_complete_radius}.
\begin{enumerate}[label=(\alph*), leftmargin=*]
\item Under \cref{exam:gaussian_family}, suppose the unit and outcome latent factors are sampled independently and uniformly from $[-1,1]^2$. Then
\begin{align}
\E\brackets{\knorm{\what\mu_{1,1,\eta^\star}-\mu_{1,1}}^2}
\leq \tilde O\biggbrackets{\frac{d^2}{\sqrt{npN}}+\frac{d^2}{p\sqrt T}}.
\nonumber
\end{align}
\item Under \cref{exam:infinite_family}, suppose the unit and outcome latent factors are sampled independently and uniformly from $[-1,1]^r$. Let $L=(\sum_{\ell\geq1}L_\ell^2)^{1/2}\in(0,\infty)$, where $L_\ell$ are the Lipschitz constants of the coefficient functions $\alpha_\ell$. Then
\begin{align}
\E\brackets{\knorm{\what\mu_{1,1,\eta^\star}-\mu_{1,1}}^2}
\leq \tilde O\biggbrackets{\left(\frac{L^r}{npN}\right)^{\frac{2}{r+2}}+\frac{1}{p\sqrt T}}.
\nonumber
\end{align}
\end{enumerate}
\end{corollary}

We refer the reader to the discussion that follows \cref{cor:stagger-bound} for a detailed explanation on how the model parameters interact in \cref{cor:mcar-bound}. The conditions in \cref{eq:mcar_complete_parameters,eq:mcar_complete_coverage} ensure sufficient overlap for row comparisons, sufficiently many nearby observed units, and an effective radius within the range of \cref{lem:stagger_complete_count}.

\begin{remark}[Dependent column factors]
Both staggered adoption in \cref{assump : confounded stagger} and the HeartSteps application involve temporally indexed data, where adjacent column factors may be dependent. For the infinite-dimensional family in \cref{exam:infinite_family}, \cref{sec:dependent-column-appendix} extends the staggered adoption and MCAR guarantees to stationary $\beta$-mixing column factors using the same estimator. Under the additional conditions stated there, the first terms of \cref{cor:stagger-bound} (b) and \cref{cor:mcar-bound} (b) remain unchanged, while the row metric estimation terms are multiplied by $\sqrt{b_\delta}$, where $b_\delta$ is the mixing block length defined in \cref{eq:noniid-mixing-block-length}. 

\end{remark}

\subsection{Rate optimality}
\label{subsec:noniid-minimax}

We identify conditions under which \kerNN, with a specific choice of radius, attains the minimax rate up to a factor of $\log n$ for estimating the distribution $\mu_{1,1}$. The minimax lower bound has the same order as the first term in \cref{cor:mcar-bound} (b).

Let $\mc M_{r,L}$ denote the class of distribution valued maps $(u,v)\mapsto\mu(u,v)$ satisfying the MCAR model of \cref{cor:mcar-bound} (b), with $\sum_{\ell\geq1}L_\ell^2\leq L^2$, conditional on $A_{1,1}=0$; see \cref{eq:full-minimax-model-class} for the formal definition. For each $\mu\in\mc M_{r,L}$, write $\Prob_\mu$ and $\E_\mu$ for probability and expectation under the induced joint law, with target distribution $\mu_{1,1}=\mu(u_1,v_1)$. Fix $r>4$, $d\geq1$, and $n,N,T\geq2$, with $L>0$ and $p\in(0,1)$, and consider the window
\begin{align}
    \frac{128\log(2N)}{p^2a_\star^2}\leq T\leq npNa_\star \qtext{where} a_\star\defeq\left(\frac{L^r}{npN}\right)^{2/(r+2)}.
    \label{eq:full-minimax-joint-window}
\end{align}
The lower limit on $T$ controls the error in estimating row similarities; it ensures that such error, of order $\sqrt{\log N}/(p\sqrt T)$, is bounded by the order of $a_\star$. 
The upper limit of \cref{eq:full-minimax-joint-window} serves a purpose in the lower-bound construction; it keeps the number of columns small enough that with probability bounded away from zero, no other latent column factor is close to that of the target entry. On this event, measurements in the other columns cannot distinguish the alternatives used in the proof, so the target column smoothing error remains unavoidable even for estimators using the full observed matrix.

\begin{theorem}[\tbf{Minimax optimality up to logarithmic factor under MCAR}]
\label{thm:full-matrix-minimax-optimality}
Under \cref{eq:full-minimax-joint-window} and the sample-size conditions \cref{eq:full-minimax-joint-sample-conditions} in \cref{sec:full-matrix-minimax}, there exist $c,C>0$ depending only on $r$ such that, with the radius $\eta^\star$ in \cref{eq:full-minimax-common-radius},
\begin{align}
    ca_\star \leq\inf_{\what\mu} \sup_{\mu\in\mc M_{r,L}} \E_\mu\brackets{\|\what\mu-\mu_{1,1}\|_\kernel^2} \leq \sup_{\mu \in \mc M_{r, L}}\E_\mu \brackets{\| \what \mu_{1, 1, \eta^\star} - \mu_{1, 1} \|_\kernel^2} \leq Ca_\star \log n.
    \label{eq:full-minimax-main-matching}
\end{align}
The infimum ranges over measurable estimators using observed measurements and missingness indicators. 
\end{theorem}

For fixed $L$, the restriction $r>4$ is necessary for the window \cref{eq:full-minimax-joint-window} to remain nonempty as $N\to\infty$. For example, fix $n,p,L$ and take $T=\lfloor N^\gamma\rfloor$ with $4/(r+2)<\gamma<r/(r+2)$. Then both the window and the additional sample-size conditions in \cref{eq:full-minimax-joint-sample-conditions} hold for all sufficiently large $N$.

The proof in \cref{subsec:full-minimax-optimality-proof} combines the minimax lower bound in \cref{thm:full-matrix-oracle-minimax} with a uniform refinement of \cref{cor:mcar-bound} (b) over a class $\mc M_{r,L}$. Optimality outside this regime, including the sharpness of the $1/(p\sqrt T)$ term, remains open. For larger $T$, nearby columns may provide additional observations. Two-sided smoothing attains the latent-factor oracle rate in scalar matrix completion under suitable H\"older regularity and sample-size conditions~\cite{sadhukhan2024twosided}. Extending this approach to distributional estimation under the $\mmd$ loss is left for future work.

\subsection{Proof strategy of \cref{prop:most-raw-bound}}

Here we briefly discuss the proof strategy of \cref{prop:most-raw-bound}; see \cref{sec:proof_of_instance_based} for details. Notably, a more granular $\mmd$ error $ \E \brackets{\| \what{\mu}_{1, 1, \eta} \!-\! \mu_{1, 1}\|_\kernel^2 \vert \wholerand}$ is first bounded, where $\mc V_{-1} = \{v_2, ..., v_T\}$ and $\mc D_{-1}$ refers to all the measurements excluding those in the first column (say $\mc D_1$) of the matrix. Then integrating the granular error and its error bound over $\mc V_{-1}$ and $\mc D_{-1}$ yields the desired result. 

A fully stochastic analysis of the squared $\mmd$ error $\| \what{\mu}_{1, 1, \eta} \!-\! \mu_{1, 1}\|_\kernel^2$ without marginalization is difficult as $\kerNN$ is a two step procedure where the second~(averaging) step depends heavily on the random neighbor constructed in the first step. Specifically, the randomness of $\kerNN$ is characterized by the stochastic neighborhood $\mbf N_{1, \eta}$~(driven by randomness $\mc V_{-1}, \mc D_{-1}, \mc U, \mc A $) and the measurements therein that are averaged upon~(driven by randomness $v_1$, $\mc D_1$). 
Conditioning on the randomness of the neighborhood $\mbf N_{1, \eta}$ fixes the membership of the measurements, but as a result the joint distribution of those measurements within the neighborhood becomes unclear. So when conditioned upon the neighborhood, any concentration type results or Gaussian approximations~(e.g., Yurinskii coupling, see \cite{pollard2002user}) cannot be applied on the average of the measurements in the neighborhood. Instead, we marginalize over the measurements in the neighborhood (again driven by $v_1$ and $\mc D_{1}$) and deal with the remaining stochasticity of the neighborhood $\mbf N_{1, \eta}$.

The difference of the embeddings $\what \mu_{1, 1, \eta}\kernel - \mu_{1, 1}\kernel$ is decomposed into bias and variance~(see \cref{eq:embedding-decomp}): 
\begin{align}
   b(j, 1) = \mu_{j, 1}\kernel - \mu_{1, 1}\kernel \qtext{and} v_n(j, 1) = \empdistZ{j, 1}\kernel - \mu_{j, 1}\kernel \qtext{for $j \in \mbf N_{1, \eta}$.} 
\end{align}
So bounding the marginalized bias $\E[\|b(j, 1)\|_{\kernel}^2|u_1, u_j]$ and the marginalized variance $\E[\| v_n(j, 1)\|_\kernel^2|u_j]$ is sufficient to bound $\E\big[ \| \what{\mu}_{1, 1, \eta} - \mu_{1, 1} \|_\kernel^2 | \mc V_{-1}, \mc D_{-1}, \mc U, \mc A \big]$; note that $b(j, 1)$ is a function of $u_1, u_j, v_1$ and $v_n(j, 1)$ a function of $u_j, v_1$ and the measurements at the $(j, 1)$th entry. 

To be more specific, the marginalized error $\E[\| \what \mu_{1, 1, \eta} - \mu_{1, 1}\|_\kernel^2 | \mc V_{-1}, \mc D_{-1}, \mc U, \mc A]$ is bounded by the two terms~(see \cref{lem : MMD decomp}),
\begin{align}\label{eq:bias-var-bound}
    \max_{j \in \mbf N_{1, \eta}}A_{j, 1}\cdot \E\big[ \| b(j, 1)\|_\kernel^2 | u_1, u_j \big] , \ \ \ \ \frac{1}{(\sum_{j \in \mbf N_{1, \eta}}A_{j, 1})^2}\sum_{j \in \mbf N_{1, \eta}} A_{j, 1}\cdot \E\big[ \| v_n(j, 1)\|_\kernel^2|u_j \big].
\end{align}
The row metrics $\rho_{j, 1}$ are uniformly~(across $j$) concentrated around $\E[ \|b(j, 1)\|_\kernel^2|u_1, u_j ]$~(see \cref{lem:conditional-concentration}) whenever there is large overlap of observed entries across rows; i.e., when $\errja$ defined in \cref{eq:err_defn} is small. From the identity $\E\big[ \| b(j, 1)\|_\kernel^2 | u_1, u_j \big] = \gio$, we observe that the uniform concentration yields a sandwiched inclusion of $\mbf N_{1, \eta}$ between the two neighborhoods $\lnbhdstarA$ and $\unbhdstarA$ defined in \cref{def : pop nbhd data centered}. 

Using the inclusion $\mbf N_{1, \eta} \subseteq \unbhdstarA$, the first term in the above display \cref{eq:bias-var-bound} is bounded by $\eta + \max_{j \in \unbhdstarA} \errja$. The variance component $\E\big[ \|v_n(j, 1)\|_\kernel^2|u_j \big]$ in the above display \cref{eq:bias-var-bound} is bounded using the CLT for kernel mean embeddings~\cite[Thm.~3.4]{muandet2017kernel}, yielding $\E\big[ \|v_n(j, 1)\|_\kernel^2|u_j \big] \leq \tilde O(n^{-1})$ for all $j$~(see \cref{eq:mmd_bound}). We then invoke the inclusion $\lnbhdstarA \subseteq \mbf N_{1, \eta}$ derived from the uniform concentration of row metrics $\rho_{j, 1}$ around $\gio$, to bound the second term in \cref{eq:bias-var-bound} by $\tilde O\big(1/(n\sum_{j \in \lnbhdstarA}A_{j, 1})\big)$.

\begin{figure}[!t]%
    \centering
    \includegraphics[width=0.95\linewidth]{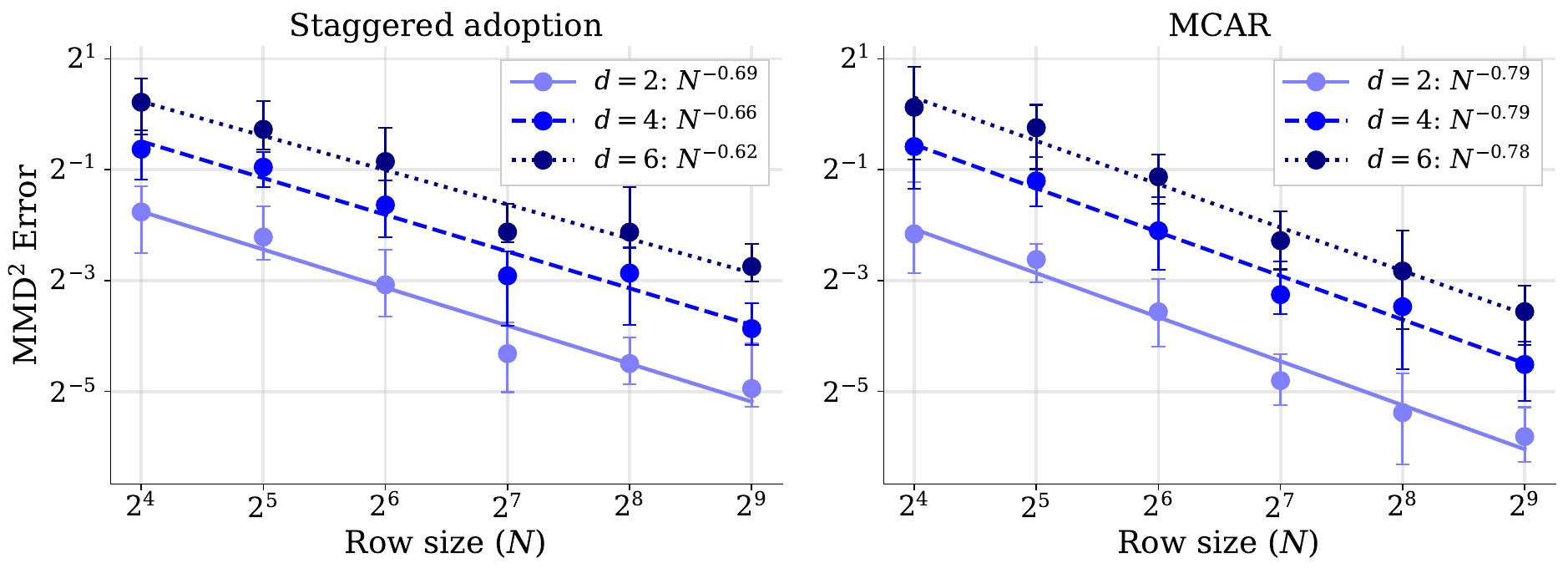}
    \caption{\tbf{Squared $\mmd$ error of cross-validated \kerNN by dimension $d$ and missing pattern.} The left panel depicts the squared-$\mmd$ error decay of $\kerNN$ as $N$ increases for different measurement dimensions $d$ under staggered-adoption missingness~(see panel (a) of \cref{fig:missingness}); the right panel gives the analogous MCAR result. Medians over 50 paired replications with 95\% bootstrap intervals are presented, and lines are log--log regression fits whose slopes are given in the legends.}%
    \label{fig:sim_mmdbyN}%
\end{figure}

\section{Experiments}
\label{sec:sim}

This section implements synthetic and real data experiments. We also introduce computationally efficient methods for selecting the radius $\eta$. We compare the standard cross validation~(CV in short, with radius $\eta=\etacv$, see \cref{app:experiment-protocol}) with two efficient methods: Direct, which chooses $\eta=\etadirect$ by minimizing the data-driven objective motivated by the instance-dependent squared-$\mmd$ bound in \cref{prop:most-raw-bound} and Entry-KT, that thins observed $n$ measurements per entry before applying CV. The simulations show error decay consistent with the theory and illustrate how distributional information can improve mean estimation. Under staggered adoption, Direct achieves empirical error decay comparable to CV with lower runtime. Entry-KT retains comparable accuracy using substantially fewer measurements for tuning.

\subsection{Simulation study}
Each cell distribution is Gaussian, with mean and covariance determined by row and column latent factors of dimension $r=2$ and all panels have $T=80$ columns. We study two missingness patterns. Under MCAR, each cell is observed independently with probability $p=0.5$, and we estimate the distribution of a missing cell. Under confounded staggered adoption~(MNAR), adoption times depend on the row latent factors; we treat post-adoption cells as missing and estimate, at time $T$, the control distribution of a unit that has already adopted, as in \cref{thm:stagger-bound}. Errors are computed against the true distribution of the target cell. The data-generating process and adoption mechanism are specified in \cref{app:synthetic-designs}.

\subsubsection{Alignment of theory and empirical results}
\cref{fig:sim_mmdbyN} presents the squared-$\mmd$ error of cross-validated \kerNN as the row size $N$ increases and the measurement dimension $d$ varies. We use the degree two polynomial kernel for both the algorithm and evaluation. The empirical results reflect the theoretical insights derived from \cref{cor:stagger-bound,cor:mcar-bound}. For both missingness patterns, the squared-$\mmd$ error decreases with $N$ at broadly similar fitted rates across the three values of $d$. This aligns with the fact that the convergence exponent of $\kerNN$ is governed by the latent dimension $r$~(see discussion following \cref{cor:stagger-bound,cor:mcar-bound}); here the simulated data has fixed latent dimension $r = 2$.
Further, increasing the measurement dimension $d$ raises the squared-$\mmd$ error while preserving the same qualitative decay with $N$. This again aligns with the theory, where $d$ contributes to the scale of the error bound.

\begin{figure}[!t]%
    \centering
    \includegraphics[width=0.95\linewidth]{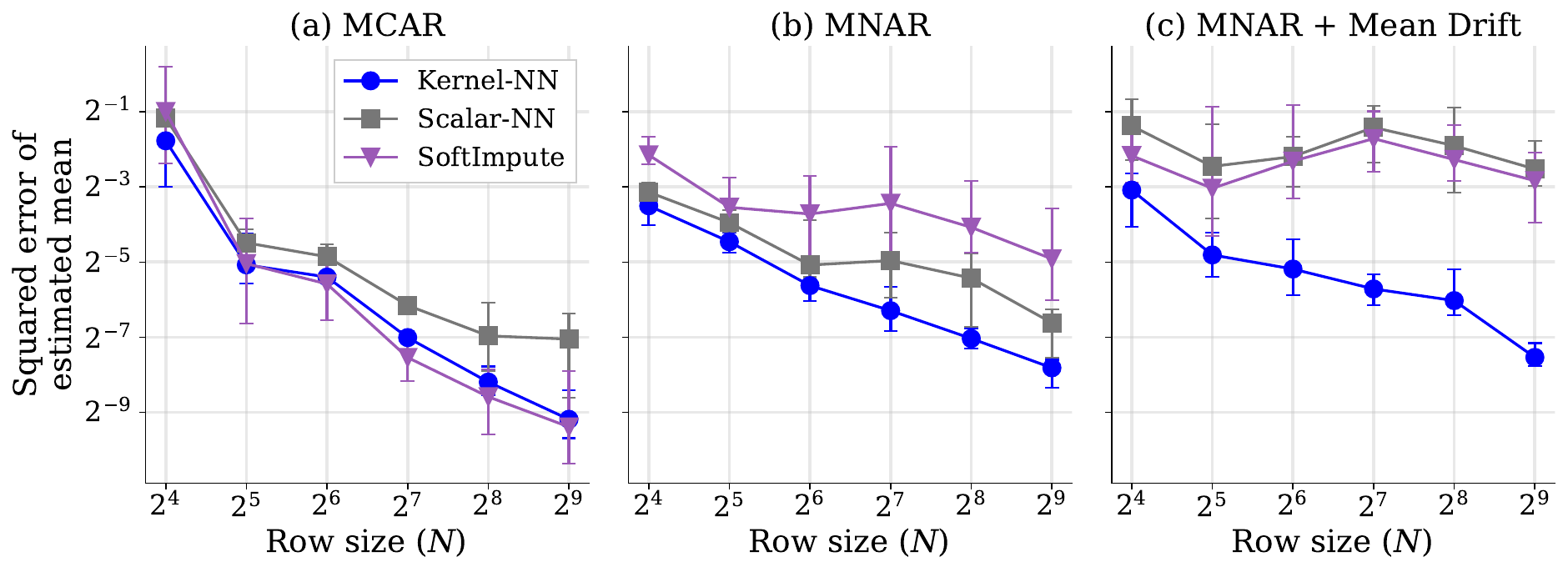}
    \caption{\tbf{Mean estimation with \kerNN, Scalar-NN, and SoftImpute.} Points show median squared error; bars show 95\% bootstrap intervals. Measurements are Gaussian, with means and variances linked through a common row factor. (a) Missing completely at random (MCAR). (b) Confounded staggered adoption (MNAR). (c) MNAR with means that drift apart late in the panel.}%
    \label{fig:feature-kernel-choice}%
\end{figure}

\subsubsection{Can distributional information improve mean estimation?}

\cref{fig:feature-kernel-choice} compares three methods for estimating the mean of distributions, evaluated by squared Euclidean error. We use the exponential kernel for the \kerNN algorithm and take the mean of its estimate. Scalar-NN applies the scalar nearest-neighbor procedure in \cref{eq:snn-distance,eq:snn-average} separately to each coordinate's observed sample means. SoftImpute~\cite{mazumder2010softimpute} is applied to the matrix of observed sample means. The designs and tuning procedures are specified in \cref{app:mean-imputation}.

The three panels of \cref{fig:feature-kernel-choice} are increasingly difficult settings. Under MCAR~(panel (a)), the errors of all three methods decrease with $N$, with Scalar-NN the least accurate at large $N$. Under confounded staggered adoption~(panel (b)), the two nearest-neighbor methods continue to improve, whereas error of SoftImpute decreases slowly and is the largest at every $N$. In panel (c), the units' means drift apart late in the panel, and \kerNN is the only method whose error still decreases with $N$; at $N=512$ its median error is more than $25$ times smaller than that of either baseline.

In panel (b), adoption depends on the unit latent factors, so the entries are missing not at random. SoftImpute, which fits all entries jointly, is not robust to such missingness, whereas both nearest-neighbor methods are, in line with \cref{thm:stagger-bound}. In panel (c), the unit means are nearly identical before adoption, so Scalar-NN and SoftImpute, which use only the means, cannot identify similar units. The variances are driven by the same latent factor as the means and still differ across units, so \kerNN, which compares distributions, continues to find similar units.

\subsubsection{Independent and dependent columns}
\label{subsec:dependent-columns}
\begin{figure}[tb]%
    \centering
    \includegraphics[width=0.95\linewidth]{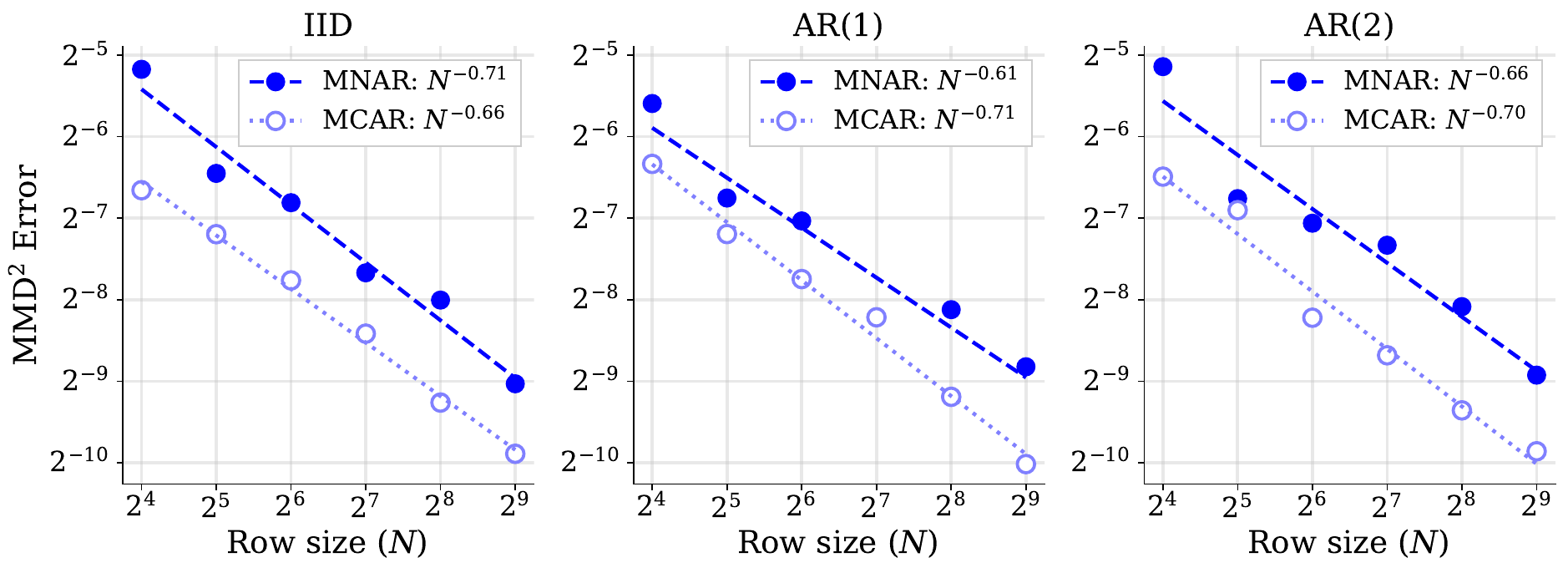}
    \caption{\tbf{Cross-validated \kerNN with independent and dependent columns under MNAR and MCAR.} Each panel overlays the confounded staggered-adoption design~(MNAR, filled markers and dashed fits) and MCAR~(hollow markers and dotted fits) for $\iid$ or AR$(k)$ column factors with $k=1,2$ and autoregressive parameter $0.85$. Points are the median squared-$\mmd$ error over targets as $N$ increases.}%
    \label{fig:dependent-n-sweep}%
\end{figure}

We study whether $\kerNN$ behaves similarly when the column factors are dependent. Under the confounded staggered-adoption design of \cref{app:synthetic-missingness}~(MNAR) and under MCAR, we compare $\iid$ columns with AR$(k)$ column factors for $k=1,2$ and autoregressive parameter $0.85$; the dependent-column model is specified in \cref{app:dependent-columns}. We use cross-validated $\kerNN$ with the exponential kernel, $d=4$, $n=30$. Under both missingness patterns, the squared-$\mmd$ error in Figure~\ref{fig:dependent-n-sweep} decays with $N$ at similar rates whether the columns are $\iid$ or dependent.

\subsection{Computational alternatives for radius selection}\label{subsec:selection-computation}
Cross validation refits $\kerNN$ for every candidate radius, and we study two ways to reduce this cost: the Direct method selects $\eta$ by minimizing a data-driven version of our error bound in \cref{prop:most-raw-bound}, while entry-wise kernel thinning~(Entry-KT) retains CV and compresses the measurements it is fitted on.

\begin{figure}[!b]%
    \centering
    \includegraphics[width=0.95\linewidth]{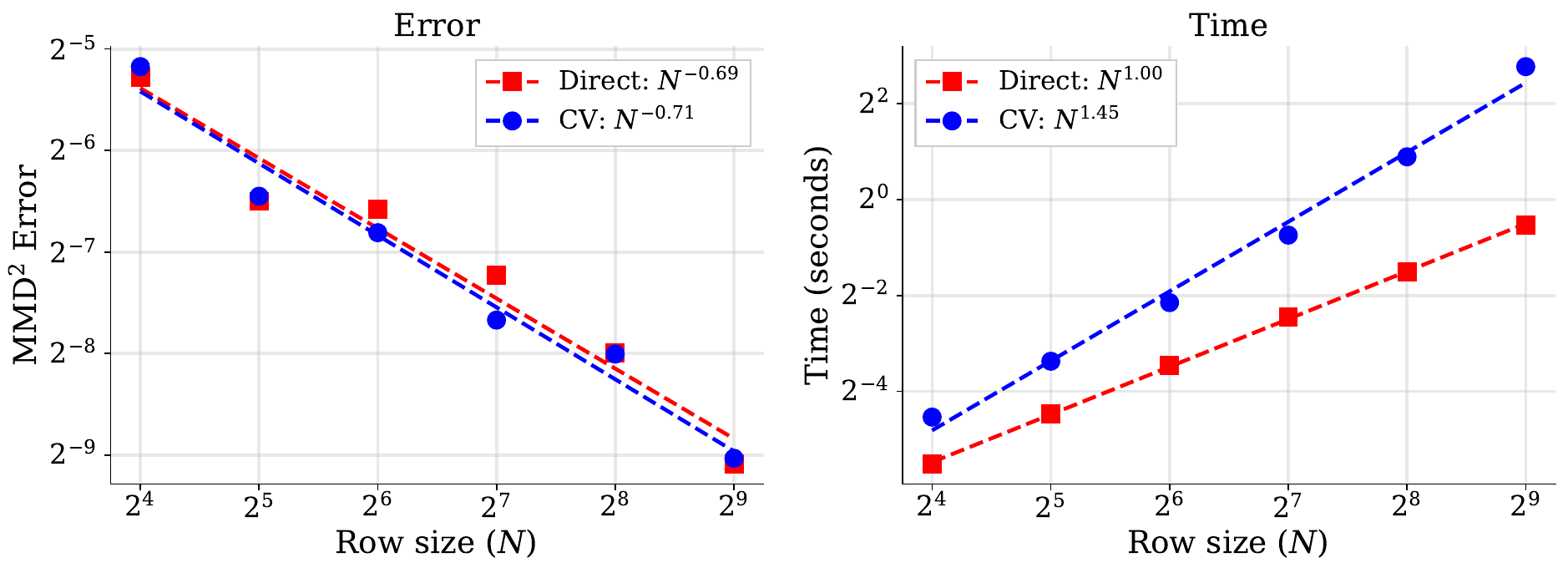}
    \caption{\tbf{CV and Direct under staggered adoption.} The left panel reports the median squared-$\mmd$ error over paired targets and the right panel the median per-target time in seconds, with log--log fits whose slopes are given in the legends. For each $N$, the targets are the missing entries $(i,T)$ of 12 treated units in each of six matrices; their donors share all observed columns of the target row. Direct is implemented as in \cref{eq:non-cv-opt}. The CV error curve coincides with the MNAR curve in the IID panel of \cref{fig:dependent-n-sweep}.}%
    \label{fig:staggered-cv-direct}%
\end{figure}

\paragraph{Analytic approach: a principled and fast way to choose hyper-parameter $\eta$}

Here we demonstrate how the instance dependent theoretical guarantee can provide practical assistance when implementing $\kerNN$ on data. Referring to the discussion following \cref{prop:most-raw-bound}, recall that whenever observation overlap between rows are non-trivial, we may substitute the non-computable neighborhoods $(\lnbhdstarA, \unbhdstarA)$ in \cref{eq:most-raw-guarantee} by its data-driven counterpart $\mbf N_{1, \eta}$ and substitute $\delta$. Then for some value $\delta \in (0, 1)$ in \cref{eq:most-raw-guarantee}~(say $\delta = 1/2$)\footnote{When all donors share the same observed columns with row $1$, as for a treated unit under staggered adoption, the second term of \cref{eq:non-cv-opt} is the same for every $\eta$; since $\delta$ enters only through that term, it does not affect $\etadirect$.}, we propose to minimize the fully data-driven version of the bound \cref{eq:most-raw-guarantee}:
\begin{align}\label{eq:non-cv-opt}
    \etadirect \defeq \argmin_{\eta} \brackets{\eta + \max_{j \in \mbf N_{1, \eta}}\!\!\! \frac{A_{j, 1}\cdot c_0\| \kernel\|_\infty \sqrt{\log(2N/\delta)} }{\sqrt{\sum_{s \neq 1} A_{1, s}A_{j, s} }} +  \frac{4\| \kernel\|_\infty (\log n + 1.5)}{ n \sum_{j \in \mbf N_{1, \eta}} A_{j, 1} }},
\end{align}
where $c_0$ is the constant in \cref{eq:err_defn}. 

\paragraph{Entry-wise kernel thinning}

Entry-KT replaces the $n$ measurements in every observed cell by a smaller subset chosen by kernel thinning~\cite{dwivedi2021kernel} and applies CV to the thinned cells. Once the neighbors are selected, the final $\kerNN$ estimate uses all of their original measurements. We compare Entry-KT with standard thinning, which keeps a uniformly random subset of each cell, and with CV on the full data. Each cell has $n=256$ measurements, of which between $8$ and $128$ are retained. We use the exponential kernel, MCAR missingness, and $N\in\{32,256\}$; implementation details are given in \cref{app:entry-thinning}.

\paragraph{Results}

Figure~\ref{fig:staggered-cv-direct} shows that Direct attains the same error decay as CV and is faster: its per-target runtime is proportional to $N$, whereas that of CV grows faster~(fitted exponents $1.00$ and $1.45$), and at $N=512$ Direct is about ten times faster. The error decay also matches under dependent columns~(\cref{fig:dependent-cv-direct} in \cref{app:direct-dependence}).

\begin{figure}[!t]%
    \centering
    \includegraphics[width=0.95\linewidth]{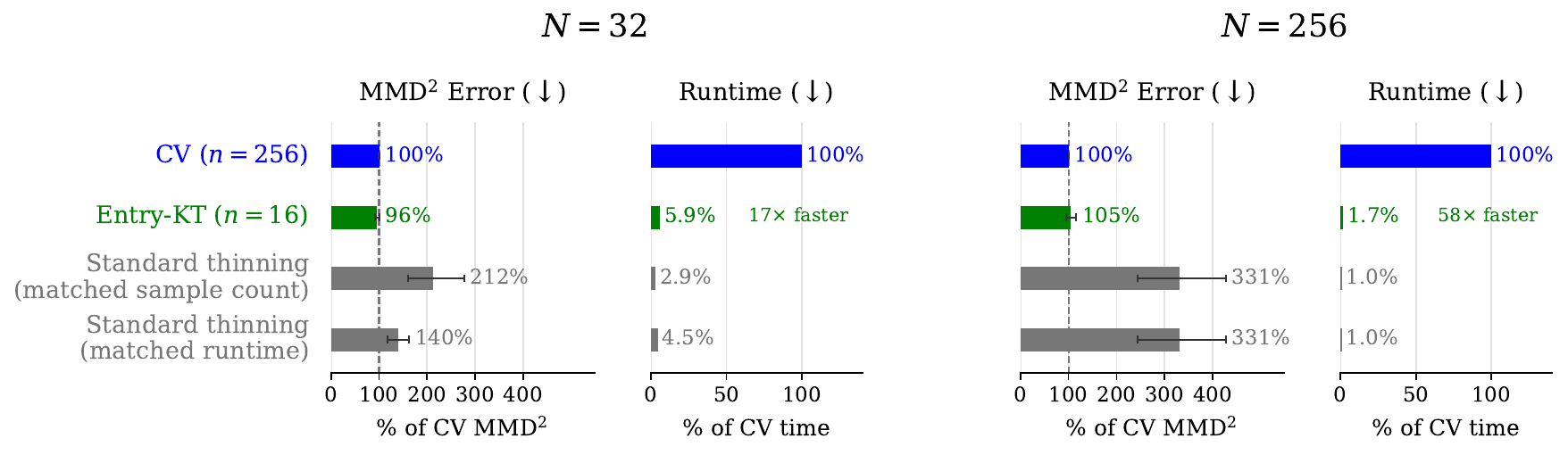}
    \caption{\tbf{Entry-wise kernel thinning preserves CV-level accuracy at substantially lower computational cost.} For $N=32$~(left) and $N=256$~(right), error is the mean per-target squared-$\mmd$ ratio to CV, expressed as a percentage, over 120 targets~(ten matrices with 12 targets each). Error bars are 95\% intervals from resampling matrices. Runtime is the median per-target time as a percentage of CV time, including thinning amortized over the 12 targets. CV uses all 256 measurements per cell; Entry-KT retains 16. Standard thinning matches either the retained sample count or the runtime budget: the latter selects the best tested configuration within Entry-KT's median-runtime budget, retaining 32 measurements for $N=32$ and 16 for $N=256$.}%
    \label{fig:entrykt-comparison}%
\end{figure}

Figure~\ref{fig:entrykt-comparison} shows that Entry-KT is as accurate as CV at a fraction of the cost. Retaining $16$ of the $256$ measurements per cell, its squared-$\mmd$ error is $0.96$ and $1.05$ times that of CV for $N=32$ and $N=256$, while it runs $17$ and $58$ times faster. Standard thinning with the same $16$ measurements has $2.12$ and $3.31$ times the error of CV; giving it the same runtime as Entry-KT lowers this to $1.40$ for $N=32$ and leaves it at $3.31$ for $N=256$. The tradeoff over all retained sizes is shown in \cref{fig:entrykt-pareto} in \cref{app:entry-thinning}.

\subsection{HeartSteps case study}
\label{sec:application}

We present the empirical performance of $\kerNN$ applied to the data collected from the HeartSteps V1 study~(HeartSteps study for short), a clinical trial designed to measure the efficacy of the HeartSteps mobile application for encouraging non-sedentary activity \cite{klasnja2019efficacy}.

\paragraph{Dataset overview and pre-processing}

In the HeartSteps study, $N = 37$ participants were under a 6-week period micro-randomized trial, where they were provided with a mobile application and an activity tracker. The mobile application was designed to send notifications to users at various times during the day to encourage anti-sedentary activity such as stretching or walking. Participants independently received a notification with probability $p = 0.6$ for $5$ pre-determined decision points per day for 40 days~($T = 200$). Participants could be marked as unavailable during decision points if they were driving, walking or running, or if their phone was offline, so notifications were only sent randomly if a participant was available and were never sent if they were unavailable. Thus, the availability of individuals encoded in the randomized trial implies that the treatment~(notification sent) here are subject to individual's latent characteristics such as their personal schedule and daily routines. Further, as notification are never sent during non-available times, positivity is clearly violated in this example.    %
There are three decision states: available and notified, available but not notified, and unavailable, hence not notified. Our analysis is restricted to available decision points.

\begin{figure}[b]
    \centering
    \includegraphics[width=0.95\linewidth]{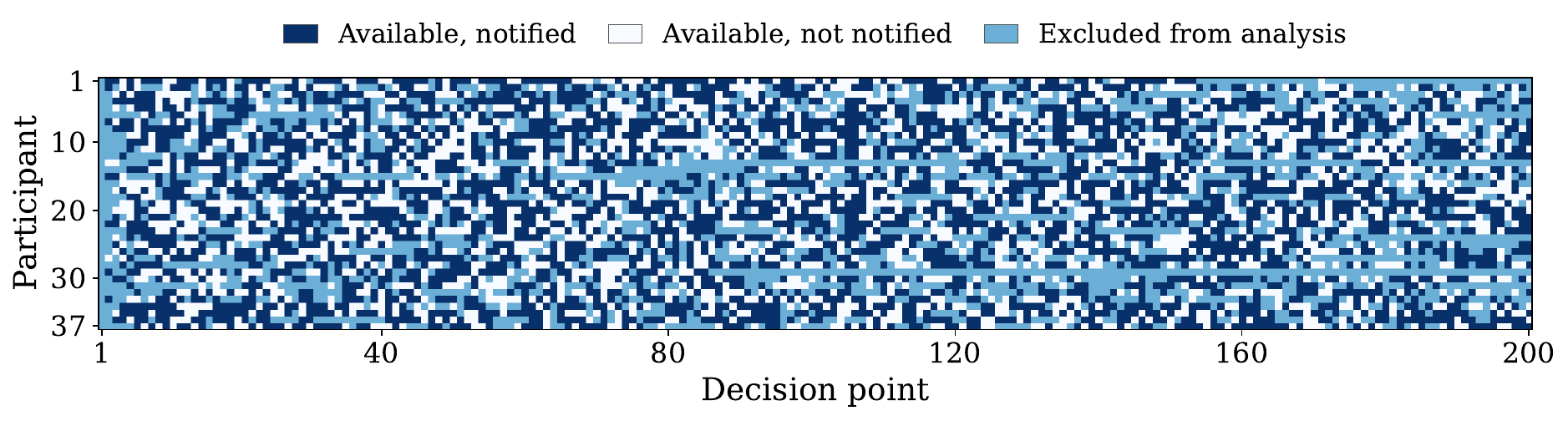}
    \caption{\tbf{HeartSteps V1 data notification pattern.} Dark blue indicates available, notified entries with retained outcomes; white indicates available, non-notified entries with retained outcomes. Light blue combines unavailable decisions, missing or inconsistent records, and entries failing the outcome checks. All light-blue entries are excluded from estimation and evaluation.}
    \label{fig:heart_missing}
\end{figure}

We proceed on our empirical analysis by imposing the potential outcome observation model \cref{mod:potential-outcome} on the HeartSteps data. For each retained decision point $t \in [200]$ of participant $i \in [37]$, the outcome is $n = 12$ five-minute step counts during the following hour. When participant $i$ received notification at decision point $t$, the corresponding step counts are denoted as $X_{1}^{(1)}(i, t), ..., X_{12}^{(1)}(i, t)$. Otherwise, when the participant was available but not given a notification, the step counts are denoted as $X_{1}^{(0)}(i, t), ..., X_{12}^{(0)}(i, t)$. The treatment assignment pattern is represented as the 37 x 200 matrix visualized in \cref{fig:heart_missing}; preprocessing details are given in \cref{app : cv_heartsteps}. 

\subsubsection{Results}

We employ the row-wise nearest neighbors approach for $\kerNN$, matching participants using their common observed decision points outside the target column and averaging the selected participants' distributions at that column. Temporal dependence across decision points are assumed and \cref{sec:dependent-column-appendix} establishes guarantees for this row-wise $\kerNN$ under dependent column factors.
Both the square polynomial and the exponential kernels are used for the algorithm and evaluation. We apply $\log(1+x)$ for step count $x$ and CV follows the synthetic experiments described in \cref{app:experiment-protocol}; the HeartSteps specific split is specified in \cref{app : cv_heartsteps}.

\begin{figure}[bt]%
    \centering
    \includegraphics[width=0.95\linewidth]{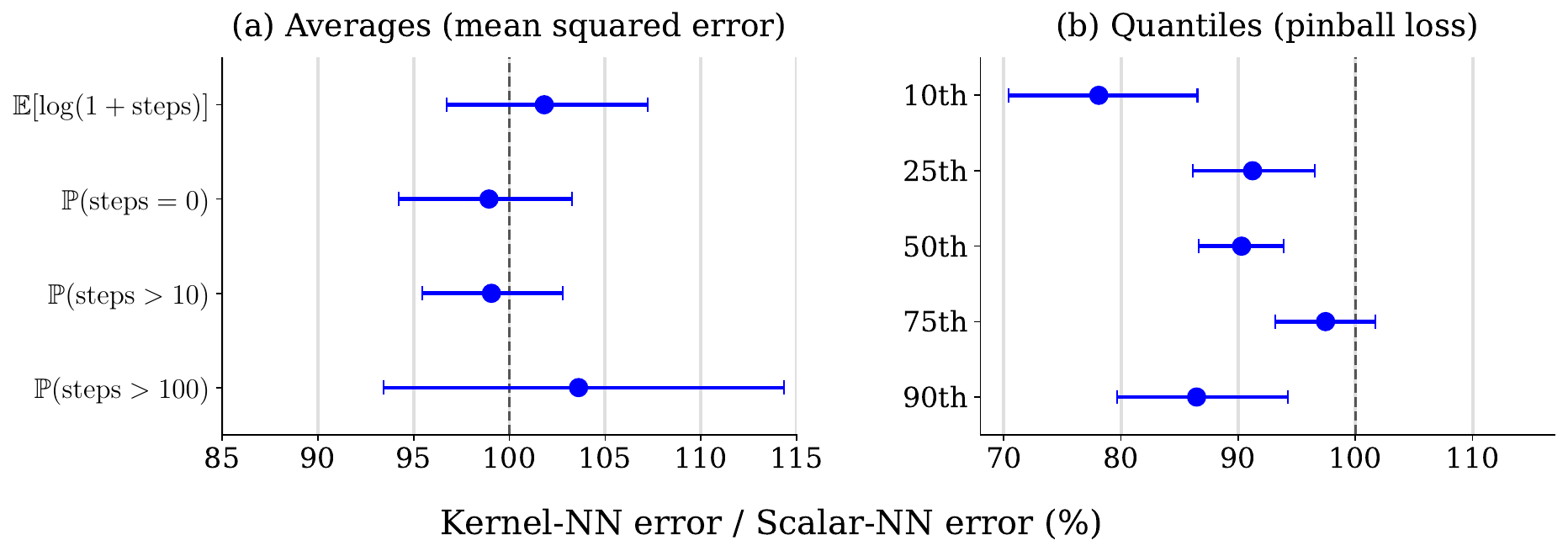}
    \caption{\tbf{Predicting step-count averages and quantiles from one distributional fit.}
    We evaluate 623 of the 3,129 observed notified entries. Each is masked in turn, and its 12 held out five minute step counts are regarded as ground truth to score the predictions. For each masked entry, $\kerNN$ uses the exponential kernel and selects a neighborhood radius by cross-validation; all nine summaries from a single fit. Scalar-NN selects neighbors separately for each step count summary.
    }%
    \label{fig:dist_learning_hs}%
\end{figure}

\begin{figure}[tb]%
    \centering
    \includegraphics[width=0.95\linewidth]{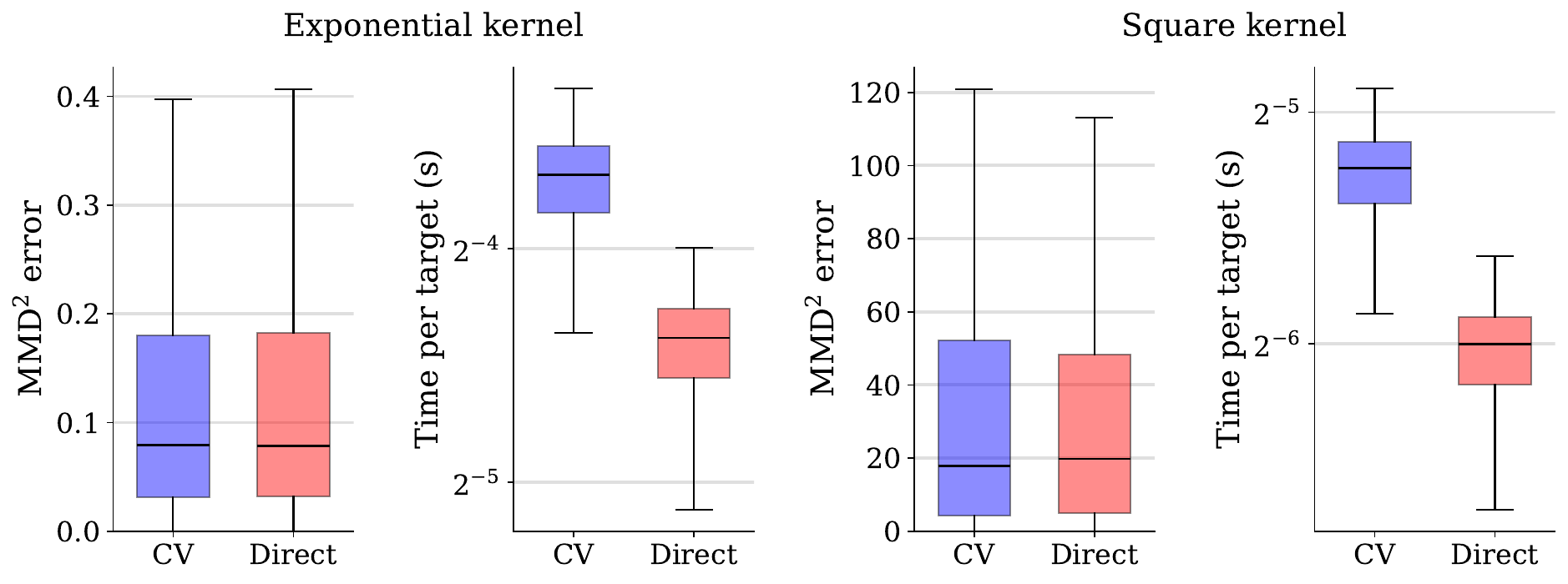}
    \caption{\tbf{Comparing the two versions of $\kerNN$ for HeartSteps data.} This plot shows the performance of \kerNN with respect to squared $\mmd$ error and computational efficiency on estimating notified entries, each masked separately, in the HeartSteps data using both $\etacv$ (denoted as CV) and $\etadirect$ (denoted as Direct). The left and right pairs of panels use the exponential and square kernels, respectively. Runtime includes row matching, radius selection, and construction of the estimate.}%
    \label{fig:direct_cv_hs}%
\end{figure}

We mask each selected notified entry in turn and apply $\kerNN$ to the remaining observations. We evaluate predictions of step count summaries in \cref{fig:dist_learning_hs} and measure distribution estimation error using squared $\mmd$ in \cref{fig:direct_cv_hs}.

\paragraph{Predicting step-count summaries}

For a target entry, $\kerNN$ uses one estimated distribution for estimating all the summary statistics, while Scalar-NN requires separate tuning per statistics. Given the same neighbors, pooling the neighboring observations and then calculating their average is the same to first averaging the measurements and then pooling those statistics across neighbors. For non-linear statistics such as quantiles, the order matters: the median of the pooled observations can differ from the average of the neighbors’ medians. In \cref{fig:dist_learning_hs}, the two methods mean squared errors were within 4$\%$ for all four averages. The single Kernel-NN fit also achieved 9–22$\%$ lower pinball loss~\cite{koenker1978regression} at the 10th, 25th, 50th, and 90th percentiles, with no clear advantage at the 75th.

\paragraph{Comparing two versions of $\kerNN$}

We compare the performance of the two versions of $\kerNN$ on the HeartSteps study data. Specifically, for each selected notified entry $(i, t)$, the empirical distribution of measurements $X_{1:12}^{(1)}(i, t)$ is compared with the output of the two versions of $\kerNN$, $\what \mu_{i, t, \etacv}$ from cross validation and $\what \mu_{i, t, \etadirect}$ from the direct optimization approach \cref{eq:non-cv-opt}, both fitted after masking the target entry. Both versions of $\kerNN$ are implemented~(and also evaluated) on both square and exponential kernels. \cref{fig:direct_cv_hs} shows that, regardless of the chosen kernel, the squared $\mmd$ errors for both versions of $\kerNN$ are similar, with median paired CV-to-Direct runtime ratios of $1.62$ and $1.69$ for the exponential and square kernels, respectively. \cref{fig:direct_cv_hs} implies the potential benefit of using direct optimization over cross-validation as it demonstrates comparable accuracy and lower runtime.

\suppressfloats %

\section{Discussion}

We study the distributional matrix completion problem where the estimand of interest per entry is a multi-dimensional distribution instead of a scalar. We propose a new method  $\kerNN$ which combines ideas from nearest neighbor methods typically used in matrix completion with kernel methods, and provide non-asymptotic guarantees under MNAR data, where missingness can be confounded and positivity can be violated. We specialize these guarantees to two widely studied patterns, staggered adoption and MCAR, and extend the guarantees when column factors are dependent. Under MCAR with independent column factors, we identify a regime where $\kerNN$ is minimax optimal up to a logarithmic factor. 

Several directions remain open. Our method averages over unit-wise nearest neighbors; a doubly robust estimator that combines unit-wise and outcome-wise neighborhoods would extend ideas from scalar matrix completion~\cite{dwivedi2022doubly} to the distributional setting. Our minimax analysis establishes optimality within a specific MCAR regime; characterizing optimal rates beyond this regime, including whether the $1/(p\sqrt{T})$ term is unavoidable, remains an open question.

\newpage

\bibliographystyle{imsart-nameyear} 
\bibliography{ref}

\begin{thebibliography}{52}

\bibitem[\protect\citeauthoryear{Abadie, Diamond and
  Hainmueller}{2010}]{abadie2010synthetic}
\begin{barticle}[author]
\bauthor{\bsnm{Abadie},~\bfnm{Alberto}\binits{A.}},
  \bauthor{\bsnm{Diamond},~\bfnm{Alexis}\binits{A.}} \AND
  \bauthor{\bsnm{Hainmueller},~\bfnm{Jens}\binits{J.}}
(\byear{2010}).
\btitle{Synthetic control methods for comparative case studies: Estimating the
  effect of California’s tobacco control program}.
\bjournal{Journal of the American statistical Association}
\bvolume{105}
\bpages{493--505}.
\end{barticle}
\endbibitem

\bibitem[\protect\citeauthoryear{Abadie {\it et~al.}}{2024}]{abadie2024doubly}
\begin{barticle}[author]
\bauthor{\bsnm{Abadie},~\bfnm{Alberto}\binits{A.}},
  \bauthor{\bsnm{Agarwal},~\bfnm{Anish}\binits{A.}},
  \bauthor{\bsnm{Dwivedi},~\bfnm{Raaz}\binits{R.}} \AND
  \bauthor{\bsnm{Shah},~\bfnm{Abhin}\binits{A.}}
(\byear{2024}).
\btitle{Doubly Robust Inference in Causal Latent Factor Models}.
\bjournal{arXiv preprint arXiv:2402.11652}.
\end{barticle}
\endbibitem

\bibitem[\protect\citeauthoryear{Agarwal, Shah and
  Shen}{2020}]{agarwal2020synthetic}
\begin{barticle}[author]
\bauthor{\bsnm{Agarwal},~\bfnm{Anish}\binits{A.}},
  \bauthor{\bsnm{Shah},~\bfnm{Devavrat}\binits{D.}} \AND
  \bauthor{\bsnm{Shen},~\bfnm{Dennis}\binits{D.}}
(\byear{2020}).
\btitle{Synthetic interventions}.
\bjournal{arXiv preprint arXiv:2006.07691}.
\end{barticle}
\endbibitem

\bibitem[\protect\citeauthoryear{Agarwal {\it
  et~al.}}{2023}]{agarwal2023causal}
\begin{binproceedings}[author]
\bauthor{\bsnm{Agarwal},~\bfnm{Anish}\binits{A.}},
  \bauthor{\bsnm{Dahleh},~\bfnm{Munther}\binits{M.}},
  \bauthor{\bsnm{Shah},~\bfnm{Devavrat}\binits{D.}} \AND
  \bauthor{\bsnm{Shen},~\bfnm{Dennis}\binits{D.}}
(\byear{2023}).
\btitle{Causal matrix completion}.
In \bbooktitle{The Thirty Sixth Annual Conference on Learning Theory}
\bpages{3821--3826}.
\bpublisher{PMLR}.
\end{binproceedings}
\endbibitem

\bibitem[\protect\citeauthoryear{Athey and Imbens}{2022}]{athey2022design}
\begin{barticle}[author]
\bauthor{\bsnm{Athey},~\bfnm{Susan}\binits{S.}} \AND
  \bauthor{\bsnm{Imbens},~\bfnm{Guido~W}\binits{G.~W.}}
(\byear{2022}).
\btitle{Design-based analysis in difference-in-differences settings with
  staggered adoption}.
\bjournal{Journal of Econometrics}
\bvolume{226}
\bpages{62--79}.
\end{barticle}
\endbibitem

\bibitem[\protect\citeauthoryear{Athey {\it et~al.}}{2021}]{athey2021matrix}
\begin{barticle}[author]
\bauthor{\bsnm{Athey},~\bfnm{Susan}\binits{S.}},
  \bauthor{\bsnm{Bayati},~\bfnm{Mohsen}\binits{M.}},
  \bauthor{\bsnm{Doudchenko},~\bfnm{Nikolay}\binits{N.}},
  \bauthor{\bsnm{Imbens},~\bfnm{Guido}\binits{G.}} \AND
  \bauthor{\bsnm{Khosravi},~\bfnm{Khashayar}\binits{K.}}
(\byear{2021}).
\btitle{Matrix completion methods for causal panel data models}.
\bjournal{Journal of the American Statistical Association}
\bvolume{116}
\bpages{1716--1730}.
\end{barticle}
\endbibitem

\bibitem[\protect\citeauthoryear{Bai and Ng}{2021}]{bai2021matrix}
\begin{barticle}[author]
\bauthor{\bsnm{Bai},~\bfnm{Jushan}\binits{J.}} \AND
  \bauthor{\bsnm{Ng},~\bfnm{Serena}\binits{S.}}
(\byear{2021}).
\btitle{Matrix completion, counterfactuals, and factor analysis of missing
  data}.
\bjournal{Journal of the American Statistical Association}
\bvolume{116}
\bpages{1746--1763}.
\end{barticle}
\endbibitem

\bibitem[\protect\citeauthoryear{Ben-Michael, Feller and
  Rothstein}{2022}]{ben2022synthetic}
\begin{barticle}[author]
\bauthor{\bsnm{Ben-Michael},~\bfnm{Eli}\binits{E.}},
  \bauthor{\bsnm{Feller},~\bfnm{Avi}\binits{A.}} \AND
  \bauthor{\bsnm{Rothstein},~\bfnm{Jesse}\binits{J.}}
(\byear{2022}).
\btitle{Synthetic controls with staggered adoption}.
\bjournal{Journal of the Royal Statistical Society Series B: Statistical
  Methodology}
\bvolume{84}
\bpages{351--381}.
\end{barticle}
\endbibitem

\bibitem[\protect\citeauthoryear{Bhattacharya and
  Chatterjee}{2022}]{bhattacharya2022matrix}
\begin{barticle}[author]
\bauthor{\bsnm{Bhattacharya},~\bfnm{Sohom}\binits{S.}} \AND
  \bauthor{\bsnm{Chatterjee},~\bfnm{Sourav}\binits{S.}}
(\byear{2022}).
\btitle{Matrix completion with data-dependent missingness probabilities}.
\bjournal{IEEE Transactions on Information Theory}
\bvolume{68}
\bpages{6762--6773}.
\end{barticle}
\endbibitem

\bibitem[\protect\citeauthoryear{Borgwardt {\it
  et~al.}}{2006}]{borgwardt2006integrating}
\begin{barticle}[author]
\bauthor{\bsnm{Borgwardt},~\bfnm{Karsten~M}\binits{K.~M.}},
  \bauthor{\bsnm{Gretton},~\bfnm{Arthur}\binits{A.}},
  \bauthor{\bsnm{Rasch},~\bfnm{Malte~J}\binits{M.~J.}},
  \bauthor{\bsnm{Kriegel},~\bfnm{Hans-Peter}\binits{H.-P.}},
  \bauthor{\bsnm{Sch{\"o}lkopf},~\bfnm{Bernhard}\binits{B.}} \AND
  \bauthor{\bsnm{Smola},~\bfnm{Alex~J}\binits{A.~J.}}
(\byear{2006}).
\btitle{Integrating structured biological data by kernel maximum mean
  discrepancy}.
\bjournal{Bioinformatics}
\bvolume{22}
\bpages{e49--e57}.
\end{barticle}
\endbibitem

\bibitem[\protect\citeauthoryear{Bradley}{2005}]{bradley2005basic}
\begin{barticle}[author]
\bauthor{\bsnm{Bradley},~\bfnm{Richard~C.}\binits{R.~C.}}
(\byear{2005}).
\btitle{Basic Properties of Strong Mixing Conditions. A Survey and Some Open
  Questions}.
\bjournal{Probability Surveys}
\bvolume{2}
\bpages{107--144}.
\end{barticle}
\endbibitem

\bibitem[\protect\citeauthoryear{Candes and Recht}{2012}]{candes2012exact}
\begin{barticle}[author]
\bauthor{\bsnm{Candes},~\bfnm{Emmanuel}\binits{E.}} \AND
  \bauthor{\bsnm{Recht},~\bfnm{Benjamin}\binits{B.}}
(\byear{2012}).
\btitle{Exact matrix completion via convex optimization}.
\bjournal{Communications of the ACM}
\bvolume{55}
\bpages{111--119}.
\end{barticle}
\endbibitem

\bibitem[\protect\citeauthoryear{Cand{\`e}s and Tao}{2010}]{candes2010power}
\begin{barticle}[author]
\bauthor{\bsnm{Cand{\`e}s},~\bfnm{Emmanuel~J}\binits{E.~J.}} \AND
  \bauthor{\bsnm{Tao},~\bfnm{Terence}\binits{T.}}
(\byear{2010}).
\btitle{The power of convex relaxation: Near-optimal matrix completion}.
\bjournal{IEEE transactions on information theory}
\bvolume{56}
\bpages{2053--2080}.
\end{barticle}
\endbibitem

\bibitem[\protect\citeauthoryear{Chatterjee}{2015}]{chatterjee2015matrix}
\begin{barticle}[author]
\bauthor{\bsnm{Chatterjee},~\bfnm{Sourav}\binits{S.}}
(\byear{2015}).
\btitle{Matrix estimation by universal singular value thresholding}.
\bjournal{The Annals of Statistics}
\bvolume{43}
\bpages{177--214}.
\end{barticle}
\endbibitem

\bibitem[\protect\citeauthoryear{Chen and Shah}{2018}]{chen2018explaining}
\begin{barticle}[author]
\bauthor{\bsnm{Chen},~\bfnm{George~H}\binits{G.~H.}} \AND
  \bauthor{\bsnm{Shah},~\bfnm{Devavrat}\binits{D.}}
(\byear{2018}).
\btitle{Explaining the success of nearest neighbor methods in prediction}.
\bjournal{Foundations and Trends{\textregistered} in Machine Learning}
\bvolume{10}
\bpages{337--588}.
\end{barticle}
\endbibitem

\bibitem[\protect\citeauthoryear{Chernozhukov, Newey and
  Singh}{2022}]{chernozhukov2022automatic}
\begin{barticle}[author]
\bauthor{\bsnm{Chernozhukov},~\bfnm{Victor}\binits{V.}},
  \bauthor{\bsnm{Newey},~\bfnm{Whitney~K}\binits{W.~K.}} \AND
  \bauthor{\bsnm{Singh},~\bfnm{Rahul}\binits{R.}}
(\byear{2022}).
\btitle{Automatic debiased machine learning of causal and structural effects}.
\bjournal{Econometrica}
\bvolume{90}
\bpages{967--1027}.
\end{barticle}
\endbibitem

\bibitem[\protect\citeauthoryear{Cohen, Arbel and
  Deisenroth}{2020}]{cohen2020estimating}
\begin{barticle}[author]
\bauthor{\bsnm{Cohen},~\bfnm{Samuel}\binits{S.}},
  \bauthor{\bsnm{Arbel},~\bfnm{Michael}\binits{M.}} \AND
  \bauthor{\bsnm{Deisenroth},~\bfnm{Marc~Peter}\binits{M.~P.}}
(\byear{2020}).
\btitle{Estimating barycenters of measures in high dimensions}.
\bjournal{arXiv preprint arXiv:2007.07105}.
\end{barticle}
\endbibitem

\bibitem[\protect\citeauthoryear{Cormen {\it
  et~al.}}{2022}]{cormen2022introduction}
\begin{bbook}[author]
\bauthor{\bsnm{Cormen},~\bfnm{Thomas~H}\binits{T.~H.}},
  \bauthor{\bsnm{Leiserson},~\bfnm{Charles~E}\binits{C.~E.}},
  \bauthor{\bsnm{Rivest},~\bfnm{Ronald~L}\binits{R.~L.}} \AND
  \bauthor{\bsnm{Stein},~\bfnm{Clifford}\binits{C.}}
(\byear{2022}).
\btitle{Introduction to algorithms}.
\bpublisher{MIT press}.
\end{bbook}
\endbibitem

\bibitem[\protect\citeauthoryear{Dwivedi and Mackey}{2021}]{dwivedi2021kernel}
\begin{barticle}[author]
\bauthor{\bsnm{Dwivedi},~\bfnm{Raaz}\binits{R.}} \AND
  \bauthor{\bsnm{Mackey},~\bfnm{Lester}\binits{L.}}
(\byear{2021}).
\btitle{Kernel thinning}.
\bjournal{arXiv preprint arXiv:2105.05842}.
\end{barticle}
\endbibitem

\bibitem[\protect\citeauthoryear{Dwivedi {\it
  et~al.}}{2022}]{dwivedi2022doubly}
\begin{barticle}[author]
\bauthor{\bsnm{Dwivedi},~\bfnm{Raaz}\binits{R.}},
  \bauthor{\bsnm{Chin},~\bfnm{Caleb}\binits{C.}},
  \bauthor{\bsnm{Tomkins},~\bfnm{Sabina}\binits{S.}},
  \bauthor{\bsnm{Klasnja},~\bfnm{Predrag}\binits{P.}},
  \bauthor{\bsnm{Murphy},~\bfnm{Susan}\binits{S.}} \AND
  \bauthor{\bsnm{Shah},~\bfnm{Devavrat}\binits{D.}}
(\byear{2022}).
\btitle{Doubly robust nearest neighbors in factor models}.
\bjournal{arXiv preprint arXiv:2211.14297}.
\end{barticle}
\endbibitem

\bibitem[\protect\citeauthoryear{Dwivedi {\it
  et~al.}}{2025}]{dwivedi2022counterfactual}
\begin{barticle}[author]
\bauthor{\bsnm{Dwivedi},~\bfnm{Raaz}\binits{R.}},
  \bauthor{\bsnm{Tian},~\bfnm{Katherine}\binits{K.}},
  \bauthor{\bsnm{Tomkins},~\bfnm{Sabina}\binits{S.}},
  \bauthor{\bsnm{Klasnja},~\bfnm{Predrag}\binits{P.}},
  \bauthor{\bsnm{Murphy},~\bfnm{Susan}\binits{S.}} \AND
  \bauthor{\bsnm{Shah},~\bfnm{Devavrat}\binits{D.}}
(\byear{2025}).
\btitle{Counterfactual inference in sequential experiments}.
\bjournal{The Annals of Statistics}
\bvolume{53}
\bpages{2380--2408}.
\end{barticle}
\endbibitem

\bibitem[\protect\citeauthoryear{Gong, Choi and
  Dwivedi}{2024}]{gong2024supervised}
\begin{barticle}[author]
\bauthor{\bsnm{Gong},~\bfnm{Albert}\binits{A.}},
  \bauthor{\bsnm{Choi},~\bfnm{Kyuseong}\binits{K.}} \AND
  \bauthor{\bsnm{Dwivedi},~\bfnm{Raaz}\binits{R.}}
(\byear{2024}).
\btitle{Supervised Kernel Thinning}.
\bjournal{arXiv preprint arXiv:2410.13749}.
\end{barticle}
\endbibitem

\bibitem[\protect\citeauthoryear{Gretton {\it
  et~al.}}{2007}]{gretton2007kernel}
\begin{barticle}[author]
\bauthor{\bsnm{Gretton},~\bfnm{Arthur}\binits{A.}},
  \bauthor{\bsnm{Fukumizu},~\bfnm{Kenji}\binits{K.}},
  \bauthor{\bsnm{Teo},~\bfnm{Choon}\binits{C.}},
  \bauthor{\bsnm{Song},~\bfnm{Le}\binits{L.}},
  \bauthor{\bsnm{Sch{\"o}lkopf},~\bfnm{Bernhard}\binits{B.}} \AND
  \bauthor{\bsnm{Smola},~\bfnm{Alex}\binits{A.}}
(\byear{2007}).
\btitle{A kernel statistical test of independence}.
\bjournal{Advances in neural information processing systems}
\bvolume{20}.
\end{barticle}
\endbibitem

\bibitem[\protect\citeauthoryear{Gretton {\it
  et~al.}}{2012}]{JMLR:v13:gretton12a}
\begin{barticle}[author]
\bauthor{\bsnm{Gretton},~\bfnm{Arthur}\binits{A.}},
  \bauthor{\bsnm{Borgwardt},~\bfnm{Karsten~M.}\binits{K.~M.}},
  \bauthor{\bsnm{Rasch},~\bfnm{Malte~J.}\binits{M.~J.}},
  \bauthor{\bsnm{Sch{{\"o}}lkopf},~\bfnm{Bernhard}\binits{B.}} \AND
  \bauthor{\bsnm{Smola},~\bfnm{Alexander}\binits{A.}}
(\byear{2012}).
\btitle{A Kernel Two-Sample Test}.
\bjournal{Journal of Machine Learning Research}
\bvolume{13}
\bpages{723-773}.
\end{barticle}
\endbibitem

\bibitem[\protect\citeauthoryear{Gunsilius}{2023}]{gunsilius2023distributional}
\begin{barticle}[author]
\bauthor{\bsnm{Gunsilius},~\bfnm{Florian~F}\binits{F.~F.}}
(\byear{2023}).
\btitle{Distributional synthetic controls}.
\bjournal{Econometrica}
\bvolume{91}
\bpages{1105--1117}.
\end{barticle}
\endbibitem

\bibitem[\protect\citeauthoryear{Hastie {\it et~al.}}{2015}]{hastie2015matrix}
\begin{barticle}[author]
\bauthor{\bsnm{Hastie},~\bfnm{Trevor}\binits{T.}},
  \bauthor{\bsnm{Mazumder},~\bfnm{Rahul}\binits{R.}},
  \bauthor{\bsnm{Lee},~\bfnm{Jason~D}\binits{J.~D.}} \AND
  \bauthor{\bsnm{Zadeh},~\bfnm{Reza}\binits{R.}}
(\byear{2015}).
\btitle{Matrix completion and low-rank SVD via fast alternating least squares}.
\bjournal{The Journal of Machine Learning Research}
\bvolume{16}
\bpages{3367--3402}.
\end{barticle}
\endbibitem

\bibitem[\protect\citeauthoryear{Hofmann, Sch{\"o}lkopf and
  Smola}{2008}]{hofmann2008kernel}
\begin{barticle}[author]
\bauthor{\bsnm{Hofmann},~\bfnm{Thomas}\binits{T.}},
  \bauthor{\bsnm{Sch{\"o}lkopf},~\bfnm{Bernhard}\binits{B.}} \AND
  \bauthor{\bsnm{Smola},~\bfnm{Alexander~J}\binits{A.~J.}}
(\byear{2008}).
\btitle{Kernel methods in machine learning}.
\bjournal{The Annals of Statistics}
\bvolume{36}
\bpages{1171--1220}.
\end{barticle}
\endbibitem

\bibitem[\protect\citeauthoryear{Imbens and Rubin}{2015}]{imbens2015causal}
\begin{bbook}[author]
\bauthor{\bsnm{Imbens},~\bfnm{Guido~W}\binits{G.~W.}} \AND
  \bauthor{\bsnm{Rubin},~\bfnm{Donald~B}\binits{D.~B.}}
(\byear{2015}).
\btitle{Causal inference for statistics, social, and biomedical sciences}.
\bpublisher{Cambridge university press}.
\end{bbook}
\endbibitem

\bibitem[\protect\citeauthoryear{Klasnja {\it
  et~al.}}{2019}]{klasnja2019efficacy}
\begin{barticle}[author]
\bauthor{\bsnm{Klasnja},~\bfnm{Predrag}\binits{P.}},
  \bauthor{\bsnm{Smith},~\bfnm{Shawna}\binits{S.}},
  \bauthor{\bsnm{Seewald},~\bfnm{Nicholas~J}\binits{N.~J.}},
  \bauthor{\bsnm{Lee},~\bfnm{Andy}\binits{A.}},
  \bauthor{\bsnm{Hall},~\bfnm{Kelly}\binits{K.}},
  \bauthor{\bsnm{Luers},~\bfnm{Brook}\binits{B.}},
  \bauthor{\bsnm{Hekler},~\bfnm{Eric~B}\binits{E.~B.}} \AND
  \bauthor{\bsnm{Murphy},~\bfnm{Susan~A}\binits{S.~A.}}
(\byear{2019}).
\btitle{Efficacy of contextually tailored suggestions for physical activity: a
  micro-randomized optimization trial of HeartSteps}.
\bjournal{Annals of Behavioral Medicine}
\bvolume{53}
\bpages{573--582}.
\end{barticle}
\endbibitem

\bibitem[\protect\citeauthoryear{Koenker and
  Bassett}{1978}]{koenker1978regression}
\begin{barticle}[author]
\bauthor{\bsnm{Koenker},~\bfnm{Roger}\binits{R.}} \AND
  \bauthor{\bsnm{Bassett},~\bfnm{Gilbert}\binits{G.}\bsuffix{Jr.}}
(\byear{1978}).
\btitle{Regression quantiles}.
\bjournal{Econometrica}
\bvolume{46}
\bpages{33--50}.
\end{barticle}
\endbibitem

\bibitem[\protect\citeauthoryear{Laumann {\it
  et~al.}}{2023}]{laumann2023kernel}
\begin{barticle}[author]
\bauthor{\bsnm{Laumann},~\bfnm{Felix}\binits{F.}},
  \bauthor{\bsnm{Von~K{\"u}gelgen},~\bfnm{Julius}\binits{J.}},
  \bauthor{\bsnm{Park},~\bfnm{Junhyung}\binits{J.}},
  \bauthor{\bsnm{Sch{\"o}lkopf},~\bfnm{Bernhard}\binits{B.}} \AND
  \bauthor{\bsnm{Barahona},~\bfnm{Mauricio}\binits{M.}}
(\byear{2023}).
\btitle{Kernel-based independence tests for causal structure learning on
  functional data}.
\bjournal{Entropy}
\bvolume{25}
\bpages{1597}.
\end{barticle}
\endbibitem

\bibitem[\protect\citeauthoryear{Li {\it et~al.}}{2019}]{li2019nearest}
\begin{barticle}[author]
\bauthor{\bsnm{Li},~\bfnm{Yihua}\binits{Y.}},
  \bauthor{\bsnm{Shah},~\bfnm{Devavrat}\binits{D.}},
  \bauthor{\bsnm{Song},~\bfnm{Dogyoon}\binits{D.}} \AND
  \bauthor{\bsnm{Yu},~\bfnm{Christina~Lee}\binits{C.~L.}}
(\byear{2019}).
\btitle{Nearest neighbors for matrix estimation interpreted as blind regression
  for latent variable model}.
\bjournal{IEEE Transactions on Information Theory}
\bvolume{66}
\bpages{1760--1784}.
\end{barticle}
\endbibitem

\bibitem[\protect\citeauthoryear{Ma and Chen}{2019}]{ma2019missing}
\begin{barticle}[author]
\bauthor{\bsnm{Ma},~\bfnm{Wei}\binits{W.}} \AND
  \bauthor{\bsnm{Chen},~\bfnm{George~H}\binits{G.~H.}}
(\byear{2019}).
\btitle{Missing not at random in matrix completion: The effectiveness of
  estimating missingness probabilities under a low nuclear norm assumption}.
\bjournal{Advances in neural information processing systems}
\bvolume{32}.
\end{barticle}
\endbibitem

\bibitem[\protect\citeauthoryear{Mazumder, Hastie and
  Tibshirani}{2010}]{mazumder2010softimpute}
\begin{barticle}[author]
\bauthor{\bsnm{Mazumder},~\bfnm{Rahul}\binits{R.}},
  \bauthor{\bsnm{Hastie},~\bfnm{Trevor}\binits{T.}} \AND
  \bauthor{\bsnm{Tibshirani},~\bfnm{Robert}\binits{R.}}
(\byear{2010}).
\btitle{Spectral Regularization Algorithms for Learning Large Incomplete
  Matrices}.
\bjournal{Journal of Machine Learning Research}
\bvolume{11}
\bpages{2287--2322}.
\end{barticle}
\endbibitem

\bibitem[\protect\citeauthoryear{Muandet {\it
  et~al.}}{2017}]{muandet2017kernel}
\begin{barticle}[author]
\bauthor{\bsnm{Muandet},~\bfnm{Krikamol}\binits{K.}},
  \bauthor{\bsnm{Fukumizu},~\bfnm{Kenji}\binits{K.}},
  \bauthor{\bsnm{Sriperumbudur},~\bfnm{Bharath}\binits{B.}},
  \bauthor{\bsnm{Sch{\"o}lkopf},~\bfnm{Bernhard}\binits{B.}} \betal{et~al.}
(\byear{2017}).
\btitle{Kernel mean embedding of distributions: A review and beyond}.
\bjournal{Foundations and Trends{\textregistered} in Machine Learning}
\bvolume{10}
\bpages{1--141}.
\end{barticle}
\endbibitem

\bibitem[\protect\citeauthoryear{Muandet {\it
  et~al.}}{2021}]{muandet2021counterfactual}
\begin{barticle}[author]
\bauthor{\bsnm{Muandet},~\bfnm{Krikamol}\binits{K.}},
  \bauthor{\bsnm{Kanagawa},~\bfnm{Motonobu}\binits{M.}},
  \bauthor{\bsnm{Saengkyongam},~\bfnm{Sorawit}\binits{S.}} \AND
  \bauthor{\bsnm{Marukatat},~\bfnm{Sanparith}\binits{S.}}
(\byear{2021}).
\btitle{Counterfactual mean embeddings}.
\bjournal{Journal of Machine Learning Research}
\bvolume{22}
\bpages{1--71}.
\end{barticle}
\endbibitem

\bibitem[\protect\citeauthoryear{Pollard}{2002}]{pollard2002user}
\begin{bbook}[author]
\bauthor{\bsnm{Pollard},~\bfnm{David}\binits{D.}}
(\byear{2002}).
\btitle{A user's guide to measure theoretic probability}
\banumber{8}.
\bpublisher{Cambridge University Press}.
\end{bbook}
\endbibitem

\bibitem[\protect\citeauthoryear{Rahimi and Recht}{2007}]{rahimi2007random}
\begin{barticle}[author]
\bauthor{\bsnm{Rahimi},~\bfnm{Ali}\binits{A.}} \AND
  \bauthor{\bsnm{Recht},~\bfnm{Benjamin}\binits{B.}}
(\byear{2007}).
\btitle{Random features for large-scale kernel machines}.
\bjournal{Advances in neural information processing systems}
\bvolume{20}.
\end{barticle}
\endbibitem

\bibitem[\protect\citeauthoryear{Rio}{2017}]{rio2017asymptotic}
\begin{bbook}[author]
\bauthor{\bsnm{Rio},~\bfnm{Emmanuel}\binits{E.}}
(\byear{2017}).
\btitle{Asymptotic Theory of Weakly Dependent Random Processes}.
\bseries{Probability Theory and Stochastic Modelling}
\bvolume{80}.
\bpublisher{Springer}.
\end{bbook}
\endbibitem

\bibitem[\protect\citeauthoryear{Rubin}{1976}]{rubin1976inference}
\begin{barticle}[author]
\bauthor{\bsnm{Rubin},~\bfnm{Donald~B}\binits{D.~B.}}
(\byear{1976}).
\btitle{Inference and missing data}.
\bjournal{Biometrika}
\bvolume{63}
\bpages{581--592}.
\end{barticle}
\endbibitem

\bibitem[\protect\citeauthoryear{Sadhukhan, Paul and
  Dwivedi}{2024}]{sadhukhan2024twosided}
\begin{barticle}[author]
\bauthor{\bsnm{Sadhukhan},~\bfnm{Tathagata}\binits{T.}},
  \bauthor{\bsnm{Paul},~\bfnm{Manit}\binits{M.}} \AND
  \bauthor{\bsnm{Dwivedi},~\bfnm{Raaz}\binits{R.}}
(\byear{2024}).
\btitle{Two-Sided Nearest Neighbors: An adaptive and minimax optimal procedure
  for matrix completion}.
\bjournal{arXiv preprint arXiv:2411.12965}.
\end{barticle}
\endbibitem

\bibitem[\protect\citeauthoryear{Sch{\"o}lkopf and
  Smola}{2002}]{scholkopf2002learning}
\begin{bbook}[author]
\bauthor{\bsnm{Sch{\"o}lkopf},~\bfnm{Bernhard}\binits{B.}} \AND
  \bauthor{\bsnm{Smola},~\bfnm{Alexander~J}\binits{A.~J.}}
(\byear{2002}).
\btitle{Learning with kernels: support vector machines, regularization,
  optimization, and beyond}.
\bpublisher{MIT press}.
\end{bbook}
\endbibitem

\bibitem[\protect\citeauthoryear{Sejdinovic {\it
  et~al.}}{2013}]{sejdinovic2013equivalence}
\begin{barticle}[author]
\bauthor{\bsnm{Sejdinovic},~\bfnm{Dino}\binits{D.}},
  \bauthor{\bsnm{Sriperumbudur},~\bfnm{Bharath}\binits{B.}},
  \bauthor{\bsnm{Gretton},~\bfnm{Arthur}\binits{A.}} \AND
  \bauthor{\bsnm{Fukumizu},~\bfnm{Kenji}\binits{K.}}
(\byear{2013}).
\btitle{Equivalence of distance-based and RKHS-based statistics in hypothesis
  testing}.
\bjournal{The Annals of Statistics}
\bvolume{41}
\bpages{2263--2291}.
\end{barticle}
\endbibitem

\bibitem[\protect\citeauthoryear{Shah {\it
  et~al.}}{2022}]{shah2022counterfactual}
\begin{barticle}[author]
\bauthor{\bsnm{Shah},~\bfnm{Abhin}\binits{A.}},
  \bauthor{\bsnm{Dwivedi},~\bfnm{Raaz}\binits{R.}},
  \bauthor{\bsnm{Shah},~\bfnm{Devavrat}\binits{D.}} \AND
  \bauthor{\bsnm{Wornell},~\bfnm{Gregory~W}\binits{G.~W.}}
(\byear{2022}).
\btitle{On counterfactual inference with unobserved confounding}.
\bjournal{arXiv preprint arXiv:2211.08209}.
\end{barticle}
\endbibitem

\bibitem[\protect\citeauthoryear{Singh, Xu and Gretton}{2024}]{singh2023kernel}
\begin{barticle}[author]
\bauthor{\bsnm{Singh},~\bfnm{Rahul}\binits{R.}},
  \bauthor{\bsnm{Xu},~\bfnm{Liyuan}\binits{L.}} \AND
  \bauthor{\bsnm{Gretton},~\bfnm{Arthur}\binits{A.}}
(\byear{2024}).
\btitle{Kernel methods for causal functions: dose, heterogeneous and
  incremental response curves}.
\bjournal{Biometrika}
\bvolume{111}
\bpages{497--516}.
\end{barticle}
\endbibitem

\bibitem[\protect\citeauthoryear{Steinwart and
  Christmann}{2008}]{steinwart2008support}
\begin{bbook}[author]
\bauthor{\bsnm{Steinwart},~\bfnm{Ingo}\binits{I.}} \AND
  \bauthor{\bsnm{Christmann},~\bfnm{Andreas}\binits{A.}}
(\byear{2008}).
\btitle{Support vector machines}.
\bpublisher{Springer Science \& Business Media}.
\end{bbook}
\endbibitem

\bibitem[\protect\citeauthoryear{Szab{\'o} {\it
  et~al.}}{2016}]{szabo2016learning}
\begin{barticle}[author]
\bauthor{\bsnm{Szab{\'o}},~\bfnm{Zolt{\'a}n}\binits{Z.}},
  \bauthor{\bsnm{Sriperumbudur},~\bfnm{Bharath~K}\binits{B.~K.}},
  \bauthor{\bsnm{P{\'o}czos},~\bfnm{Barnab{\'a}s}\binits{B.}} \AND
  \bauthor{\bsnm{Gretton},~\bfnm{Arthur}\binits{A.}}
(\byear{2016}).
\btitle{Learning theory for distribution regression}.
\bjournal{Journal of Machine Learning Research}
\bvolume{17}
\bpages{1--40}.
\end{barticle}
\endbibitem

\bibitem[\protect\citeauthoryear{Vershynin}{2018}]{vershynin2018high}
\begin{bbook}[author]
\bauthor{\bsnm{Vershynin},~\bfnm{Roman}\binits{R.}}
(\byear{2018}).
\btitle{High-dimensional probability: An introduction with applications in data
  science}
\bvolume{47}.
\bpublisher{Cambridge university press}.
\end{bbook}
\endbibitem

\bibitem[\protect\citeauthoryear{Wainwright}{2019}]{wainwright2019high}
\begin{bbook}[author]
\bauthor{\bsnm{Wainwright},~\bfnm{Martin~J}\binits{M.~J.}}
(\byear{2019}).
\btitle{High-dimensional statistics: A non-asymptotic viewpoint}
\bvolume{48}.
\bpublisher{Cambridge University Press}.
\end{bbook}
\endbibitem

\bibitem[\protect\citeauthoryear{Wenliang {\it
  et~al.}}{2023}]{wenliang2023distributional}
\begin{barticle}[author]
\bauthor{\bsnm{Wenliang},~\bfnm{Li~Kevin}\binits{L.~K.}},
  \bauthor{\bsnm{D{\'e}letang},~\bfnm{Gr{\'e}goire}\binits{G.}},
  \bauthor{\bsnm{Aitchison},~\bfnm{Matthew}\binits{M.}},
  \bauthor{\bsnm{Hutter},~\bfnm{Marcus}\binits{M.}},
  \bauthor{\bsnm{Ruoss},~\bfnm{Anian}\binits{A.}},
  \bauthor{\bsnm{Gretton},~\bfnm{Arthur}\binits{A.}} \AND
  \bauthor{\bsnm{Rowland},~\bfnm{Mark}\binits{M.}}
(\byear{2023}).
\btitle{Distributional Bellman Operators over Mean Embeddings}.
\bjournal{arXiv preprint arXiv:2312.07358}.
\end{barticle}
\endbibitem

\bibitem[\protect\citeauthoryear{Williams and Seeger}{2000}]{williams2000using}
\begin{barticle}[author]
\bauthor{\bsnm{Williams},~\bfnm{Christopher}\binits{C.}} \AND
  \bauthor{\bsnm{Seeger},~\bfnm{Matthias}\binits{M.}}
(\byear{2000}).
\btitle{Using the Nystr{\"o}m method to speed up kernel machines}.
\bjournal{Advances in neural information processing systems}
\bvolume{13}.
\end{barticle}
\endbibitem

\bibitem[\protect\citeauthoryear{Xu}{2017}]{xu2017generalized}
\begin{barticle}[author]
\bauthor{\bsnm{Xu},~\bfnm{Yiqing}\binits{Y.}}
(\byear{2017}).
\btitle{Generalized synthetic control method: Causal inference with interactive
  fixed effects models}.
\bjournal{Political Analysis}
\bvolume{25}
\bpages{57--76}.
\end{barticle}
\endbibitem

\end{thebibliography}

\newpage

\numberwithin{theorem}{section}
\numberwithin{mylemma}{section}
\numberwithin{mydefinition}{section}
\numberwithin{myproposition}{section}
\numberwithin{mycorollary}{section}
\renewcommand{\theHtheorem}{\theHsection.\arabic{theorem}}
\renewcommand{\theHmylemma}{\theHsection.\arabic{mylemma}}
\renewcommand{\theHmydefinition}{\theHsection.\arabic{mydefinition}}
\renewcommand{\theHmyproposition}{\theHsection.\arabic{myproposition}}
\renewcommand{\theHmycorollary}{\theHsection.\arabic{mycorollary}}
\numberwithin{figure}{section}
\numberwithin{table}{section}
\numberwithin{algocf}{section}

\begin{appendix}

\section{Discussion on notations}\label{sec : disc finite dim model}

We first set the ground on the notations used throughout the Appendix. Next, we elaborate on the implications and the extensions made in our proposed model, that was introduced in \cref{exam:gaussian_family} but not thoroughly discussed in the main text.

\paragraph{Additional notation}
    
    For any random variable $X \in \real$, the $\psi_2$-Orlicz norm is defined as $ \|X\|_{\psi_2 }\defeq \inf \braces{ c > 0 : \E[ \psi_2(|X|/c) ] \leq 1 }$ where $\psi_2 \defeq \exp\{ x^2 \} - 1$. 
    We use $c$~(or $c'$) to be positive universal constants that could be different from line to line. 
    
    Recall that, without loss of generality, our target estimand was set as the distribution $\mu_{1, 1}$. Accordingly, we use 
    \begin{align}
        \AN &\defeq \sbraces{ A_{j, 1}, j \in [N] } \qtext{and}\Arho \defeq  \sbraces{ A_{j, s}, s \geq [T] \setminus \sbraces{1} }, \\ \mc D_1 &\defeq \{ X_k(i, 1): A_{i, 1} = 1,  i\in [N], k \in [n]\}
        \qtext{and} \\ 
        \mc D_{-1} &\defeq \{ X_k(i, t), k \in [n] : A_{i, t} = 1,  i\in [N], t\in[T], t\neq 1, k \in [n] \}.
    \end{align}
    That is, $\AN$ denotes the missingness of the first outcome (column) and $\mc D_{1}$ denotes the corresponding measurements, while $\Arho$ and $\mc D_{-1}$ denote the corresponding quantities for the remaining outcomes (columns).

    Similarly, define $\mc V_{-1} \defeq \sbraces{ v_2, v_3, ..., v_T }$ and $\mc U_{-1} \defeq \sbraces{ u_2, ..., u_N }$. Notice that conditioned on $\sbraces{\wholerand}$, the set $\mbf N_{1, \eta}$ is deterministic as  the set $\mc D_{-1}$ is used in the first step of \kerNN\ while $\mc D_{1}$ is used in the second step.

\section{Generalization of prior work}
\label{sec : generalization and hetero}

We show here that the model and algorithm proposed in \cite{li2019nearest} can be recovered by our model \cref{model : dist matrix completion} and a slight modification of the $\kerNN$ algorithm. Let $\phi(x) = \frac{1}{\sqrt{2\pi}}e^{-x^2/2}$ be the density of a standard Gaussian distribution on a real line.

Consider the (scalar) matrix completion problem from \cite{li2019nearest}, where $(i, t)$-th entry in the matrix satisfies
\begin{align}
    X_1(i, t) = \begin{cases}
    \theta_{i, t} + \vareps_{i, t} &\qtext{if} A_{i, t} = 1 \\ 
    \trm{unknown}        &\qtext{otherwise}
    \end{cases}
    \label{eq:scalar_mc}
\end{align}
with $\vareps_{i, t}$ drawn \iid from $\mc N(0, \sigma^2)$ and $\theta_{i, t}$, the mean of $X_1(i, t)$ satisfying a factor model $\theta_{i, t} = m_1(u_i, v_t)$ for some function $m_1$, and a collection of latent factors $\mc U = \sbraces{u_i}_{i \in [N]}$ and $\mc V = \sbraces{v_t}_{t \in [T]}$. The following result formalizes our claim,

\begin{lemma}[\tbf{Recovering model and algorithm of \cite{li2019nearest}}]
\label{prop : generalization of prior work}
 The scalar matrix completion set-up~\cref{eq:scalar_mc} of \cite{li2019nearest} can be recovered as a special case of distributional matrix completion problem~\cref{model : dist matrix completion} with $n=1$ measurements in each observed entry, where \cref{assump:factorization} holds for a Gaussian location family $\mc P = \sbraces{\mu_{i, t}}$ with $\mu_{i, t} = \mc N(\theta_{i, t}, \sigma^2)$ and the linear kernel $\kernel(x, x') = xx'$. Furthermore, the scalar nearest neighbor algorithm of \cite{li2019nearest} can be recovered as a special case of $\kerNN$ with linear kernel and distance $\rho^{\textsc{V}}_{i, j}$ defined in \cref{def : modified dissim} with $n=1$.
\end{lemma}

We emphasize the distance $\rho_{i, j}$ from \kerNN\ \cref{eq:row-metric} cannot be constructed when only one sample~($n = 1$) is available, since U-statistics of two arguments is well-defined when at least two samples are available. For \cite{li2019nearest}, as stated in \cref{prop : generalization of prior work}, homogeneous variance assumption across samples is a critical assumption. Note that for this case where $\empdistZ{j, s} = \dirac_{X_1(j, s)}$ whenever $A_{j, s} = 1$, we have
\begin{align}
\E[ \rho^{\textsc{V}}_{i, j} | \mc U ] = \|g(u_i, \cdot) - g(u_j, \cdot)\|_2^2 + \text{Var}(X_{1}(i, t)) + \text{Var}(X_1(j, t)).
\label{eqn : biased V stat}
\end{align}
And hence when constructing neighbors, the analysis requires that the equality holds $\text{Var}(X_{1}(i, t)) = \text{Var}(X_1(j, t)) = \sigma^2$ for consistent estimates. In contrast, our U-statistics-based distance $\rho_{i, j}$ with $n\geq 2$ samples debiases, i.e., $\E[\rho_{i, j} | \mc U] = \knorm{g(u_i, \cdot) - g(u_j, \cdot)}$, thereby allowing for heterogeneous variances in each entry.

\subsection{Proof of \cref{prop : generalization of prior work}: Recovering model and algorithm of \cite{li2019nearest}}

We set the missingness $A_{i,t }$ of \cref{eq:scalar_mc,model : dist matrix completion} follow MCAR structure, which corresponds to the missing pattern considered in \cite{li2019nearest}. Without loss of generality, the latent factors $u_i, v_t$ for both models \cref{model : dist matrix completion,eq:scalar_mc} have identical finite discrete distribution on a compact support $\mc S_u, \mc S_v \subset [-1, 1]^r$ respectively.
It suffices to show that the measurements for both models have the same distribution --- for that end, we first show that the marginal distributions of measurements are identical and then show that the joint distribution of measurements are identical as well.

Whenever latent values are fixed as $u_i = u , v_t = v $, the kernel mean embedding of each Gaussian distribution $\mu_{i,t}$ is a linear function through the center, i.e. $T_\kernel(\mu_{i, t})y = \theta_{i, t}y$. Set the operator $g$ of interest to be $g(u, v)(x) = m_1(u, v)x$, meaning that image of $g$ for every $u, v$ is a linear mapping through the origin with slope $m_1(u, v)$. Then a linear kernel along with \cref{assump:factorization} and the operator $g$ induces $\mu_{i, t}\kernel(y) = \int x d\mu_{i, t}\cdot y = m_{1}(u_i, v_t)\cdot y$ for all $y \in \real$, thereby implying $\theta_{i, t} = m_1(u_i, v_t)$.

Next we recover the algorithm of \cite{li2019nearest}. Notice that $\int \kernel(\cdot, x) d \dirac_{X_1(i, s)}(x) = \kernel(\cdot,X_1(i, s))$. Then under the linear kernel $\kernel(x, x') = xx'$, we have
\begin{align}
    \mmd_\kernel^2( \dirac_{X_1(i, s)}, \dirac_{X_1(j, s)} ) &= \knorm{\dirac_{X_1(i, s)}\kernel - \dirac_{X_1(j, s)}\kernel}^2\\
    &= \knorm{\kernel(\cdot, X_1(i, s)) - \kernel(\cdot, X_1(j, s))}^2\\
    &= X_1(i, s)^2 + X_1(j, s)^2 - 2X_1(i, s)X_1(j, s).
\end{align}
So we may conclude that
\begin{align}
\rho^{\textsc{V}}_{i, j} &\defeq \frac{\sum_{s \neq t} A_{i, s}A_{j, s} \text{MMD}_\kernel^2(\dirac_{X_1(i, s)}, \dirac_{X_1(j, s)}) }{\sum_{s \neq t}A_{i, s}A_{j, s}} \\
&=\frac{\sum_{s \neq t} A_{i, s}A_{j, s} ( X_1(i, s) - X_1(j, s) )^2 }{\sum_{s \neq t}A_{i, s}A_{j, s}} \defeq \varrho_{i, j}.
\end{align}
So the dissimilarity measure \cref{eq:snn-distance} used in scalar nearest neighbor is recovered using $\rho^{\textsc{V}}_{i, j}$ with linear kernels, implying that the neighborhood would be identical for the modified $\kerNN$ and that of \cite{li2019nearest}. Further, by plugging $n = 1$ for the barycenter formula in \cref{eq:mmd-barycenter}, we simply recover the sample averaging of observations within the neighborhood, which again matches the final output of \cite{li2019nearest}.

\section{Proof of Prop. \ref{prop:most-raw-bound}: Instance-based guarantee}
\label{sec:proof_of_instance_based}
We briefly summarize the proof outline: the proof starts by decomposing a partially integrated $\mmd$ metric~\cref{lem : MMD decomp}, then the decomposed terms are bounded separately on a high-probability event at which the row metric $\rho_{i, j}$ concentrates around its mean.

Without loss of generality, we assume that $\uarand$ are such that for any $j \in [N]$ and $j\neq 1$, 
\begin{align}
     A_{j, 1} = 1 \implies \sum_{s\neq 1}A_{1, s}A_{j, s} > 0 \quad \text{and} \quad \sum_{j \in \lnbhdstarA}A_{j, 1} > 0,
     \label{event:good2}
\end{align}
because otherwise the terms defined in \cref{prop:most-raw-bound} are not well-defined, hence the guarantee therein is vacuous.

Now define 
\begin{align}
  b(j, 1) &\defeq \int \kernel(x, \cdot) d\mu_{j, 1}(x) - \int \kernel (x, \cdot) d\mu_{1, 1}(x) \qtext{and}\\
  v_n(j, 1) &\defeq \int \kernel(x, \cdot ) d \empdistZ{j, 1}(x) - \int \kernel(x, \cdot) d \mu_{j, 1} (x).
  \label{eq:bias_variance}
\end{align}
Notice that $b(j, 1)$ is analogous to a bias term that characterizes how far the (unknown) distribution $\mu_{j, 1}$ is from the target distribution $\mu_{1, 1}$. On the other hand, the term $v_n$ is analogous to a sampling error as its kernel norm characterizes how far the empirical (observed) distribution $\empdistZ{j, 1}$ is from the true distribution $\mu_{j, 1}$. Note the two identities,
\begin{align}
    \E[\knorm{b(j,1)}^2\vert u_1, u_j] \seq{\cref{eq:factor_model},\eqref{def : Delta and etai}} \gio
    \qtext{and}
    \E[v_n(j, 1) \vert v_1, u_j] = 0.
    \label{eq:bias_variance_mean}
\end{align}
The first identify of \cref{eq:bias_variance_mean} can be shown by applying the following in order: assumption \cref{eq:factor_model}, the definition \cref{def : Delta and etai}, and the independence $(u_j, u_1) \indep v_1$ from \cref{assump:latent-independence}. For the second identity of \cref{eq:bias_variance_mean}, observing the following sequence of equalities is sufficient, 
\begin{align}
    \int \kernel(x, \cdot ) d\empdistZ{j, 1}(x) &= \frac{1}{n}\sum_{\ell = 1}^{n}\E\biggbrackets{   \int \kernel(x, \cdot)d\delta_{X_\ell(j, 1)}(x) \Big| v_1, u_j}\\
    &= \frac{1}{n}\sum_{\ell = 1}^n \E[\kernel(X_\ell, \cdot) | v_1, u_j] = \int \kernel(x, \cdot) d\mu_{j, 1}(x); 
\end{align}
where the first equality is due to linearity of empirical distributions, the second equality due to integrating over the delta measure $\delta_{X_\ell(i,t)}$, and the last equality due to identically distributed $X_\ell(i, t)$ across $\ell \in [n]$, according to \cref{assump : measurement generation}.

The next lemma (proven in \cref{proof : decomp}) provides a characterization of the MMD error for the \kerNN estimate in terms of these bias-variance like terms.

\begin{lemma}[\tbf{Conditional $\mmd$ error decomposition}]
\label{lem : MMD decomp}
Let \cref{assump:factorization,assump:latent-independence,assump : measurement generation} hold. Then the estimate $\what{\mu}_{1, 1, \eta}$ satisfies
\begin{align}
    \E\brackets{ \| \what{\mu}_{1, 1, \eta} - \mu_{1, 1}\|_{\kernel}^2 | \mc V_{-1}, \mc D_{-1}, \mc A, \mc U } &\leq \frac{\bindic{ \sum_{j \in \mbf N_{1, \eta}} A_{j, 1} \geq 1 }}{(\sum_{j \in \mbf N_{1, \eta}} A_{j, 1})^2} \sum_{j \in \mbf N_{1, \eta}} A_{j, 1} \cdot \E\brackets{ \| v_n(j, 1)\|_{\kernel}^2 | u_j } \\
    &+\bindic{ \sum_{j \in \mbf N_{1, \eta}} A_{j, 1} \geq 1 } \max_{j \in \mbf N_{1, \eta} }A_{j, 1} 
    \cdot \E[\knorm{b(j, 1)}^2\vert u_1, u_j]\\
    &+ 2\|\kernel\|_\infty \cdot \bindic{\sum_{j \in \mbf N_{1, \eta}} A_{j, 1} = 0},
    \label{eqn : conditional decomp}
\end{align}
for any $(\mc V_{-1}, \mc D_{-1}, \mc A, \mc U)$ on which the RHS of \cref{eqn : conditional decomp} is well defined, i.e. $\sum_{j \in \mbf N_{1, \eta}}A_{j, 1} > 0$.

\end{lemma}

The next lemma, with proof in \cref{proof : concentration}, shows that the dissimilarity measure $\rho_{j, 1}$ has mean $\gio$ and exhibits a tight concentration around it:
\begin{lemma}[\tbf{Conditional concentration for row metric}]\label{lem:conditional-concentration}
Let \cref{assump:factorization,assump:latent-independence,assump : unobs confounding,assump : measurement generation} hold.
Then for any unit $j$ with $A_{j, 1} = 1$ and $\overoj >0$ and any $\delta \in (0, 1)$, we have
\begin{align}
    \Prob \biggbraces{ | \rho_{j, 1} - \E[\knorm{b(j, 1)}^2\vert u_1, u_j] | > \frac{ 8e^{1/e - 1/2} \| \kernel \|_\infty \sqrt{\log(2/\delta)}}{\sqrt{ 2\log 2\overoj }} \bigg| \mc U, \mc A } \leq \delta.
\end{align}
\end{lemma}

Recall from \cref{eq:err_defn} that $\errja =\frac{ 8e^{1/e - 1/2} \| \kernel \|_\infty \sqrt{\log(2/\delta)}}{\sqrt{ 2\log 2\overoj }} $. 
Given the two lemmas, we now proceed to establish \cref{prop:most-raw-bound}, which builds on the RHS of \cref{eqn : conditional decomp} once we have a handle on the bias-like term $\E\brackets{ \| b(j, 1)\|_{\kernel}^2 | u_1, u_j }$ and the variance-like term $\E\brackets{ \| v_n(j, 1)\|_{\kernel}^2 | u_j}$.

\paragraph{Controlling $\E\brackets{ \| b(j, 1)\|_{\kernel}^2 | u_1, u_j }$}
Conditioned on $\sbraces{\uarand}$, define the event
\begin{align}
    \overlapconc &\defeq \braces{ | \rho_{j, 1} -\E\brackets{ \| b(j, 1)\|_{\kernel}^2 | u_1, u_j} | \leq \errja \quad \text{for all $j$ such that $A_{j, 1} = 1$}}
    \label{event:dist-conc}
\end{align}
and note that \cref{lem:conditional-concentration} implies that $\Prob[\overlapconc \vert \uarand] \geq 1-N\delta$. 

Next, recall the definitions of $(\lnbhdstarA, \unbhdstarA)$ from \cref{def : pop nbhd data centered}, both of which are well-defined by assuming values $\sbraces{ \uarand }$ satisfying \cref{event:good2}.
We note that on the event $\overlapconc$
\begin{align}\label{eq:computable-bound-bias}
    \gio = \E\brackets{ \| b(j, 1)\|_{\kernel}^2 | u_1, u_j} 
    \leq \rho_{j, 1} + \errja,
\end{align}
so that on this event for any $j \in \mbf N_{1, \eta}$, defined in \cref{eq:actual_nbhd}, we have $\gio \leq \eta + \errja$ so that 
\begin{align}
    \mbf N_{1, \eta} \subseteq \unbhdstarA \qtext{on the event} \overlapconc.
    \label{eq:n_ubnd}
\end{align}
Similarly, for $j \in \lnbhdstarA$, on the event $\overlapconc$, we find that
\begin{align}
    \gio \leq \eta - \errja 
    \implies \rho_{j, 1} \leq \gio + \errja 
    \leq \eta
\end{align}
so that
\begin{align}
    \lnbhdstarA \subseteq \mbf N_{1, \eta} 
    \qtext{on the event} \overlapconc.
    \label{eq:n_lbnd}
\end{align}
Thus, we also have
\begin{align}
    \sum_{j \in \lnbhdstarA}A_{j, 1} \leq \sum_{j \in \mbf N_{1, \eta}}A_{j, 1} \leq \sum_{j \in \unbhdstarA} A_{j, 1}
    \qtext{on the event} \overlapconc,
    \label{eq:a_sum}
\end{align}
and the immediate consequence of \cref{eq:a_sum} along with $\sbraces{\uarand}$ satisfying \cref{event:good2} is that the RHS of \cref{eqn : conditional decomp} is well-defined, thereby allowing us to utilize \cref{lem : MMD decomp}. 

Consequently on $\overlapconc$, we can write 
\begin{align}
    \bindic{ \sum_{j \in \mbf N_{1, \eta}} A_{j, 1} \geq 1 } \max_{j \in \mbf N_{1, \eta} }A_{j, 1} 
    \cdot \E[\knorm{b(j, 1)}^2\vert u_1, u_j]
    &\sless{\cref{eq:n_ubnd}} \eta + \max_{j \in \unbhdstarA} \errja \label{eq:bias_bnd} \qtext{and}\\
    \|\kernel\|_\infty \cdot \bindic{\sum_{j \in \mbf N_{1, \eta}} A_{j, 1} = 0}
    &\sless{\cref{eq:a_sum}}  \|\kernel\|_\infty \cdot \bindic{\sum_{j \in \lnbhdstarA} A_{j, 1} = 0}\label{eq:last_term_bnd}
\end{align}

\paragraph{Controlling $\E\brackets{ \| v_n(j, 1)\|_{\kernel}^2 | u_j }$} Applying \cite[Thm.~3.4]{muandet2017kernel}, we find that
\begin{align}
\|v_n(j, 1) \|_{\kernel}^2 = \mmd_\kernel^2( \what{\mu}_{j, 1}, \mu_{j, 1} ) \leq \frac{2\|\kernel\|_\infty}{n} + \frac{4\|\kernel \|_\infty\log(1/\delta_0)}{n}
    \label{eqn : mmd lln}
\end{align}
with probability at least $1 - \delta_0$   conditioned on $u_i, v_1$, where the randomness is taken over the measurements $X_{1:n}(j, 1)$. Note that for any pairs of distributions $(\mu, \nu)$, we have
\begin{align}
    \mmd_{\kernel}^2(\mu, \nu)&\leq \E_{X\sim\mu,X'\sim\mu}[\kernel(X, X')]
    + \E_{X\sim\nu,X'\sim\nu}[\kernel(X, X')]
    - 2\E_{X\sim\mu,X'\sim\nu}[\kernel(X, X')]\\
    &\leq 4\sinfnorm{\kernel}.
    \label{eq:mmd_bound}
\end{align}
Now choosing  $\delta_0 = n^{-1}$, we thus obtain 
\begin{align}
    \E\brackets{ \| v_n(j, 1) \|_\kernel^2 | v_1, u_j } \sless{\cref{eq:mmd_bound}} \frac{2\sinfnorm{\kernel} + 4\sinfnorm{\kernel}}{n} + \frac{4\sinfnorm{\kernel}\log n}{n} = 
    4\sinfnorm{\kernel} \frac{(1.5+\log n)}{n}.
\end{align}
So on the event $\overlapconc$,  we can also bound the first term from the RHS of \cref{eqn : conditional decomp} as follows:
\begin{align}
    &\frac{\bindic{ \sum_{j \in \mbf N_{1, \eta}} A_{j, 1} \geq 1 }}{(\sum_{j \in \mbf N_{1, \eta}} A_{j, 1})^2} \sum_{j \in \mbf N_{1, \eta}} A_{j, 1} \cdot \E\brackets{ \| v_n(j, 1)\|_{\kernel}^2 | u_j } \\
    &\leq \frac{\bindic{ \sum_{j \in \mbf N_{1, \eta}} A_{j, 1} \geq 1 }}{(\sum_{j \in \mbf N_{1, \eta}} A_{j, 1})} 4\sinfnorm{\kernel} \frac{(1+\log n)}{n} \\ 
    &\sless{\cref{eq:n_lbnd}} \frac{\bindic{ \sum_{j \in \lnbhdstarA} A_{j, 1} \geq 1 }}{(\sum_{j \in \lnbhdstarA} A_{j, 1})} 4\sinfnorm{\kernel} \frac{(1+\log n)}{n}.
    \label{eq:vbnd}
\end{align}

Note that assuming $\sum_{j \in \lnbhdstarA}A_{j, 1} > 0$ in \cref{event:good2} and the condition \cref{eq:a_sum} from the event $\overlapconc$ jointly induces $\sum_{j \in \mbf N_{1, \eta}}A_{j, 1} > 0$, on which the RHS of inequality \cref{eqn : conditional decomp} is well defined---this allow us to invoke \cref{lem : MMD decomp}.

\paragraph{Putting the pieces together}
Whenever $\mc V_{-1}, \mc D_{-1}$ satisfies $\overlapconc$, under \cref{event:good2}, we invoke \cref{lem : MMD decomp}. Then on the event $\overlapconc$, combine \cref{eq:bias_bnd,eq:vbnd,eq:last_term_bnd} together with the fact that $\Prob[\overlapconc \vert \mc U, \mc A] \geq 1-N\delta$. On the other hand, if $\mc V_{-1}, \mc D_{-1}$ does not satisfy $\overlapconc$, then we observed 
\begin{align}
    \E\brackets{ \| \what{\mu}_{1, 1, \eta} - \mu_{1, 1}\|_{\kernel}^2 | \mc V_{-1}, \mc D_{-1}, \mc A, \mc U } \leq 4\sinfnorm{\kernel}.
\end{align}
As a last step, marginalize over $\mc V_{-1}, \mc D_{-1}$ and we yield the desired bound of  \cref{prop:most-raw-bound}.

\subsection{Proof of \cref{lem : MMD decomp}: Conditional $\mmd$ error decomposition}
\label{proof : decomp}

We have
\begin{align}
\label{eq:mse_basic_decmp_lem1}
    &\E\brackets{ \| \what \mu_{1, 1, \eta} - \mu_{1, 1} \|_{\kernel}^2 | \wholerand} \\ 
    &\leq \bindic{\sum_{j \in \mbf N_{1, \eta}}A_{j, 1}  = 0 } \cdot 4\|\kernel\|_\infty \\
    &+ \bindic{ \sum_{j \in \mbf N_{1, \eta}}A_{j, 1}  \geq 1  }  \cdot \E\brackets{\| \what\mu_{1, 1, \eta} - \mu_{1,1}\|_{\kernel}^2| \wholerand},
\end{align}
where for the first term we have used the fact that $\knorm{\mu-\nu}^2\sless{\cref{eq:mmd_bound}} 4\sinfnorm{\kernel}$ for two arbitrary distributions $\mu$ and $\nu$. On the event $\bindic{ \sum_{j \in \mbf N_{1, \eta}}A_{j, 1}  \geq 1  }$, recalling the definitions~\cref{eq:bias_variance}, we can write 
\begin{align}\label{eq:embedding-decomp}
    \what\mu_{1, 1, \eta}\kernel - \mu_{1, 1}\kernel&= \frac{1}{| \mbf N_{1, \eta} |} \sum_{j \in \mbf N_{1, \eta}} \parenth{\empdistZ{j, 1}\kernel - \mu_{1, 1}\kernel} \\
    &=  \frac{1}{| \mbf N_{1, \eta} |} \sum_{j \in \mbf N_{1, \eta}} \parenth{ v_n(j, 1) + b(j, 1) }\\
    &= \frac{\sum_{j \in \mbf N_{1, \eta}} A_{j, 1} ( v_n(j, 1) + b(j, 1) ) }{\sum_{j \in \mbf N_{1, \eta}} A_{j, 1} }.
\end{align}
Note that by the bilinearity of inner product, i.e. for any $w_i \in \real$, $\alpha_i, \beta_i \in \mc H$ and index $i, i' \in \mc I$, we have
\begin{align}
    \angles{ \sum_{i \in \mc I} w_i \parenth{\alpha_i + \beta_i} , \sum_{i \in \mc I} w_i \parenth{\alpha_i + \beta_i} }_\kernel &= \sum_{i, i' \in \mc I} w_i w_{i'} \angles{\alpha_i + \beta_i, \alpha_{i'} + \beta_{i'}}_\kernel\\
    &= \sum_{i, i' \in \mc I} w_i w_{i'} \cdot\braces{\angles{\alpha_i, \alpha_{i'}}_\kernel + \angles{\beta_i, \beta_{i'}}_\kernel + 2 \angles{ \alpha_i, \beta_{i'} }_\kernel},
\end{align}
so that the squared MMD error can be expanded as follows:
\begin{align}
\label{eqn : decomposition}
     \| \what\mu_{1, 1, \eta} - \mu_{1, 1}\|_{\kernel}^2  &=  \angles{ \frac{\sum_{j \in \mbf N_{1, \eta}} A_{j, 1} ( v_n(j, 1) + b(j, 1) ) }{\sum_{j \in \mbf N_{1, \eta}} A_{j, 1} }, \frac{\sum_{j \in \mbf N_{1, \eta}} A_{j, 1} ( v_n(j, 1) + b(j, 1) ) }{\sum_{j \in \mbf N_{1, \eta}} A_{j, 1} } }_\kernel \\ 
     &= \frac{1}{( \sum_{j \in \mbf N_{1, \eta}}A_{j, 1} )^2} \sum_{j, m \in \mbf N_{1, \eta}} A_{j, 1}A_{m, 1}\langle v_n(j, 1), v_n(m, 1) \rangle_{\kernel} \\
    &+  \frac{1}{( \sum_{j \in \mbf N_{1, \eta}}A_{j, 1} )^2}\sum_{j, m \in \mbf N_{1, \eta}}  A_{j, 1}A_{m, 1}\langle b(j, 1), b(m, 1) \rangle_{\kernel}\\
    &+  \frac{2}{( \sum_{j \in \mbf N_{1, \eta}}A_{j, 1} )^2}\sum_{j, m \in \mbf N_{1, \eta}} A_{j, 1}A_{m, 1}\langle v_n(j, 1), b(m, 1) \rangle_{\kernel}. 
    \label{eq:lem1_basic_decomp}
\end{align}
We now bound the conditional expectation for each of the terms in the above display, one-by-one.
    \paragraph{Bound on $\angles{v_n(j, 1), v_{m}(j, 1)}_\kernel$}For $j \neq m$, we have
    \begin{align}
        &\E\brackets{ \langle v_n(j, 1) , v_n(m, 1) \rangle_{\kernel} | \wholerand }\\
        =&\E\brackets{ \langle v_n(j, 1), v_n(m, 1) \rangle_{\kernel} | u_j, u_m }\\
        =&\E\brackets{ \E  \brackets{  \langle v_n(j, 1) , v_n(m, 1) \rangle_{\kernel} | v_1, u_j, u_m }  } \\ 
        =&\E\brackets{  \langle  \E  \brackets{v_n(j, 1)\vert v_1, u_j} ,  \E  \brackets{v_n(m, 1)\vert v_1,u_m} \rangle_{\kernel} \vert u_j, u_m } \seq{\cref{eq:bias_variance_mean}} 0,
    \end{align}
    where second equality is by using independence of column latent factors $v_1 \indep \mc V_{-1}$.
    For $j=m$, we have
    \begin{align}
     \E\brackets{ \langle v_n(j, 1) , v_n(m, 1) \rangle_{\kernel} | \wholerand }
     = \E\brackets{ \knorm{ v_n(j, 1)}^2| u_j}.
    \end{align}
    As a result, we have
    \begin{align}
        &\frac{1}{( \sum_{j \in \mbf N_{1, \eta}}A_{j, 1} )^2} 
        \sum_{j, m \in \mbf N_{1, \eta}} A_{j, 1}A_{m, 1} \E[\langle v_n(j, 1), v_n(m, 1) \rangle_{\kernel} \vert \wholerand]
        \\
        &\quad= 
        \frac{1}{( \sum_{j \in \mbf N_{1, \eta}}A_{j, 1} )^2}
        \sum_{j \in \mbf N_{1, \eta}} A_{j, 1}A_{m, 1} \E[ \knorm{ v_n(j, 1)}^2 \vert u_j].
        \label{eq:ee_bnd}
    \end{align}
    
\paragraph{Bound on $\angles{b(j, 1), b(m, 1)}_\kernel$} 

Cauchy-Schwarz inequality yields that
 \begin{align}
     &\max_{j, m \in \mbf N_{1, \eta}} A_{j, 1}A_{m, 1}\E\brackets{ \| b(j, 1) \|_{\kernel} \| b(m, 1) \|_{\kernel} | \wholerand }\\
     \leq& \braces{\max_{j \in \mbf N_{1, \eta}} A_{j, 1} \sqrt{ \E\brackets{ \|b(j, 1)\|_\kernel^2 | \wholerand } } }^2\\
     =& \max_{j \in \mbf N_{1, \eta}}A_{j, 1} \E \brackets{ \| b(j, 1) \|_\kernel^2 | \wholerand}
     =  \max_{j \in \mbf N_{1, \eta}}A_{j, 1} \E \brackets{ \| g(u_j, v_1) - g(u_1, v_1)\|_{\kernel}^2 | u_1, u_j } 
 \end{align}
 Consequently, we have
 \begin{align}
     &\frac{1}{( \sum_{j \in \mbf N_{1, \eta}}A_{j, 1} )^2}\sum_{j, m \in \mbf N_{1, \eta}}  A_{j, 1}A_{m, 1} \E[\langle b(j, 1), b(m, 1) \rangle_{\kernel} \vert \wholerand] \\ 
     &\quad\leq \max_{j \in \mbf N_{1, \eta}}A_{j, 1} \E \brackets{ \| g(u_j, v_1) - g(u_1, v_1)\|_{\kernel}^2 | u_1, u_j }.
     \label{eq:bb_bnd}
 \end{align}
\paragraph{Bound on $\langle v_n(j, 1), b(m, 1) \rangle$}
We can mimic the reasoning used to control variance and bias terms to find that for any $j, m$, we have
\begin{align}
  &\E\brackets{ \langle v_n(j, 1), b(m, 1) \rangle_{\kernel} | \wholerand }\\
  &= \E\brackets{ \langle \E[v_n(j, 1) \vert b(m, 1), \wholerand], b(m, 1) \rangle_{\kernel} | \wholerand }
  \seq{(i)} 0,
  \label{eqn : cross product}.
\end{align}
where step~(i) follows from \cref{eq:bias_variance_mean}.
Consequently, we find that
\begin{align}
    \frac{2}{( \sum_{j \in \mbf N_{1, \eta}}A_{j, 1} )^2}\sum_{j, m \in \mbf N_{1, \eta}} A_{j, 1}A_{m, 1} \E[\langle v_n(j, 1), b(m, 1) \rangle_{\kernel} \vert\wholerand]
    = 0
    \label{eq:eb_bnd}
\end{align}

Collecting \cref{eq:mse_basic_decmp_lem1,eq:lem1_basic_decomp,eq:ee_bnd,eq:bb_bnd,eq:eb_bnd} yields the bound~\cref{eqn : conditional decomp} as claimed in \cref{lem : MMD decomp}.

\subsection{Proof of \cref{lem:conditional-concentration}: Conditional concentration for row metric}
\label{proof : concentration}

Conditioned on $\sbraces{\uarand}$, we have
\begin{align}
    \rho_{j, 1} = \sum_{s\neq 1} w_s \what{\mmd}_\kernel^2( \empdistZ{j, s}, \empdistZ{1, s})
    \qtext{where}w_s = \frac{A_{1,s}A_{j, s}}{\overoj}.
\end{align}
Note that $\what{\mmd}_\kernel^2$ is an unbiased estimator of $\mmd_{\kernel}^2$~\cite[Cor.~2.3]{borgwardt2006integrating}, i.e., for $s \neq 1$, we have
\begin{align}
    \E\brackets{\what{\mmd}_\kernel^2( \empdistZ{j, s}, \empdistZ{1, s} ) \vert u_j, u_1, v_s, A_{j,s}=1, A_{1, s}=1 } = \mmd_{\kernel}^2(\mu_{j, s}, \mu_{1, s}).
\end{align}
As a result, we find that 
\begin{align}
    \E\brackets{ \what{\mmd}_\kernel^2( \empdistZ{j, s}, \empdistZ{1, s} )|\uarand} = \E[\mmd_{\kernel}^2(\mu_{j, s}, \mu_{1, s}) \vert u_1, u_j] \seq{\eqref{def : Delta and etai},\eqref{eq:factor_model}} \gio, \quad \forall s \neq 1,
\end{align}
and further, in conjuction with the fact that $\sum_{s \neq 1}w_s = 1$, we have the identity 
\begin{align}
    \rho_{j, 1} - \gio = \sum_{s \neq 1}w_s \braces{\what{\mmd}_\kernel^2(\empdistZ{j, s}, \empdistZ{1, s}) - \gio }. 
\end{align}

Next we apply a sub-Gaussian concentration result~\cite[Thm.~2.6.2]{vershynin2018high}, on the centered dissimilarity measure $\rho_{j, 1} - \gio$, which requires (i) the control of the $\psi_2$-Orlicz norm of each of its summands, and (ii) independence across these summands. 

Accordingly, we claim that
\begin{align}
\left\|\what{\mmd}^2_\kernel(\empdistZ{j, s}, \empdistZ{1, s}) - \gio \right\|_{\psi_2}  \leq \frac{8 \|\kernel\|_\infty}{\sqrt{\log 2}},
    \label{eq:mmd_psinorm}
\end{align}
by utilizing the fact that any random variable $X$ satisfies $\| X \|_{\psi_2} \leq \sinfnorm{X}/\sqrt{\log 2}$ whenever its supremum norm $\sinfnorm{X}$ is bounded~\cite[Ex.~2.5.8]{vershynin2018high}. To show \cref{eq:mmd_psinorm}, we first
observe the inequality
$\sinfnorm{ \what{\mmd}^2(\empdistZ{j, s}, \empdistZ{1, s}) } \leq 4\sinfnorm{\kernel}$ follows directly from \cref{eq:mmd_bound}. Second, observe the following inequality,
\begin{align}
    \gio \leq \int 2 \| g(u_j, v) \|_{\kernel}^2 + 2 \| g(u_1, v)\|_{\kernel}^2 d\P_{v},
    \label{eq:gio_ineq}
\end{align}
by triangle inequality and the inequality $(a + b)^2 \leq 2a^2 + 2b^2$ that holds for any $a, b \in \real$. Combining \cref{eq:gio_ineq} with the following inequality,
\begin{align}
\|g(u_i, v_t)\|_{\kernel}^2 = \angles{\mu_{i, t}\kernel, \mu_{i, t}\kernel}_\kernel = \iint k(x, y) d\mu_{i, t}(x) d\mu_{i, t}(y)\leq \| \kernel\|_\infty, 
\end{align}
we attain $\sinfnorm{\gio} \leq 4\sinfnorm{\kernel}$.
Lastly, the following triangle inequality completes \cref{eq:mmd_psinorm},
\begin{align}
    \left\|\what{\mmd}^2_\kernel(\empdistZ{j, s}, \empdistZ{1, s}) - \gio \right\|_{\infty} \leq \sinfnorm{\what{\mmd}^2_\kernel(\empdistZ{j, s}, \empdistZ{1, s})} + \sinfnorm{\gio} \leq 8 \sinfnorm{\kernel}.
\end{align}

Another ingredient for sub-Gaussian concentration is the $\sbraces{\uarand}$-conditional independence of the following terms across $s \neq 1$,
\begin{align}
    W_{j, s} \defeq w_s \braces{\what{\mmd}^2_\kernel(\empdistZ{j, s}, \empdistZ{1, s}) - \gio }.
\end{align}
It is sufficient to check independence of $\what{\mmd}_\kernel^2(\empdistZ{j, s}, \empdistZ{1, s})$ across $s \neq 1$, as $w_s$ are constant conditioned on $\mc A$ and $\gio$ are constant conditioned on $\mc U$. The exogenous nature of $\mc U$, and the independence across column latent factors in \cref{assump:latent-independence}, along with conditional independence of $\mc A$ in \cref{assump : cond_ind_a} yields conditional independence we desire. Equipped with conditional independence, and $\psi_2$-Orlicz norm bound in \cref{eq:mmd_psinorm}, we can apply sub-Gaussian concentration~\cite[Thm.~2.6.2]{vershynin2018high} on $\rho_{j, 1} - \gio$, yielding,
\begin{align}
    \Prob\braces{ \Big|\sum_{s \neq 1} W_{j, s} \Big| > \frac{c_0\|\kernel\|_\infty \sqrt{\log(2/\delta)}}{\sqrt{ \overoj }} \Big| \uarand } \leq \delta
\end{align}
for any $\delta > 0$. Note that the constant $c_0$ does not depend on $\mc U, \mc A$ or index $j$.

\section{Proof of Thm. \ref{thm:stagger-bound}: Staggered adoption guarantee}
\label{proof : staggered adoption}

Under the staggered adoption assumption \cref{assump : confounded stagger}, the adoption times are conditionally independent to the column latent factors given the row latent factors. Since adoption times determine the treatment assignments in this setting, \cref{assump : unobs confounding} here as well. We therefore apply \cref{prop:most-raw-bound} to the control outcome matrix, with target entry $(1, T)$ in place of $(1, 1)$.

For the remainder of this proof, we follow the notation of \cref{prop:most-raw-bound} and write $A_{j, s} = \indic{s \leq \tau_j}$ for the indicator that the control outcome for unit $j$ is observed at time $s$. So the number of periods before $T$ where units $1$ and $j$ both have observed control outcomes is 
\begin{align}
    \sum_{s \neq T} A_{1, s}A_{j, s} = 
    \adtime[1] \wedge \adtime[j] \wedge (T - 1).
\end{align}
According to the local definition of the indicator $A_{j, s}$, at time $T$, $A_{j, T}=1$ if and only if the unit $j \in \neverad$ and for all these units
$A_{j, s} = 1$ for all $s \leq T$. Consequently, for any $j \in \neverad$, we have
\begin{align}
    \sum_{s\neq T} A_{1, s}A_{j, s} = \tau_1 \wedge (T - 1)
    \qtext{and}
    \errja = \frac{c_0\| \kernel\|_\infty \log(2N/\delta) }{\sqrt{\adtime[1] \wedge (T - 1) }}.
    \label{eq:err_never_ad}
\end{align}
Recalling the definition~\cref{eq:err_defn} of $\errja$, we find that
\begin{align}
    \max_{j \in \unbhdstarA}A_{j, T}\errja[j, \mc A] \leq \max_{j \in \neverad}\errja[j, \mc A]  \seq{\cref{eq:err_never_ad}} \frac{c_0\| \kernel\|_\infty \log(2N/\delta) }{\sqrt{\adtime[1] \wedge (T - 1) }}.
    \label{eq:err_ub}
\end{align}
Next, using the definition~\cref{def : pop nbhd data centered} of $\lnbhdstarA$, we find that
\begin{align}
    \sum_{j \in \lnbhdstarA} A_{j, T} &\geq | \sbraces{ j \in \neverad : \gio < \eta - \errja[j, \mc A] }  | \\ &\seq{\cref{eq:err_never_ad}}| \sbraces{ j \in \neverad : \gio < \eta - \frac{c_0\| \kernel\|_\infty \log(2N/\delta) }{\sqrt{\adtime[1] \wedge (T - 1) }}}  | \\
    &\seq{(i)} \abss{\nbhdstar}
    \label{eq:nb_ub}
\end{align}
where step~(i) follows from the definition of $\nbhdstar$ stated in the statement of \cref{thm:stagger-bound}. 

Finally, invoking \cref{prop:most-raw-bound} and putting it together with \cref{eq:err_ub,eq:nb_ub} we find that
\begin{align}
    \E \brackets{\| \what{\mu}_{1, T, \eta} - \mu_{1, T}\|_\kernel^2 \vert\mc V_{-1}, \mc D_{-1}, \mc U, \mc A }
    &\leq \eta + \max_{j \in \unbhdstarA}\!\!\! 
    A_{j, T}\cdot \errja 
    +  \frac{4\| \kernel\|_\infty (\log n + 1.5)}{ n \sum_{j \in \lnbhdstarA} A_{j, T} },
     \\ 
     &\leq \eta + \frac{c_0\| \kernel\|_\infty \log(2N/\delta) }{\sqrt{\adtime[1] \wedge (T - 1) }} + 
     \frac{4\| \kernel\|_\infty (\log n + 1.5)}{n\abss{\nbhdstar}}
\end{align}
as claimed. Lastly marginalize with respect to $\mc V_{-1}$ and $\mc D_{-1}$ and the proof is complete.

\subsection{Kernel Treatment Effect}
\label{app:dte-proof}

Here we give a formal discussion on the estimation of kernel treatment effects~\cref{eq:dte}, that is specific to the staggered adoption setting in \cref{sec : staggered adoption}. We introduce our proposed estimator for learning $\dte_{1, T} = \snorm{ \mu_{1, T}^{(1)} - \mu_{1, T}^{(0)} }_\kernel$, and introduce additional structural assumptions that make analysis feasible. We emphasize that the framework, estimator, and guarantees provided in this section can be easily extended to the more general potential outcome framework of \cref{mod:potential-outcome}.

\paragraph{Proposed estimator for $\dte_{1, T}$}

Fix entry $(1, T)$ and radii $\eta_0, \eta_1 > 0$. Available observations are the missingness $\mc A$, and  measurements $\sbraces{Z_{i, t}}_{(i, t)\in [N]\times [T]}$ from \cref{mod:potential-outcome staggered}. Then implement the general version of $\kerNN$~(see \cref{sec : fully general kerNN}) in the following way:
\begin{enumerate}[leftmargin = 0.7cm]
    \item[(1)] Construct estimators $\what{\mu}^{(1)}_{1, T, \eta_1}, \what{\mu}^{(0)}_{1, T, \eta_0}$ for distributions $\mu_{1, T}^{(1)}$ and $\mu_{1, T}^{(0)}$ respectively through
    \begin{align}
\begin{cases}
     \text{Apply $\kerNN$ with $\eta = \eta_1, a = 1$} &{\Longrightarrow} \quad  \what{\mu}^{(1)}_{1, T, \eta_1},\\
    \text{Apply $\kerNN$ with $\eta = \eta_0, a = 0$} & {\Longrightarrow} \quad \what{\mu}^{(0)}_{1, T, \eta_0}.
\end{cases}
\end{align}
    \item[(2)] Calculate $\what{\dte}_{1, T, \eta} = \| \what{\mu}^{(1)}_{1, T, \eta_1} - \what{\mu}^{(0)}_{1, T, \eta_0} \|_{\kernel}$, where $\eta = (\eta_0, \eta_1)$.
\end{enumerate}

We emphasize $\what{\dte}_{1, T, \eta}$ is computable from data due to linearity of inner product $\langle \cdot, \cdot \rangle_\kernel$ and the mixture expression of $\kerNN$. Also, we propose to tune radii $\eta_0, \eta_1$ separately --- practically, do grid search~(see \cref{app:sim}) for $\eta_0, \eta_1$ separately, and theoretically, apply the reasoning of \cref{cor:stagger-bound} separately to get two different optimal values $\eta_0^\star, \eta_1^\star$.

\paragraph{Data generating process}
Measurements $\sbraces{Z_{i, t}}_{(i, t) \in [N]\times [T]}$ of model \cref{mod:potential-outcome staggered} are generated through the following process,
\begin{enumerate}[leftmargin = 0.7cm]
    \item[(1)] Row latent factors $\mc U = \braces{u_1, ..., u_N}$ are generated i.i.d. from compact hypercube $[-1, 1]^r$, and two separate column latent factors are generated --- for $q = 0, 1$, column latent factors $\sbraces{ v_1^{(q)}, ..., v_T^{(q)} } = \mc V^{(q)}$ are both generated i.i.d. uniformly from a compact space $[-1, 1]^r$ and $\mc V^{(0)} \indep \mc V^{(1)}$ hold. This latent factor generation is analogous to \cref{assump:latent-independence}. 
    \item[(2)] Next, for each entry $(i, t)$, we assign two different distributions. For fixed $u_i, v_t^{(0)}, v_t^{(1)}$, define distributions $\mu_{i, t}^{(q)}, q = 0, 1,$ so that embedding factorization holds, i.e. $\mu_{i, t}^{(q)}\kernel = g^{(q)}(u_i, v_t^{(q)})$ for some non-parametric functions $g^{(q)}, q = 0, 1$. This is analogous to \cref{assump:factorization}.
    \item[(3)] Lastly, given treatment assignment $\mc A$ were generated according to \cref{assump : confounded stagger}, generate measurements $X_1^{(q)}(i, t), ..., X_n^{(q)}(i, t)$ whenever $A_{i, t} = q$. This step is analogous to \cref{assump : measurement generation}.
\end{enumerate}
It is possible to make two~(indexed by $q \in \{ 0, 1 \}$) separate distributional matrix completion models \cref{model : dist matrix completion} from the observations generated in this section,
\begin{align}
 \text{ for}\ \ i \in [N], t\in[T],\quad
    \begin{cases}
        [X_{1}^{(q)}(i, t), \ldots, X_{n}^{(q)}(i, t)]&\qtext{if} A_{i, t}=q, \\
        \mrm{unknown}&\qtext{if} A_{i, t}= 1 - q.
    \end{cases}
    \label{model : counterfactual double}
\end{align}

\subsection{Proof of \cref{cor:dte-stagger}}
\label{proof:dte-stagger}
Take $r=2$, $L=4d$, and $\sinfnorm{\kernel}=(d+1)^2$, and use the parameter choices in \cref{eq:stagger_complete_parameters}. Require $N-1-N^\alpha>0$ and $T-1-T^{1-\beta}>0$. Choose $\eta_0^\star$ by \cref{eq:stagger_complete_radius}, and choose $\eta_1^\star$ by the same formula with $(N^\alpha, T^\beta)$ substituted by $(N-1-N^\alpha, T-1-T^{1-\beta})$. Require \cref{eq:stagger_complete_coverage} both without and with these substitutions. 
For any pair $\eta = (\eta_0, \eta_1)$, the triangle inequality and $(a+b)^2 \leq 2 a^2 + 2 b^2$ yield
\begin{align}
    (\what{\dte}_\eta - \dte)^2 & \leq 2\| \what{\mu}^{(1)}_{1, T, \eta_1} - \mu_{1, T}^{(1)} \|_{\kernel}^2 + 2\| \what{\mu}^{(0)}_{1, T, \eta_0} - \mu_{1, T}^{(0)}\|_{\kernel}^2,
\end{align}
so that we have
\begin{align}
    \E \brackets{ ( \what{\dte}_\eta - \dte )^2 } &\leq 2\E\brackets{ \| \what{\mu}^{(1)}_{1, T, \eta_1} - \mu^{(1)}_{1, T}\|_\kernel^2 } + 2\E\brackets{ \| \what{\mu}^{(0)}_{1, T, \eta_0} - \mu^{(0)}_{1, T}\|^2_\kernel}.
\end{align}

Corollary~\ref{cor:stagger-bound} (a) gives the control outcome bound. For the treatment outcome, the $N-1-N^\alpha$ adopters other than unit $1$ form a fixed candidate pool. Each candidate $j$ and unit $1$ are both treated from period $\max\{\tau_1,\tau_j\}+1$ through $T-1$, giving at least $T-1-T^{1-\beta}$ shared periods.

With this candidate pool count and overlap, Proposition~\ref{prop:most-raw-bound} and the argument of Appendix~\ref{proof:stagger_complete} give the treatment-arm bound. Since $N-1-N^\alpha\asymp N$ and $T-1-T^{1-\beta}\asymp T$, substituting both arm bounds into the preceding inequality gives
\begin{align}
    \E \brackets{ ( \what{\dte}_{\eta^\star} - \dte )^2 } \leq \tilde{O}\biggbrackets{ \frac{d^2}{\sqrt{n \cdot N^{(1 - \alpha) \wedge \alpha}}} + \frac{d^2}{\sqrt{ T^{(1 - \beta) \wedge \beta}}}}, 
\end{align}
as claimed.

\subsection{\kerNN\ for potential outcome setting}
\label{sec : fully general kerNN}

For the setting with potential outcomes (under finitely many interventions $a \in \sbraces{0, 1, \ldots, K-1}$), we can generalize the \kerNN algorithm by redefining the notation for the observed distribution for unit $i$ for outcome $t$ and intervention $a$ as follows:
\begin{align}
    \empdistnotag^{(Z, a)}_{j,s} \defeq 
    \begin{cases}
        \frac{1}{n} \sum_{\ell = 1}^{n} \dirac_{X^{(a)}_\ell(j, s)} & A_{j, s} = a\\
        \trm{unobserved} &\trm{otherwise}
    \end{cases},
\end{align}
Next, we define intervention-specific neighborhood via
\begin{align}
    \rho_{i,j}^{(a)} &\defeq \frac{\sum_{s \neq t} \indic{A_{i, s} = a} \indic{A_{j,s} = a }\what{\mmd}^2_{\kernel}(\empdistnotag^{(Z, a)}_{i,s},\empdistnotag^{(Z, a)}_{j,s})  }{\sum_{s \neq t} \indic{A_{i,s} = a}\indic{A_{j,s} = a}},
\end{align}
so that the \kerNN-estimate for  ${\mu}^{(a)}_{i, t, \eta}$ is given by
\begin{align}
\what{\mu}^{(a)}_{i, t, \eta} 
&\defeq 
\frac{\sum_{j \in \mbf N_{i,\eta}^{(a)}} \indic{A_{j, t} = a} \empdistnotag^{(Z, a)}_{j,t}}{\sum_{j \in \mbf N_{i,\eta}^{(a)}} \indic{A_{j,t} = a} } 
\qtext{where} 
\mbf N_{i, \eta}^{(a)} \defeq \braces{ j \in [N]\setminus\{i\} : \rho_{i,j}^{(a)} \leq \eta}.
\end{align}

\section{Proof of Thm. \ref{thm:prop-bound}: Propensity-based guarantee}
\label{proof:thm_propensity}

\newcommand{\totalconc}{\mc E_{\trm{total-conc}}}
Without loss of generality, we assume that $\mc U$ and $\eta > 0$ are such that
\begin{align}
    &\sum_{s \neq 1} p_{1, s}p_{j, s} > 0
    \quad\text{for every }j\neq1\text{ with }p_{j,1}>0, \qtext{and}\sum_{j \in \lnbhdstarp}p_{j, 1} > 0,
    \label{event:good3}
\end{align}
because otherwise the bound derived in \cref{thm:prop-bound} is vacuous.
Now, define the following two events regarding concentration of missingness around its propensities:
\begin{align}
    \nbhdconc &\defeq \biggbraces{ \sum_{j \in \lnbhdstarp} A_{j, 1} \geq \frac{1}{2}\sum_{j \in \lnbhdstarp} p_{j, 1} } \quad \text{and} \label{event:nbhd-conc} \\
    \overlapconcp &\defeq \biggbraces{ \sum_{s \neq 1}A_{1, s}A_{j, s} \geq \frac{1}{2}\sum_{s \neq 1}p_{1, s}p_{j, s} , \ \text{for all $j\neq1$ with $A_{j, 1} = 1$} }\label{event:ov-conc}.
\end{align}
Using \cref{assump : cond_ind_a} and the fact that $\lnbhdstarp$ and $p_{j, s}$ are functions of $\mc U$, we apply Binomial-Chernoff concentration~\cite[Lem.~A.2]{dwivedi2022counterfactual}, to attain the following proabaility bounds of the events,
\begin{align}
    \Prob\biggbraces{ \sum_{j \in \lnbhdstarp}A_{j, 1} &< \frac{1}{2} \sum_{j \in \lnbhdstarp}p_{j, 1} \Big| \mc U } \leq \exp\biggbraces{-\frac{1}{8} \sum_{j \in \lnbhdstarp}p_{j, 1} } \quad \text{and} \\
    \Prob\biggbraces{ \sum_{s\neq 1}A_{1, s}A_{j, s} &< \frac{1}{2} \sum_{s \neq 1}p_{1, s}p_{j, s} \Big| \mc U } \leq \exp\biggbraces{-\frac{1}{8} \sum_{s\neq 1}p_{1, s}p_{j, s} }. \label{eq:twoevent_conc}
\end{align}
The two probability bounds in \cref{eq:twoevent_conc} results in the following probability lower bound for the two events \cref{event:nbhd-conc,event:ov-conc},
\begin{align}
    \Prob\sbraces{ \nbhdconc  | \mc U} &\geq  1 - \exp\biggbraces{-\frac{1}{8} \sum_{j \in \lnbhdstarp}p_{j, 1} } \quad \text{and} \\
    \Prob\sbraces{ \overlapconcp  | \mc U} &\geq 1 - \sum_{j\neq1:\,p_{j,1}>0}\exp\biggbraces{-\frac{1}{8} \sum_{s\neq 1}p_{1, s}p_{j, s} }.
    \label{eq:prop_highprob}
\end{align}
Note that \cref{eq:prop_highprob} uses that fact that conditional on $\mc U$, $A_{j,1}=1$ implies $p_{j,1}>0$ almost surely, which means conditioned on $\mc U$
\begin{align}
    \overlapconcp^c &= \bigcup_{j \neq 1, A_{j, 1}=1} \biggbraces{ \sum_{s \neq 1}A_{1, s}A_{j, s} < \frac{1}{2}\sum_{s \neq 1}p_{1, s}p_{j, s}} \\
    &\subseteq \bigcup_{j \neq 1, p_{j, 1} > 0} \biggbraces{ \sum_{s \neq 1}A_{1, s}A_{j, s} < \frac{1}{2}\sum_{s \neq 1}p_{1, s}p_{j, s}}.
\end{align}

Next, on the events $\nbhdconc$ and $\overlapconcp$, we establish bounds on the individual terms appearing in the RHS of \cref{eq:most-raw-guarantee}. Observe that on the event $\overlapconcp$, we have for $j\neq1$ with $A_{j,1}=1$
\begin{align}
    \frac{A_{j, 1}\cdot c_0\sinfnorm{\kernel}\sqrt{\log(2N/\delta)}}{\sqrt{\sum_{s \neq 1} A_{1, s}A_{j, s} }} \leq \frac{A_{j, 1}\cdot c_0\sinfnorm{\kernel} \sqrt{2\log(2N/\delta)}}{\sqrt{\sum_{s \neq 1} p_{1, s}p_{j, s} }},
    \label{eq:radius-change}
\end{align}
from which we can deduce the following two set inclusions,
\begin{align}
    \lnbhdstarp\cap\braces{j:A_{j,1}=1}
    &\subseteq \lnbhdstarA \quad \text{and}
    \nonumber\\
    \unbhdstarA\cap\braces{j:A_{j,1}=1}
    &\subseteq \unbhdstarp \quad \text{on the event $\overlapconcp$},
    \label{eq:pop-neighbor-inclusion}
\end{align}
where $(\lnbhdstarA, \unbhdstarA)$ was defined in \cref{def : pop nbhd data centered} and $(\lnbhdstarp,\unbhdstarp)$ defined in \cref{eq:pop_neighbor_p}. One immediate consequence of the second set inclusion of \cref{eq:pop-neighbor-inclusion} is a bound on the second term of \cref{eq:most-raw-guarantee}, which is 
\begin{align}
    &\max_{\substack{j \in \unbhdstarA\\A_{j,1}=1}} \frac{ A_{j, 1} \cdot c_0 \sinfnorm{\kernel}\sqrt{\log(2N/\delta)}}{\sqrt{\sum_{s \neq 1}A_{1, s}A_{j, s}}}\\
    &\stackrel{\cref{event:ov-conc},\cref{eq:pop-neighbor-inclusion}}{\leq} \max_{\substack{j \in \unbhdstarp\\p_{j,1}>0}} \frac{ c_0 \sinfnorm{\kernel}\sqrt{2\log(2N/\delta)}}{\sqrt{\sum_{s \neq 1} p_{1, s}p_{j, s} }}, \quad\text{on the event $\overlapconcp\cap\nbhdconc$.}
    \label{eq:ineq_bias}
\end{align}

Also, we can deduce the following inequality,
\begin{align}
    \frac{4\sinfnorm{\kernel}(\log n + 1.5)}{n \sum_{j \in \lnbhdstarp}A_{j, 1}} \leq \frac{8\sinfnorm{\kernel}(\log n + 1.5)}{n \sum_{j \in \lnbhdstarp}p_{j, 1}} \quad \text{on the event $\nbhdconc$},
    \label{eq:ineq_nbhd-conc}
\end{align}
and by additionally applying the first set inclusion of \cref{eq:pop-neighbor-inclusion}, we get a bound on the third term of the RHS of \cref{eq:most-raw-guarantee}, which is
\begin{align}
    \frac{4\sinfnorm{\kernel}(\log n + 1.5)}{n \sum_{j \in \lnbhdstarA}A_{j, 1}} \stackrel{\cref{eq:pop-neighbor-inclusion},\cref{eq:ineq_nbhd-conc}}{\leq} \frac{8\sinfnorm{\kernel}(\log n + 1.5)}{n \sum_{j \in \lnbhdstarp}p_{j, 1}}, \quad \text{on the event $\overlapconcp\cap\nbhdconc$}.
    \label{eq:ineq_var}
\end{align}

Note that the new bounds established in \cref{eq:ineq_bias,eq:ineq_var} are well defined since we assume values $\mc U$ and $\eta$ to satisfy \cref{event:good3}. Further, by operating on the event $\overlapconcp \cap \nbhdconc$, the condition \cref{event:good2} that is necessary to invoke \cref{prop:most-raw-bound} is satisfied. Specifically, the first condition of \cref{event:good2} is derived using the first condition of \cref{event:good3} along with the definition of \cref{event:ov-conc}:
\begin{align}
 \text{for $j\neq1$ with $A_{j, 1}= 1$,} \quad 0 \stackrel{\cref{event:good3}}{<} \sum_{s \neq 1} p_{1, s}p_{j, s} \stackrel{\cref{event:ov-conc}}{\leq} 2 \sum_{s \neq 1}A_{1, s}A_{j, s}.
\end{align}
The second condition of \cref{event:good2} is derived using the second condition of \cref{event:good3} along with the definition of \cref{event:nbhd-conc}, as well as the set inclusion established in \cref{eq:pop-neighbor-inclusion}:
\begin{align}
 0\stackrel{\cref{event:good3}}{<} \sum_{j \in \lnbhdstarp}p_{j, 1} \stackrel{\cref{event:nbhd-conc}}{\leq}2 \sum_{j \in \lnbhdstarp} A_{j, 1} \stackrel{\cref{eq:pop-neighbor-inclusion}}{\leq}2\sum_{j \in \lnbhdstarA}A_{j, 1}   .
\end{align}

\paragraph{Putting the pieces together}
Now invoke the bound from \cref{prop:most-raw-bound} with the same $\delta$ and marginalize over $\mc A$ conditional on $\mc U$, using the event $\totalconc \defeq \overlapconcp \cap \nbhdconc$. On this event, combining \cref{eq:ineq_bias} and \cref{eq:ineq_var} bounds the bias and variance terms.
\begingroup
On $\totalconc^c$, use the uniform loss bound $4\sinfnorm{\kernel}$ from \cref{eq:stagger_complete_uniform_loss}. Including the $4\sinfnorm{\kernel}\delta$ term already in \cref{prop:most-raw-bound}, the total remainder is at most
\begin{align}
    4\sinfnorm{\kernel}\big[\delta+\Prob\braces{\overlapconcp^c\mid\mc U}
        +\Prob\braces{\nbhdconc^c\mid\mc U}\big]
    \leq\sinfnorm{\kernel}\mbf r_\delta,
    \nonumber
\end{align}
by \cref{eq:prop_highprob}. This
\endgroup
yields the claimed bound \cref{eq:prop-bound} of \cref{thm:prop-bound}.

\section{Proof of Cor. \ref{cor:stagger-bound}: Guarantees for specific examples under staggered adoption}
\label{proof:stag_adopt_opt}
\label{proof:stagger_complete}

We set $\delta = (nNT)^{-2}$ in \cref{thm:stagger-bound}, including its neighborhood definition. Next, equipped with the lower bound on adoption times, we claim that the guarantee of \cref{thm:stagger-bound} can be integrated to
\begin{align}
        &\E \bigbrackets{\| \what{\mu}^{(0)}_{1, T, \eta} - \mu_{1, T}^{(0)}\|_\kernel^2 \big| \mc U}
        \nonumber\\
        &\quad\leq \eta+ \frac{c_0\sinfnorm{\kernel}\log(2N/\delta)}{\sqrt{T^{\beta}}}
        +\frac{4\sinfnorm{\kernel}(\log n+1.5)}{n|\lnbhdstar|}
        +4\sinfnorm{\kernel}\delta, \label{eq:fully_int_stag}
\end{align}
where $\lnbhdstar \defeq \sbraces{ j \in \neverad : \gio < \eta - c_0 \sinfnorm{\kernel}\log(2N/\delta)/\sqrt{T^\beta} }$. This claim applies when $|\lnbhdstar|>0$; we defer its proof to the end of this section. Under the default convention of \cref{sec:algo}, the estimate is a probability measure or the zero measure, including when no observed control neighbors are available. For any probability measures $P,Q$, also allowing either to be the zero measure,
\begin{align}
    \knorm{P-Q}^2\leq(\knorm{P}+\knorm{Q})^2\leq4\sinfnorm{\kernel},
    \label{eq:stagger_complete_uniform_loss}
\end{align}
since $\knorm{P}^2\leq\sinfnorm{\kernel}$. Thus the loss remains bounded when $|\lnbhdstar|=0$.

Next, we use the following lemma (proof in \cref{sub:proof_of_lem_exam_lip}) to lower bound the number of neighbors:

\begingroup
\renewcommand{\theHmylemma}{\thesection.\arabic{mylemma}}
\begin{lemma}\label{lem:exam_lip}\label{lem:stagger_complete_count}
Suppose the latent factors $\mc U, \mc V$ are drawn \iid from the uniform distribution on $[-1, 1]^r$ and the function $g : [-1, 1]^r \times [-1, 1]^r \to \mc H$ in \cref{assump:factorization} is $L$-Lipschitz, with $L>0$, in the following sense:
\begin{align}
    \knorm{ g(u, v) - g(u', v') } \leq L \sbraces{ \twonorm{u - u'} \vee\twonorm{v - v'}}.
    \label{eq:lip}
\end{align}
Fix $u_1\in[-1,1]^r$, a set $\mc I\subset[N]\setminus\{1\}$ chosen before the unit latent factors are drawn, and $0<\eta'\leq L^2$.
\newcommand{\phiprob}[1][\eta']{\Phi_{#1}}
Then, over the randomness in $u_2, \ldots, u_{N}$, we have
\begin{align}
    \P\biggbraces{\abss{\sbraces{ j \in \mc I: \gio < \eta'}}
    \geq \frac12|\mc I|\cdot \phiprob~\Big\vert~ u_1}
    \geq 1 - e^{- |\mc I| \cdot \phiprob / 8}
    \stext{where}
    \phiprob
    \defeq \frac{(\sqrt{\pi\eta'}/4L)^r }{\Gamma(r/2+1)}.
    \label{eq:phi_lower_stag}
\end{align}
Moreover, we may take $r=2$, $L=4d$, $\sinfnorm{\kernel}=(d+1)^2$ for \cref{exam:gaussian_family}, and $L=\sqrt{\sum_{k=1}^\infty L_k^2}\in(0,\infty)$, $\sinfnorm{\kernel}=1$ for \cref{exam:infinite_family}, with its latent dimension $r$.
\end{lemma}
\endgroup
Write $c_r=\pi^{r/2}/(4^r\Gamma(r/2+1))$, so that $\Phi_{\eta'}=c_r(\eta')^{r/2}/L^r$. Choosing $\mc I=\neverad$ and $\eta'=\eta-c_0\sinfnorm{\kernel}\log(2N/\delta)/\sqrt{T^\beta}$, with $0<\eta'\leq L^2$, we apply the lemma to the fixed set of $N^\alpha$ never-adopters in \cref{cor:stagger-bound}. On the event $|\lnbhdstar|\geq N^\alpha\Phi_{\eta'}/2$, use \cref{eq:fully_int_stag}; on its complement, use \cref{eq:stagger_complete_uniform_loss}. Averaging over $u_2,\ldots,u_N$ conditional on $u_1$, we find that
\begin{align}
     &\E \bigbrackets{\| \what{\mu}^{(0)}_{1, T, \eta} - \mu_{1, T}^{(0)}\|_\kernel^2 \big| u_1}
     \nonumber\\
     &\leq \eta+\frac{c_0\sinfnorm{\kernel}\log(2N/\delta)}{\sqrt{T^\beta}}
     +\frac{8c_r^{-1}\sinfnorm{\kernel}L^r(1.5+\log n)}{nN^\alpha(\eta')^{r/2}}
     +4\sinfnorm{\kernel}\left(\delta+e^{-N^\alpha\Phi_{\eta'}/8}\right).
     \label{eq:temp_Bnd}
\end{align}
Since $\eta=\eta'+c_0\sinfnorm{\kernel}\log(2N/\delta)/\sqrt{T^\beta}$, balancing the radius-dependent terms in \cref{eq:temp_Bnd}, up to constants depending only on $r$, gives the following critical equality:
\begin{align}
    \eta' = \frac{\sinfnorm{\kernel}L^r(1.5+\log n)}{nN^\alpha(\eta')^{r/2}}
    \implies \eta'=\parenth{\frac{\sinfnorm{\kernel}L^r(1.5+\log n)}{nN^\alpha}}^{\frac{2}{r+2}}.
    \label{eq:eta_star}
\end{align}
Accordingly, to retain this scale while accounting for the error in estimating row similarities, we choose
\begin{align}
    \eta^\star=\frac{c_0\sinfnorm{\kernel}\log(2N/\delta)}{\sqrt{T^\beta}}
        +\parenth{\frac{\sinfnorm{\kernel}L^r(1.5+\log n)}{nN^\alpha}}^{\frac{2}{r+2}}.
    \label{eq:stagger_complete_radius}
\end{align}
The parameter conditions assumed in \cref{cor:stagger-bound} for this choice are
\begin{align}
    n,N,T\geq2,\qquad \alpha,\beta\in(0,1),\qquad
    \delta=(nNT)^{-2},\qquad C_r\geq16/c_r,
    \label{eq:stagger_complete_parameters}
\end{align}
and
\begin{align}
    L^2\left(\frac{C_r\log(nNT)}{N^\alpha}\right)^{2/r}
    &\leq\parenth{\frac{\sinfnorm{\kernel}L^r(1.5+\log n)}{nN^\alpha}}^{\frac{2}{r+2}}
    \leq L^2.
    \label{eq:stagger_complete_coverage}
\end{align}
Here $n,N,T$ and the dimensions are integers, and $N^\alpha$ is the size of the fixed never-adopter set. The middle expression in \cref{eq:stagger_complete_coverage} is $\eta'$ at $\eta=\eta^\star$. The upper inequality ensures that \cref{lem:exam_lip} applies, while the lower inequality gives $N^\alpha\Phi_{\eta'}\geq c_rC_r\log(nNT)\geq16\log(nNT)$. Consequently, $e^{-N^\alpha\Phi_{\eta'}/8}\leq\delta$, and the remainder in \cref{eq:temp_Bnd} is $O(\sinfnorm{\kernel}\delta)$, which is bounded by $O(\sinfnorm{\kernel}/\sqrt{T^\beta})$.

Substituting \cref{eq:stagger_complete_radius} into \cref{eq:temp_Bnd} and averaging over $u_1$ gives an inequality up to a constant depending only on $r$ 
\begin{align}
    &\E\bigbrackets{\|\what\mu^{(0)}_{1,T,\eta^\star}-\mu^{(0)}_{1,T}\|_\kernel^2}
    \leq \tilde O \bigg[\parenth{\frac{\sinfnorm{\kernel}L^r}{nN^\alpha}}^{\frac{2}{r+2}}
    +\frac{\sinfnorm{\kernel}}{\sqrt{T^\beta}}\bigg].
    \label{eq:stagger_complete_rate}
\end{align}
Now substituting the scalings of $L$ and $\sinfnorm{\kernel}$ from \cref{lem:exam_lip} for \cref{exam:gaussian_family,exam:infinite_family} yields the claimed bounds. 

\paragraph{Proof of claim \cref{eq:fully_int_stag}}

Use $\delta=(nNT)^{-2}$ in \cref{thm:stagger-bound}, and first retain the conditioning on $(\mc U,\adtimes)$. We prove the claim for $|\lnbhdstar|>0$; the empty-neighborhood case was handled separately above.

For every $j\in\neverad$, units $1$ and $j$ have $\tau_1$ jointly observed control outcomes before time $T$. The lower bound $T^\beta\leq\tau_1<T$ therefore induces a bound on the second term of the RHS of \cref{eq:most-raw-guarantee stagger}, which is
\begin{align}
    \frac{c_0\sinfnorm{\kernel}\log(2N/\delta)}{\sqrt{\tau_1}}
    \leq\frac{c_0\sinfnorm{\kernel}\log(2N/\delta)}{\sqrt{T^\beta}}.
    \label{eq:stag_bound1}
\end{align}
An immediate consequence of \cref{eq:stag_bound1} is 
\begin{align}
    | \mbf N_{1, \eta}^{\text{never-ad}} | \geq \sum_{j \in \neverad} \indic{\gio < \eta'} = | \lnbhdstar|,
    \label{eq:stag_bound2}
\end{align}
thereby providing an upper bound for the variance term in \cref{eq:most-raw-guarantee stagger},
\begin{align}
    \frac{4\sinfnorm{\kernel}(\log n +1.5)}{n | \mbf N_{1, \eta}^{\text{never-ad}} |} \stackrel{\cref{eq:stag_bound2}}{\leq} \frac{4\sinfnorm{\kernel} (\log n + 1.5) }{n \abss{\lnbhdstar} }.
    \label{eq:stag_bound3}
\end{align}
Because $\neverad$ is fixed, these bounds depend on $\mc U$ but not on the realized adoption times. Thus, averaging the guarantee of \cref{thm:stagger-bound} over $\adtimes$ conditional on $\mc U$, we have
\begin{align}
    &\E \bigbrackets{\| \what{\mu}^{(0)}_{1, T, \eta} - \mu_{1, T}^{(0)}\|_\kernel^2 \big| \mc U}
    \nonumber\\
    &\quad\leq\eta+\frac{c_0\sinfnorm{\kernel}\log(2N/\delta)}{\sqrt{T^\beta}}
    +\frac{4\sinfnorm{\kernel}(\log n+1.5)}{n|\lnbhdstar|}
    +4\sinfnorm{\kernel}\delta,
\end{align}
which yields the desired claim.

\newcommand{\gennbhd}{\mbf N_{1, \eta'}^\star}

\subsection{Proof of \cref{lem:exam_lip}}
\label{sub:proof_of_lem_exam_lip}
Since $\mc I$ is fixed and excludes unit $1$, the indicators $\indic{\gio<\eta'}$, $j\in\mc I$, are independent conditional on $u_1$. First, apply the Binomial-Chernoff inequality~\cite[Lem.~A.2]{dwivedi2022counterfactual} so that
\begin{align}
    \sum_{j \in \mc I} \indic{ \gio < \eta' } \geq \frac{1}{2}\sum_{j \in \mc I} \phi_{u_1, \eta'} \quad\text{w.p. at least $1 - \exp \bigbraces{   - |\mc I| \cdot \phi_{u_1, \eta'} / 8  }$},
\end{align}
where $\phi_{u_1,\eta'}\defeq\Prob\braces{\gio<\eta'\mid u_1}$ is their common success probability. Then, by the Lipschitz property \cref{eq:lip} of $g$ and the formula for the volume of a Euclidean ball in $\real^r$, we have
\begin{align}
    \phi_{u_1,\eta'}\geq\Prob\braces{\|u_j-u_1\|_2<\sqrt{\eta'}/L\mid u_1}
    \geq\frac{(\sqrt{\pi\eta'}/4L)^r}{\Gamma(r/2+1)}=\Phi_{\eta'},
    \label{eq:phi_lower_bound}
\end{align}
where $\Gamma$ is the Gamma function. Since $\sqrt{\eta'}/L\leq1$, the cube contains the orthant of this ball directed toward the more distant face in each coordinate. This orthant has $2^{-r}$ of the ball's volume; multiplying by the uniform density $2^{-r}$ gives the factor $4^{-r}$, uniformly over $u_1\in[-1,1]^r$. Substituting $\phi_{u_1,\eta'}\geq\Phi_{\eta'}$ into the Chernoff bound proves \cref{eq:phi_lower_stag}.

Next we derive the Lipschitz constant $L$ of operator $g$ and the value $\sinfnorm{\kernel}$ under \cref{exam:gaussian_family,exam:infinite_family}.
For \cref{exam:gaussian_family}, write $a=u(1)v(1)$, $b=u(2)v(2)$, and define $a',b'$ analogously for $u',v'$. By \cref{eq:moment_factor}, observe the equality
\begin{align}
    &g(u,v)(y)-g(u',v')(y)\nonumber\\
    &\quad=(a-a')\sum_{j=1}^d(-1)^jy_j
    +\frac{b^2-b'^2}{4}\sum_{j=1}^d2^{-j}y_j^2
    +\frac{a^2-a'^2}{4}\left(\sum_{j=1}^d(-1)^jy_j\right)^2.
\end{align}
Then recalling the basis expansion for the RKHS generated by the second-order polynomial kernel~\cite[Example 12.8]{wainwright2019high}, and using $(s+t)^2\leq2s^2+2t^2$ and $\sum_{j=1}^d4^{-j}\leq1/3$, we obtain
\begin{align}
    \|g(u,v)-g(u',v')\|_\kernel^2
    &\leq\frac d2(a-a')^2+\frac{d^2}{8}(a^2-a'^2)^2+\frac1{24}(b^2-b'^2)^2
    \nonumber\\
    &\leq\left(2d+2d^2+\frac23\right)h^2\leq16d^2h^2,
\end{align}
where $h=\|u-u'\|_2\vee\|v-v'\|_2$, and we used $|a-a'|,|b-b'|\leq2h$ and $|a^2-a'^2|,|b^2-b'^2|\leq4h$. So we may take $L=4d$, and also observe $\|\kernel\|_\infty=\max_{x\in[-1,1]^d}(1+\|x\|_2^2)^2=(d+1)^2$.

For \cref{exam:infinite_family}, recall that $\psi_j$ are orthonormal basis of $\mc H$. Observe the following inequality,
\begin{align}
    \| g(u, v) - g(u', v') \|_\kernel^2 &= \sum_{k = 1}^{\infty} \braces{ \alpha_k(u, v)  - \alpha_k(u', v') }^2 \\
    &\leq \left(\sum_{k = 1}^{\infty} L_k^2\right)\big(\|u-u'\|_2\vee\|v-v'\|_2\big)^2.
\end{align}
which implies that a Lipschitz constant of $g$ is $L = \sqrt{\sum_{k = 1}^{\infty}L_k^2}$. As we are assuming the exponential kernel, we have $\sinfnorm{\kernel} = 1$.

\section{Proof of Cor. \ref{cor:mcar-bound}: Guarantees for specific examples under positivity}
\label{proof:positivity_opt}
\label{proof:positivity_complete}

Fix $\delta$ as $(nNT)^{-2}$, which is without loss of generality, as the guarantees appearing in \cref{prop:most-raw-bound,thm:prop-bound} hold for any $\delta\in(0,1)$. Accordingly, here we change the definitions of $(\unbhdstarp, \lnbhdstarp)$ in \cref{eq:pop_neighbor_p} by plugging in $\delta = (nNT)^{-2}$.

We claim that under MCAR, we have an integrated bound
\begin{align}
    \E[ \knorm{\what{\mu}_{1, 1, \eta} - \mu_{1, 1}}^2 | \mc U ] \leq \tilde O \biggbrackets{\eta + \frac{ \sinfnorm{\kernel}}{p \sqrt{T}} + \frac{\sinfnorm{\kernel} }{n p | \underline{\mbf N}^\star_{1, \eta} |}}+\chi_{\mc U},
    \label{eq:fully_int_pos}
\end{align}
where $\underline{\mbf N}^\star_{1, \eta} = \sbraces{j \neq 1:\gio < \eta - \frac{c_0 \sinfnorm{\kernel}\sqrt{2\log(2N/\delta)}}{p\sqrt{T-1}}}$ and $\chi_{\mc U}$ is the remainder specified in \cref{eq:chi-u}. We are assuming values of $\mc U$ and $\eta >0$ so that $| \underline{\mbf N}_{1, \eta}^\star | > 0$. The proof of this claim is deferred to the end of this section.

Invoking \cref{lem:exam_lip} by choosing $\mc I = [N]\setminus \{1\}$, $\eta' = \eta - \frac{c_0 \sinfnorm{\kernel}\sqrt{2\log(2N/\delta)}}{p\sqrt{T-1}}$, with $0<\eta'\leq L^2$, and tracking dependency only on $(n, p, N, T, \eta, L, \sinfnorm{\kernel})$~(and treating other quantities as constants), using \cref{eq:stagger_complete_uniform_loss} on the exceptional event in the lemma, we find that
\begin{align}
    \E[ \knorm{\what{\mu}_{1, 1, \eta} - \mu_{1, 1}}^2 | u_1] \leq \tilde O\biggbrackets{\eta +\frac{\sinfnorm{\kernel}}{p\sqrt{T}} +\frac{\sinfnorm{\kernel}L^r}{np N (\eta')^{r/2} } + \chi} 
    \label{eq:temp_Bnd_sec}
\end{align}
where
\begin{align}
    \chi &= \sinfnorm{\kernel}\bigg[\delta+N\exp\biggbraces{-\frac{p^2(T-1)}8}+\exp\biggbraces{-\frac{p(N-1)\Phi_{\eta'}}{16}}+\exp\biggbraces{-\frac{(N-1)\Phi_{\eta'}}8}\bigg].
    \label{eq:mcar_complete_remainder}
\end{align}

Thus balancing the radius-dependent terms up to constants depending only on $r$ gives the following critical equality and our choice of radius $\eta^\star$:
\begin{align}
    \eta' &= \frac{\sinfnorm{\kernel}L^r(1.5+\log n)}{np(N-1)(\eta')^{r/2}}\\
    &\implies \eta^\star = \parenth{ \frac{\sinfnorm{\kernel}L^r(1.5+\log n)}{np(N-1)} }^{\frac{2}{2 + r}}+\frac{c_0\sinfnorm{\kernel}\sqrt{2\log(2N/\delta)}}{p\sqrt{T-1}}.
    \label{eq:mcar_complete_radius}
\end{align}
For this choice of $\eta^\star$, the bound of \cref{eq:temp_Bnd_sec} is of the order up to constants that depend on $r$\footnote{$\chi$ is of smaller order under the constraints \cref{eq:mcar_complete_parameters} and \cref{eq:mcar_complete_coverage}}
\begin{align}
    \eta^\star +\frac{\sinfnorm{\kernel}}{p\sqrt{T}} =\tilde O\biggbrackets{\parenth{ \frac{\sinfnorm{\kernel}L^r}{npN} }^{\frac{2}{2 + r}}  + \frac{\sinfnorm{\kernel}}{p\sqrt{T}}}.
\end{align}
under the constraints
\begin{align}
    n,N,T\geq2,\qquad \delta=(nNT)^{-2},\qquad C_r\geq32/c_r, \qquad p^2(T-1)&\geq32\log(nNT),
\label{eq:mcar_complete_parameters}
\end{align}
and 
\begin{align}
    L^2\left(\frac{C_r\log(nNT)}{p(N-1)}\right)^{2/r} \leq\parenth{\frac{\sinfnorm{\kernel}L^r(1.5+\log n)}{np(N-1)}}^{\frac{2}{2+r}}\leq L^2.
\label{eq:mcar_complete_coverage}
\end{align}

Here $c_r=\pi^{r/2}/(4^r\Gamma(r/2+1))$, and $r,L,\sinfnorm{\kernel}$ are as in \cref{lem:exam_lip}. At the chosen radius, these conditions give $0<\eta'\leq L^2$, $p(N-1)\Phi_{\eta'}\geq32\log(nNT)$, and $\chi\leq24\sinfnorm{\kernel}\delta$. This remainder is absorbed by $\sinfnorm{\kernel}/(p\sqrt T)$.
\par
Averaging over $u_1$ and plugging the scalings of $L$ and $\sinfnorm{\kernel}$ from \cref{lem:exam_lip} for \cref{exam:gaussian_family,exam:infinite_family} yields the claimed bounds.

\paragraph{Proof of claim \cref{eq:fully_int_pos}} 

Under MCAR, we have the lower bound $\sum_{s\neq 1}p_{1, s}p_{j, s} \geq p^2 (T-1)$ that holds for any value of $\mc U$. An immediate consequence is that we may bound the second term of the RHS of guarantee \cref{eq:prop-bound} by
\begin{align}
    \max_{j \in \unbhdstarp} \frac{c_0 \sinfnorm{\kernel} \sqrt{2 \log (2N/\delta) } }{\sqrt{\sum_{s \neq 1} p_{1, s}p_{j, s} }} \leq \frac{c_0 \sinfnorm{\kernel} \sqrt{ 2\log (2N/\delta) }}{ p \sqrt{T-1} },
\end{align}
and further the set inclusion $\underline{\mbf N}_{1, \eta}^\star \subseteq \lnbhdstarp$ can be derived, from which we observe
\begin{align}
    \sum_{j \in \lnbhdstarp}p_{j, 1} \geq \sum_{j \neq 1}  p_{j, 1} \cdot \indic{\gio < \eta'}  \geq p | \underline{\mbf N}_{1, \eta}^\star |.
    \label{eq:nbhd_lower_last}
\end{align}
So under MCAR, \cref{eq:nbhd_lower_last} induces a bound on the variance term of the RHS of \cref{eq:prop-bound},
\begin{align}
    \frac{\sinfnorm{\kernel}(8\log n + 12)}{n \sum_{j \in \lnbhdstarp}p_{j, 1}} \leq \frac{\sinfnorm{\kernel}(8\log n + 12)}{n p | \underline{\mbf N}_{1, \eta}^\star |}.
\end{align}
So integrating the guarantee of \cref{thm:prop-bound} while conditioning on $\mc U$, we have
\begin{align}
    \E[ \knorm{\what{\mu}_{1, 1, \eta} - \mu_{1, 1}}^2 | \mc U] \leq \eta + \frac{c_0 \sinfnorm{\kernel} \sqrt{ 2\log (2N/\delta) }}{ p \sqrt{T-1} } + \frac{\sinfnorm{\kernel}(8\log n + 12)}{n p | \underline{\mbf N}_{1, \eta}^\star |}  + \chi_{\mc U}
\end{align}
where
\begin{align}
    \chi_{\mc U}=\sinfnorm{\kernel}\biggbrackets{4\delta+8N\exp\biggbraces{-\frac{p^2(T-1)}8}+8\exp\biggbraces{-\frac{p|\underline{\mbf N}_{1,\eta}^\star|}8}}.
    \label{eq:chi-u}
\end{align}

\section{Detailed guarantees under dependent column factors}
\label{sec:dependent-column-appendix}

\providecommand{\mixblock}{b_\delta}
\providecommand{\ndoverlap}{{\mc E_{\rm row}^{\diamond}}}

Here we present the dependent latent column factor assumption and the marginalized error rates for $\kerNN$. We provide a generalized version of \cref{lem:conditional-concentration}, which establishes the row metric concentration result under $\beta$-mixing latent columns.

\subsection{Sampling assumptions and marginalized rates}
\label{sec:noniid}

Both the staggered adoption setting in \cref{sec : staggered adoption} and the HeartSteps example in \cref{sec:application} involve temporally indexed data, where adjacent columns may be dependent and the $\iid$ column factor assumption may be restrictive. We relax the $\iid$ column-factor condition in \cref{assump:latent-independence} while keeping the \kerNN estimator unchanged. For the column latent process, define $\mc F_a^b\defeq\sigma(\{v_t:a\leq t\leq b\})$ and
\begin{align}
    \beta_{\rm mix}(\ell)
    \defeq
    \sup_{t\geq1}\beta\left(\mc F_1^t,\mc F_{t+\ell}^{\infty}\right),
    \label{eq:noniid-mixing-coeff}
\end{align}
where $\beta(\mc F,\mc G)$ is the absolute-regularity coefficient between two sigma-fields~\cite{bradley2005basic}.

\begin{assumption}[Latent factor sampling with dependent columns]
\label{assump:noniid-latent-sampling}
The row latent factors $u_1,\ldots,u_N$ are drawn \iid from a distribution $\Prob_u$ on $\real^r$ and independently of the column latent process $\mc V=\{v_1,\ldots,v_T\}$. The column process is stationary with common marginal distribution $\Prob_v$ on $\real^r$ and is $\beta$-mixing with $\beta_{\rm mix}(\ell)\to0$ as $\ell\to\infty$.
\end{assumption}

For any $\delta>0$, let the block length $\mixblock$ be a positive integer satisfying
\begin{align}
    \ceils{T/\mixblock}\beta_{\rm mix}(\mixblock)
    \leq \frac{\delta}{16N}.
    \label{eq:noniid-mixing-block-length}
\end{align}
The independent column latent factor setting is recovered with $\mixblock=1$. Under geometric mixing, $\mixblock$ can be chosen logarithmic in $(N,T,\delta^{-1})$.

\paragraph{Adjusting the proof technique for dependent latent column factors}
Under \cref{assump:noniid-latent-sampling}, the target factor $v_1$ may be dependent on the non-target columns used to construct the neighborhood, so the independence step in the max bias bound \cref{eq:bb_bnd} no longer applies. We use Berbee coupling to control the row-metric error, then bound the average bias using inner and outer population neighborhoods. The shell conditions in \cref{subsec:dependent-rate-details} control the inner and outer neighborhoods' relative sizes. 

\begin{corollary}[Marginalized rates under dependent columns]
\label{cor:noniid-rate-form}
Assume \cref{assump:factorization,assump:noniid-latent-sampling,assump : measurement generation}. Consider the settings of \cref{cor:stagger-bound,cor:mcar-bound} for \cref{exam:infinite_family}, except that the column factors form the stationary $\beta$-mixing process in \cref{assump:noniid-latent-sampling} with the same marginal distribution. Suppose the coefficient functions $\alpha_k$ are $L_k$-Lipschitz and $L=(\sum_{k\geq1}L_k^2)^{1/2}<\infty$. Additional missingness specific conditions are stated in \cref{subsec:dependent-rate-details}.
\begin{enumerate}[label=(\alph*), leftmargin=*]
    \item Under staggered adoption, for some $\eta^\star$,
    \begin{align}
        \E\brackets{\|\what\mu^{(0)}_{1,T,\eta^\star}-\mu^{(0)}_{1,T}\|_\kernel^2}
        \leq \widetilde O\biggbrackets{
            \parenth{\frac{L^r}{nN^\alpha}}^{2/(2+r)}
            +\sqrt{\frac{\mixblock}{T^\beta}}
        }.
        \label{eq:noniid-stagger-rate}
    \end{align}
    \item Under MCAR, for some $\eta^\star$,
    \begin{align}
        \E\brackets{\|\what\mu_{1,1,\eta^\star}-\mu_{1,1}\|_\kernel^2}
        \leq \widetilde O\biggbrackets{
            \parenth{\frac{L^r}{npN}}^{2/(2+r)}
            +\sqrt{\frac{\mixblock}{p^2T}}
        }.
        \label{eq:noniid-mcar-rate}
    \end{align}
\end{enumerate}
\end{corollary}

Compared with the corresponding \cref{exam:infinite_family} bounds in \cref{cor:stagger-bound,cor:mcar-bound}, the dependence inflates the second term by $\sqrt{\mixblock}$, which corresponds to the estimation error induced by the row metrics of \kerNN. These bounds match the iid rate orders when $\mixblock=1$. Under geometric mixing, $\mixblock$ grows logarithmically, so dependence introduces at most an additional logarithmic factor in the second term.

\begin{remark}[Role of the shell conditions]
Under independent columns, the squared bias is bounded by the largest population distance among selected donors, since conditioning on neighbor selection retains the target factor's law; see \cref{eq:bb_bnd}. With dependent columns, we bound the squared bias by summing population distances over all possible outer neighbors and dividing by the number of guaranteed inner neighbors. The bound in \cref{eq:noniid-corollary-risk} is therefore proportional to the outer distance threshold, but is inflated by the ratio of outer to inner neighbor counts. 
The shell conditions \cref{eq:noniid-stagger-shell,eq:noniid-mcar-shell} ensure comparable neighborhood masses.
\end{remark}

\subsection{Proof of Corollary~\ref{cor:noniid-rate-form}}
\label{proof:noniid-rate-form}
\label{subsec:dependent-rate-details}

\newcommand{\lnbhdstare}{\underline{\mbf N}_{1,\eta,e}^{\star}}
\newcommand{\unbhdstare}{\overline{\mbf N}_{1,\eta,e}^{\star}}

Set $c_{\rm mix}=8$ and $\delta=(NT)^{-2}$. For \cref{exam:infinite_family}, $g$ is $L$-Lipschitz in the row factor and $\|\kernel\|_\infty=1$. Fix $\kappa>1$. Without loss of generality, assume $\sum_{j\in\lnbhdstare}A_{j,1}>0$.

Fix $e>0$ as some non-random error bound, which is specified later in the proof when we fix a missing pattern. Define population neighborhoods
\begin{align}
    \label{eq:dep-nbhd}
    \lnbhdstare\defeq\{j\neq1:\gio\leq\eta-e\},\qquad
    \unbhdstare\defeq\{j\neq1:\gio\leq\eta+e\}.
\end{align}
We replace \cref{lem:conditional-concentration} by the following lemma, proved in \cref{subsec:noniid-row-concentration-proof}.

\begin{lemma}[Row-metric concentration under beta-mixing columns]
\label{lem:noniid-row-concentration}
Let
\cref{assump:factorization,assump:noniid-latent-sampling,assump : unobs confounding,assump : measurement generation}
hold. For any $\delta\in(0,1)$, fix an integer $b\geq1$ such that
$\ceils{T/b}\beta_{\rm mix}(b)\leq\delta/16$. Then for any row $j$ with
$A_{j,1}=1$ and $\overoj>0$,
\begin{align}
    \Prob\biggbraces{\abss{\rho_{j,1}-\gio}>c_{\rm mix}\|\kernel\|_\infty\sqrt{\frac{b\log(4/\delta)}{\overoj}}\biggm| \mc U,\mc A} \leq \delta.
\end{align}
\end{lemma}

Assume that the error radius in \cref{lem:noniid-row-concentration}, with $b=\mixblock$ and probability $\delta/N$, is at most $e$ for $j\neq1$ with $A_{j, 1}=1$. Applying \cref{lem:noniid-row-concentration} and taking a union bound gives an event
\begin{align}
    \ndoverlap \defeq \{ |\rho_{j,1}-\gio|\leq e \text{ for all $j\neq 1$ with $A_{j, 1} = 1$} \}
\end{align}
of probability at least $1-\delta$ conditional on $(\mc U,\mc A)$. On the event $\ndoverlap$, the argument of \cref{eq:n_ubnd,eq:n_lbnd} yield inclusions
\begin{align}
    j \in \lnbhdstare \implies j \in \mbf N_{1, \eta} \implies j \in \unbhdstare.
    \label{eq:dep-nbhd-inclusion}
\end{align}
Bounding the bias component is what differs from the proof of \cref{lem:conditional-concentration}, and is the reason for why we introduce the shell condition. 

From the proof of \cref{lem : MMD decomp}, recall that conditionally integrating out the measurements via the decomposition \cref{eq:embedding-decomp} yields
\begin{align}
    \E\big[ \|\what \mu_{1, 1, \eta} - \mu_{1, 1}\|_\kernel^2 \mid \mc V, \mc D_{-1} \mc U, \mc A \big] \leq \bigg\| \frac{\sum_{j \in \mbf N_{1,\eta}}A_{j, 1}b(j, 1)}{\sum_{j \in\mbf N_{1, \eta}}A_{j, 1}} \bigg\|_\kernel^2 + \frac{4\|\kernel\|_\infty (\log n + 1.5)}{n \sum_{j \in \mbf N_{1, \eta}}A_{j, 1}}.
\end{align}
Multiply the above display by $\indic{\ndoverlap}$  and then integrate it conditional on $(\mc U, \mc A)$, then we observe
\begin{align}
    \E\big[ \indic{\ndoverlap} \| \what \mu_{1, 1, \eta} -\mu_{1, 1}\|_\kernel^2 \mid \mc U, \mc A\big] \stackrel{\cref{eq:dep-nbhd}\cref{eq:dep-nbhd-inclusion}}{\leq} (\eta+e)\frac{\sum_{j\in\unbhdstare}A_{j,1}}{\sum_{j\in\lnbhdstare}A_{j,1}} + \frac{4\|\kernel\|_\infty(\log n+1.5)}{n\sum_{j\in\lnbhdstare}A_{j,1}}.
\end{align}

Lastly, we observe $\E[ \indic{(\ndoverlap)^c}\|\what \mu_{1, 1, \eta} - \mu_{1, 1}\|_\kernel^2|\mc U, \mc A] \leq 4 \|\kernel\|_\infty \delta$; see \cref{eq:stagger_complete_uniform_loss}. Combining all the results, we observe
\begin{align}
    &\E\big[\|\what\mu_{1,1,\eta}-\mu_{1,1}\|_\kernel^2
        \mid\mc U,\mc A\big]\\
    &\qquad\leq(\eta+e)
        \frac{\sum_{j\in\unbhdstare}A_{j,1}}
             {\sum_{j\in\lnbhdstare}A_{j,1}}
        +\frac{4\|\kernel\|_\infty(\log n+1.5)}
              {n\sum_{j\in\lnbhdstare}A_{j,1}}
        +4\|\kernel\|_\infty\delta.
    \label{eq:noniid-corollary-risk}
\end{align}
The same argument applies to the target entry $(1,T)$ using the non-target overlapping columns in chronological order.

\paragraph{Staggered adoption}
\label{subsec:noniid-stagger-rate-details}

Use the row-factor model and fixed donor construction of \cref{cor:stagger-bound} for \cref{exam:infinite_family}. Thus $\neverad$ is the entire control donor pool at $T$. The remaining adoption times may depend on $\mc U$. Take $n,N,T\geq2$ and $L>0$.

Every donor in $\neverad$ has at least $T^\beta$ overlapping control outcomes with unit $1$. Take $\eta = \eta^\star$ and $e = e_{\mc A}^\diamond$ where
\begin{align}
    e_{\mc A}^\diamond \defeq c_{\rm mix}\|\kernel\|_\infty \sqrt{\frac{\mixblock\log(4N/\delta)}{T^\beta}} \qtext{and} \eta^\star=\max\biggbraces{\kappa e_{\mc A}^\diamond,\parenth{\frac{L^r(1+\log n)}{nN^\alpha}}^{2/(2+r)}}.
    \label{eq:noniid-stagger-radius}
\end{align}
With $c_r$ as defined in \cref{proof:stagger_complete}, require
\begin{align}
    0<\eta^\star-e_{\mc A}^\diamond\leq L^2,\qquad
    c_rN^\alpha\parenth{\frac{\eta^\star-e_{\mc A}^\diamond}{L^2}}^{r/2}
    \geq8\log(2/\delta).
    \label{eq:noniid-stagger-coverage}
\end{align}

Since $\neverad$ is a fixed set, we observe 
\begin{align}
    \sum_{j \in \lnbhdstare} A_{j, T} = \sum_{j \in \neverad} \indic{\gio \leq \eta - e}.
\end{align}
Let's set $\eta' = \eta - e$, and recall $\phi_{u_1, \eta'} = \Prob(\gio \leq \eta - e\mid u_1)$ and $c_r \defeq \frac{\pi^{r/2}}{4^r\Gamma(r/2+1)}$. Conditioned on $u_1$, the $\iid$ structure of $\mc U$ allows the statistics in the above display to follow binomial distribution, $\mrm{Bin}(N^\alpha, \phi_{u_1, \eta'})$, so with probability at least $1 - \delta/2$ and $h = \sqrt{\eta - e}/L$, 
\begin{align}
    \sum_{j \in \lnbhdstare} A_{j, T} \stackrel{\cref{eq:noniid-stagger-coverage}}{\geq} \frac{N^\alpha}{2} \phi_{u_1, \eta'} \stackrel{\cref{eq:lip}}{\geq} \frac{N^\alpha}{2}\Prob(u_j \in B(u_1, h)\mid u_1) \geq \frac{c_rN^\alpha}{2} \bigg( \frac{\eta - e}{L^2} \bigg)^{r/2}.
\end{align}

Assume existence of a universal constant $C > 0$ such that
\begin{align}
    \Prob\{\gio\leq\eta + e\mid u_1\} \leq C\Prob\{\gio\leq\eta - e\mid u_1\}.
    \label{eq:noniid-stagger-shell}
\end{align}
Again, invoke the binomial distribution for $\sum_{j \in \unbhdstare}A_{j, T}$ so that with probability at least $1 - \delta$ (which is a result of a union bound), we observe
\begin{align}
    \sum_{j \in \unbhdstare}A_{j, T} \leq 2N^\alpha \Prob(\gio \leq \eta + e \mid u_1) \stackrel{\cref{eq:noniid-stagger-shell}}{\implies} \frac{\sum_{j \in \unbhdstare}A_{j, T}}{\sum_{j \in \lnbhdstare}A_{j, T}} \leq 4C.
    \label{eq:stagg-shell-results}
\end{align}
So on the event $\mc E_{\mrm{Bin}}$ of probability at least $1 - \delta$, we see 
\begin{align}
    \E\big[ \indic{\mc E_{\mrm{Bin}}} \| \what \mu_{1, T, \eta}^{(0)}-\mu_{1, T}\|_\kernel^2 \mid \mc U, \mc A \big] \leq 4C(\eta+e) + \frac{8\|\kernel\|_\infty (\log n +1.5)L^r}{c_r N^\alpha (\eta')^{r/2}} + 4\|\kernel\|_\infty\delta
    \label{eq:stagg-bin-bound}
\end{align}
and again observe $\E\big[ \indic{\mc E^c_{\mrm{Bin}}} \| \what \mu_{1, T, \eta}^{(0)}-\mu_{1, T}\|_\kernel^2 \mid \mc U, \mc A \big] \leq 4\|\kernel\|_\infty \delta$. 

Then \cref{eq:noniid-stagger-radius} implies 
\begin{align}
    \eta + e \leq (1 + \kappa^{-1})\eta^\star \qtext{and} \eta - e \geq (1 - \kappa^{-1})\eta^\star.
\end{align}
The first inequality in the above display induce $4C(\eta+e)\leq 4C(1 +\kappa^{-1})\eta^\star$ and the second implies 
\begin{align}
    \frac{8(\log n+1.5)}{c_r n N^\alpha} \bigg( \frac{L^2}{\eta'} \bigg)^{r/2} \leq \frac{8}{c_r (1 - \kappa^{-1})^{r/2}} \frac{L^2(\log n+1.5)}{n N^\alpha (\eta^\star)^{r/2}};
\end{align}
these bounds give us a bound for the sum of the first two terms in \cref{eq:stagg-bin-bound}. Next, note that $\delta = (NT)^{-2} \leq e \leq \eta^\star/\kappa$ and finally by invoking the second argument of the maximum of \cref{eq:noniid-stagger-radius}, we observe
\begin{align}
    \E\Big[ \| \what \mu_{1, T, \eta^\star}^{(0)}-\mu_{1, T}\|_\kernel^2 \Big] \leq \tilde O(\eta^\star).
\end{align}

\paragraph{MCAR}
\label{subsec:noniid-mcar-rate-details}
Take $\eta = \eta^\star$ and $e = e_p^\diamond$, where
\begin{align}
    e_p^\diamond \defeq c_{\rm mix}\|\kernel\|_\infty \sqrt{\frac{2\mixblock\log(4N/\delta)}{p^2(T-1)}} \qtext{and} \eta^\star=\max\biggbraces{\kappa e_p^\diamond,\parenth{\frac{L^r(1+\log n)}{npN}}^{2/(2+r)}}.
    \label{eq:noniid-mcar-radius}
\end{align}

Under MCAR, $A_{i, t}$ are independent with $p = \Prob(A_{i, t} = 1)$ and are independent to all remaining randomness in the model. Set $\eta'=\eta-e$ and use $\phi_{u_1,\eta'}=\Prob\{\gio\leq\eta'\mid u_1\}$ as above. 

Assume there exists universal constants $c_-, c_+ > 0$ such that
\begin{align}
    c_-\parenth{\frac{t}{L^2}}^{r/2}\leq \Prob\{\gio\leq t\mid u_1\}\leq c_+\parenth{\frac{t}{L^2}}^{r/2},\qquad t\in\{\eta-e,\eta+e\}.
    \label{eq:noniid-mcar-shell}
\end{align}
We further assume
\begin{align}
    p^2(T-1) \geq 8 \log (2N/\delta) \qtext{and} c_{-}(N-1)p\bigg( \frac{\eta'}{L^2} \bigg)^{r/2} \geq 8 \log (2N/\delta). 
    \label{eq:dep-mcar-ov}
\end{align}

For each $j\ne1$, the overlap $\overoj$ has distribution $\mrm{Bin}(T-1,p^2)$ and is independent of $A_{j,1}$. Applying \cref{eq:twoevent_conc} and taking a union bound over donors gives
\begin{align}
    \Prob\bigg\{\overoj\geq\frac{p^2(T-1)}{2}\text{ for $j\neq 1$, $A_{j,1}=1$} \biggm|\ \mc U\bigg\}
    \geq 1 - (N-1)p\exp\bigg\{-\frac{p^2(T-1)}{8}\bigg\}.
    \label{eq:noniid-mcar-overlap-failure}
\end{align}
In the high probability event in the above display, the choice of $e$ in \cref{eq:noniid-mcar-radius} bounds the row metric error radius uniformly.

Conditional on $u_1$, the following observation count has Binomial distribution $\mrm{Bin}(N-1,p\phi_{u_1,\eta'})$,
\begin{align}
    \sum_{j\in\lnbhdstare}A_{j,1}=\sum_{j\ne1}A_{j,1}\indic{\gio\leq\eta'}.
\end{align}
Similarly, the $\sum_{j \in \unbhdstare}A_{j, 1}$ has distribution $\mrm{Bin}(N-1,p\Prob\{\gio\leq\eta+e\mid u_1\})$. Binomial concentration and a union bound give
\begin{align}
    \sum_{j\in\lnbhdstare}A_{j,1}
    &\geq \frac{(N-1)p}{2}\phi_{u_1,\eta'}
    \stackrel{\cref{eq:noniid-mcar-shell}}{\geq} \frac{c_-(N-1)p}{2}\parenth{\frac{\eta'}{L^2}}^{r/2} \qtext{and}\\
    \sum_{j\in\unbhdstare}A_{j,1}
    &\leq 2(N-1)p\Prob\{\gio\leq\eta+e\mid u_1\}\notag
\end{align}
with conditional probability at least $1 - 2\exp\{-(N-1)p\phi_{u_1,\eta'}/8\}$. When both inequalities hold, \cref{eq:noniid-mcar-shell} and $\eta=\eta^\star{\geq}\kappa e$ give
\begin{align}
    \frac{\sum_{j\in\unbhdstare}A_{j,1}}{\sum_{j\in\lnbhdstare}A_{j,1}}\leq 4\frac{c_+}{c_-}\parenth{\frac{\kappa+1}{\kappa-1}}^{r/2}.
\end{align}
Using the identical reasoning in the staggered adoption, we observe
\begin{align}
    \E\big[\|\what\mu_{1,1,\eta^\star}-\mu_{1,1}\|_\kernel^2\big]\leq 4\frac{c_+}{c_-}\parenth{\frac{\kappa+1}{\kappa-1}}^{r/2}(\eta+e)+\frac{8(\log n+1.5)}{c_-n(N-1)p}\parenth{\frac{L^2}{\eta'}}^{r/2}+8\|\kernel\|_\infty \delta.
\end{align}
Using  $\eta+e\leq(1+\kappa^{-1})\eta^\star$ and $\eta'\geq(1-\kappa^{-1})\eta^\star$, \cref{eq:noniid-mcar-radius} yields
\begin{align}
    (\eta^\star)^{1+r/2}\geq\frac{L^r(1+\log n)}{npN}.
\end{align}
Using $\delta=(NT)^{-2}\leq e_p^\diamond\leq\eta^\star/\kappa$, we can derive 
\begin{align}
    \E\Big[ \| \what \mu_{1, 1, \eta^\star}-\mu_{1, 1}\|_\kernel^2 \Big] \leq \tilde O(\eta^\star).
\end{align}

\subsection{Proof of \cref{lem:noniid-row-concentration}}
\label{subsec:noniid-row-concentration-proof}

Condition on $(\mc U,\mc A)$ and index the $m_j=\overoj$ overlapping columns as $s_1 < \cdots s_{m_j}$.
Set
\[
    Y_\ell=\what{\mmd}_\kernel^2(\empdistZ{j,s_\ell},\empdistZ{1,s_\ell})-\gio,
    \qquad \ell=1,\ldots,m_j.
\]
As in the $\iid$ proof, unbiasedness of the U-statistic and \cref{assump : unobs confounding} imply that $Y_\ell$ is conditionally centered and lies in an interval of length at most $8\|\kernel\|_\infty$. Conditioning on $(\mc U,\mc A)$ leaves the column-process law unchanged, by invoking the conditional independence between $\mc A$ and $\mc V$, and that $\mc U \indep \mc V$. Since $s_{\ell+h}-s_\ell\geq h$ and the measurement noise is conditionally independent across columns, the lag-$h$ mixing coefficient of $Y_\ell$ is at most $\beta_{\rm mix}(h)$.

Set $q_j=\ceils{m_j/b}$ and partition the sequence into blocks
\[
    I_r=\{(r-1)b+1,\ldots,\min(rb,m_j)\},\qquad
    B_r=(Y_\ell)_{\ell\in I_r},\qquad r=1,\ldots,q_j.
\]
Each block contains $b$ entries, except possibly the last. Within the odd-numbered sequence $B_1,B_3,B_5,\ldots$, each block is separated from the next by an intervening block of $b$ entries; the same holds for the even-numbered sequence $B_2,B_4,B_6,\ldots$. Conditional on $(\mc U,\mc A)$, Berbee's coupling lemma~\cite[Lemma~5.1]{rio2017asymptotic} gives copies $B_r^*$ with the same blockwise marginal laws as $B_r$, and that are independent within the odd-indexed and within the even-indexed group. A union bound over coupling failures gives
\begin{align}
    \Prob\{B_r^*\neq B_r\text{ for some }r\mid\mc U,\mc A\} \leq 8\ceils{T/b}\beta_{\rm mix}(b).
    \label{eq:berbee-event}
\end{align}
For the parity class $a\in\{\mathrm{odd},\mathrm{even}\}$, let $S_a^*$ be the sum of all entries in the copied blocks within group $a$, and let $m_a$ be their total number of entries. Outside the mismatch event in \cref{eq:berbee-event}, we observe $\sum_{\ell=1}^{m_j}Y_\ell=S_{\rm odd}^*+S_{\rm even}^*$. Each copied block sum is centered and has range length at most $8\|\kernel\|_\infty|I_r|$. Hoeffding's lemma and $|I_r|^2\leq b|I_r|$ therefore give
\[
    \E[\exp(\lambda S_a^*)\mid\mc U,\mc A]
    \leq\exp\{8\lambda^2\|\kernel\|_\infty^2b m_a\}.
\]
Cauchy--Schwarz gives the bound $\exp\{16\lambda^2\|\kernel\|_\infty^2b m_j\}$ for the moment generating function of their sum. Chernoff's inequality and the coupling failure bound imply
\begin{align}
    \Prob\bigg\{\bigg|\frac1{m_j}\sum_{\ell=1}^{m_j}Y_\ell\bigg|>x
        \biggm|\mc U,\mc A\bigg\}
    &\leq2\exp\bigg\{-\frac{m_jx^2}{64\|\kernel\|_\infty^2b}\bigg\}
        +8\ceils{T/b}\beta_{\rm mix}(b).\notag
\end{align}
Taking $x=8\|\kernel\|_\infty\sqrt{b\log(4/\delta)/m_j}$ makes the first term $\delta/2$, and the assumed block length bounds the second by $\delta/2$.

\section{Minimax lower bound}
\label{sec:full-matrix-minimax}

We establish a minimax lower bound for estimating $\mu_{1,1}$ and lower bound the marginalized squared $\mmd$ error by a constant multiple of the first term of \cref{cor:mcar-bound} (b). Then we establish minimax optimality of $\kerNN$ at a restricted regime. 

We operate under the setting of \cref{cor:mcar-bound} (b); so missingness is MCAR and the latent factors are independent samples from a uniform distribution with range $[-1, 1]$. The target entry $(1,1)$ is set unseen, $A_{1, 1} = 0$. Fix integers $r,d\geq1$, $N\geq2$, $T,n\geq1$, and parameters $L>0$ and $p\in(0,1]$. 
Let $\mc M_{r,L}$ be the class of the mappings $\mu$ from the product space of latent factors $\mc U \times \mc V$ to distributions with support $\mc X=[-1,1]^d$, with regularity condition speicified in \cref{exam:infinite_family}; for the exponential kernel $\kernel(x,y)=\exp(-\|x-y\|_2^2/2)$, we require
\begin{align}
     |\alpha_\ell(u,v)-\alpha_\ell(u',v')|\leq L_\ell\big(\|u-u'\|_2\vee\|v-v'\|_2\big) \qtext{and} \sum_{\ell\geq1}L_\ell^2\leq L^2
    \label{eq:full-minimax-model-class}
\end{align}
for all $u,u',v,v'\in[-1,1]^r$, where $\{\psi_\ell\}_{\ell\geq1}$ is the same orthonormal basis of $\mc H$ used in \cref{exam:infinite_family}, held fixed throughout. $L$ is the common upper bound on the aggregate Lipschitz constant throughout as well.

For a mapping $\mu\in\mc M_{r,L}$, let $\Prob_\mu$ denote the joint law of the latent factors, missingness indicators, and all observed measurements, and let $\E_\mu$ denote expectation under this law. 
Recall that $\mc D_1$ and $\mc D_{-1}$ denoted the collection of observed measurements in the target first column and the remaining columns, respectively; see \cref{sec : disc finite dim model}. Define $\mc D_{\mrm{obs}}\defeq(\mc A,\mc D_1,\mc D_{-1})$ as the entire available observations and define the minimax risk:
\begin{align}
    \mc R^\star
    &\defeq\inf_{\what{\mu}_{1,1}(\mc D_{\mrm{obs}})}
    \sup_{\mu\in\mc M_{r,L}}
    \E_\mu\brackets{\|\what{\mu}_{1,1}-\mu_{1,1}\|_\kernel^2}.
    \label{eq:full-minimax-risks}
\end{align}
The infimum ranges over all measurable estimators using $\mc D_{\mrm{obs}}$. The expectation averages over the latent factors, missingness, and measurements. 

Define the bandwidth scale and the target rate by
\begin{align}
    h_\star\defeq(npNL^2)^{-1/(r+2)}
    \qtext{and}
    a_\star\defeq L^2h_\star^2
    =\left(\frac{L^r}{npN}\right)^{2/(r+2)}.
    \label{eq:full-minimax-bandwidth}
\end{align}
Refer to \cref{subsec:full-matrix-minimax-proof} for the proof of the following lower bound result.

\begin{proposition}[\tbf{Minimax lower bound}]
\label{thm:full-matrix-oracle-minimax}
Fix a constant $H\in(0,\infty)$. If
\begin{align}
    h_\star\leq1,\qquad Lh_\star\leq\frac14,
    \qquad Th_\star^r\leq H,
    \label{eq:full-minimax-regime}
\end{align}
then there exists a constant depending on $r$ and $H$, $c=c(r,H)>0$, such that
\begin{align}
    \mc R^\star
    \geq ca_\star.
    \label{eq:full-minimax-lower-bound}
\end{align}
\end{proposition}

The proof involves partitioning the latent factor space into cubes of side length comparable to $h_\star$. The condition $Th_\star^r\leq H$ keeps the probability of the event that no other column factor shares the cube containing \(v_1\) bounded away from zero. This allows the proof to change the target distribution without changing measurement distributions in the other columns, limiting the information available to distinguish the models. A larger $H$ extends the lower bound to larger $T$, at the cost of a smaller lower bound constant, which contains the factor $\exp(-2^{1-r}H)$; a smaller $H$ gives a stronger constant under a more restrictive condition. 

\subsection{Minimax optimality and scope}
\label{subsec:full-matrix-minimax-connections}

\cref{thm:full-matrix-minimax-optimality} combines the lower bound in \cref{thm:full-matrix-oracle-minimax} with a refinement of \cref{cor:mcar-bound}(b) that holds uniformly over $\mc M_{r,L}$. For $n,T\geq2$ and $0<p<1$, set the common neighborhood radius by
\begin{align}
    \varepsilon_N=\frac{8\sqrt{\log(2N)}}{p\sqrt{T-1}},
    \qquad \eta^\star=2(a_\star+\varepsilon_N).
    \label{eq:full-minimax-common-radius}
\end{align}
The worst-case risk of \kerNN at this radius is
\begin{align}
    \mc R_{\text{\kerNN}}\defeq\sup_{\mu\in\mc M_{r,L}}
    \E_\mu\brackets{\|\what{\mu}_{1,1,\eta^\star}-\mu_{1,1}\|_\kernel^2}.
\end{align}
If there are no observed row neighbors, set $\what{\mu}_{1,1,\eta^\star}=\dirac_0$, the point mass at zero. For $0<p<1$, conditioning on $A_{1,1}=0$ preserves the estimator's MCAR risk because it uses neither the target measurements nor $A_{1,1}$.

\paragraph{Matching regimes}
In addition to the window \cref{eq:full-minimax-joint-window}, \cref{thm:full-matrix-minimax-optimality} assumes
\begin{align}
    h_\star\leq1,\qquad a_\star\leq\frac1{16},\qquad p(N-1)h_\star^r\geq c_r\log(1/a_\star),
    \label{eq:full-minimax-joint-sample-conditions}
\end{align}
where $h_\star$ is defined in \cref{eq:full-minimax-bandwidth} and $c_r$ is a constant depending only on $r$. These conditions control the smoothing scale and ensure enough nearby observed rows. Since $h_\star^{-r}=npNa_\star$, the upper limit on $T$ in \cref{eq:full-minimax-joint-window} gives $Th_\star^r\leq1$, keeping the probability of target-column isolation bounded away from zero; the lower limit makes the row-similarity error at most of order $a_\star$.

For example, fix $d$, $n\geq2$, $p\in(0,1)$, $L>0$, and $r>4$, and take
\begin{align}
    T=\lfloor N^\gamma\rfloor,
    \qquad \frac{4}{r+2}<\gamma<\frac{r}{r+2}.
    \label{eq:full-minimax-fixed-parameter-window}
\end{align}
Then the conditions of \cref{thm:full-matrix-minimax-optimality} hold for sufficiently large $N$. The lower-bound proposition itself remains valid for every $r\geq1$ under \cref{eq:full-minimax-regime}.

\subsection{Proof of \cref{thm:full-matrix-oracle-minimax}}
\label{subsec:full-matrix-minimax-proof}

Write $\mc D_{\mrm{or}}\defeq(\mc D_{\mrm{obs}},\mc U,\mc V)$ as the set of observed and latent variables and define the oracle minimax risk by 
\begin{align}
    \mc R_{\mrm{or}}^\star
    \defeq\inf_{\what{\mu}_{1,1}(\mc D_{\mrm{or}})}
    \sup_{\mu\in\mc M_{r,L}}
    \E_\mu\brackets{\|\what{\mu}_{1,1}-\mu_{1,1}\|_\kernel^2}
    \quad \text{where} \quad \mc R_{\mrm{or}}^\star\leq\mc R^\star.
    \label{eq:full-minimax-oracle-risk}
\end{align}
The infimum ranges over all measurable functions of the oracle data $\mc D_{\mrm{or}}$. The inequality holds because every estimator based on $\mc D_{\mrm{obs}}$ can be recovered with $\mc D_{\mrm{or}}$ by ignoring $\mc U,\mc V$. 
It suffices to prove $\mc R_{\mrm{or}}^\star\geq ca_\star$. 

We first use a Bernoulli submodel to reduce squared $\mmd$ error to scalar squared error. We then construct a finite Lipschitz family and derive the lower bound through conditional testing and averaging.

\paragraph{Bernoulli submodel and reduction to scalar estimation}
We first restrict the model class to $\mu = \mu_f$ indexed by $f \in \mc F_{r, L}$, where measurements are supported on two values, $x_0=0$ and $x_1=e_1=(1,0,\ldots,0)$ in $[-1,1]^d$. 
Let $\mc F_{r,L}$ be the class of functions $f:[-1,1]^r\times[-1,1]^r\to[-1/4,1/4]$ that are $L$-Lipschitz for the metric $\|u-u'\|_2\vee\|v-v'\|_2$. For $f\in\mc F_{r,L}$, define the probability $q_f$, Bernoulli distribution $\mu^f$
\begin{align}
    q_f(u,v)\defeq\frac12+f(u,v) \qtext{and}
    \mu^f(u,v)\defeq(1-q_f(u,v))\dirac_{x_0}+q_f(u,v)\dirac_{x_1}
    \label{eq:full-minimax-bernoulli-submodel}
\end{align}
Here $q_f(u,v)\in[1/4,3/4]$ is the probability that a measurement equals $x_1$; equivalently, its indicator of equaling $x_1$ has distribution $\Bern(q_f(u,v))$. The function $f$ perturbs this probability from $1/2$. 

Define the difference of their feature maps and its squared RKHS norm by
\begin{align}
    d_\kernel&\defeq\kernel(x_1,\cdot)-\kernel(x_0,\cdot) \qtext{and}
    \kappa_\kernel\defeq\|d_\kernel\|_\kernel^2
    =2(1-e^{-1/2})\in(0,1).
    \label{eq:full-minimax-kernel-direction}
\end{align}
Then the kernel mean embedding of the Bernoulli distribution $g_f(u, v)$ is 
\begin{align}
    g_f(u,v)&\defeq\mu^f(u,v)\kernel
    =(1-q_f(u,v))\kernel(x_0,\cdot)+q_f(u,v)\kernel(x_1,\cdot) =\kernel(x_0,\cdot)+q_f(u,v)d_\kernel.
\end{align}

For every $f\in\mc F_{r,L}$, the model $\mu^f$ belongs to $\mc M_{r, L}$, which we verify below in \cref{lem:full-minimax-inclusions} (a). For each $f\in\mc F_{r,L}$, we write $\Prob_f\defeq\Prob_{\mu^f}$ for the joint law induced by $\mu^f$ and $\E_f\defeq\E_{\mu^f}$ for expectation under this law. The target distribution is $\mu_{1,1}^f=\mu^f(u_1,v_1)$. So we can claim that for every oracle estimator $\what\mu_{1,1}(\mc D_{\mrm{or}})$,
\begin{align}
    \sup_{\mu\in\mc M_{r,L}}
    \E_\mu\brackets{\|\what\mu_{1,1}-\mu_{1,1}\|_\kernel^2}
    \geq
    \sup_{f\in\mc F_{r,L}}
    \E_f\brackets{\|\what\mu_{1,1}-\mu_{1,1}^f\|_\kernel^2}.
    \label{eq:full-minimax-submodel-comparison}
\end{align}

Under the class of measures $\mu^f, f \in \mc F_{r, L}$, it is sufficient to learn $q_f(u_1, v_1)$ to recover $\mu_{1, 1}^f$. We next transform any distribution valued oracle estimator $\what\mu_{1,1}(\mc D_{\mrm{or}})$ into a scalar valued estimator of $q_f(u_1,v_1)$. Cauchy--Schwarz then lower-bounds the squared $\mmd$ error by a constant multiple of the scalar estimator's squared error.
Define the scalar valued estimator
\begin{align}
    \what q = \frac{\langle\what{\mu}_{1,1}(\mc D_{\mrm{or}})\kernel-\kernel(x_0,\cdot),d_\kernel\rangle_\kernel}{\kappa_\kernel}.
\end{align}
Using the embedding $g_f$ form above, we observe
\begin{align}
    \what q-q_f(u_1,v_1)
    =\frac{\langle\what{\mu}_{1,1}(\mc D_{\mrm{or}})\kernel-g_f(u_1,v_1),d_\kernel\rangle_\kernel}{\kappa_\kernel};
\end{align}
Cauchy--Schwarz therefore induces the pointwise inequality
\begin{align}
    \|\what{\mu}_{1,1}-\mu_{1,1}^f\|_\kernel^2
    \geq\kappa_\kernel\big(\what q-q_f(u_1,v_1)\big)^2.
    \label{eq:full-minimax-scalar-reduction}
\end{align}
Therefore, a lower bound valid for every scalar-valued estimator constructed from $\mc D_{\mrm{or}}$ via $\what q$ further yields a lower bound on \cref{eq:full-minimax-submodel-comparison}. Note that the transformed estimator $\what q$ need not lie in $[0,1]$.

Next, we construct a finite subclass of $\mc F_{r,L}$ and compare two models at a time by changing a single bump.

\paragraph{Construction of Lipschitz bumps in both latent coordinates}

We construct two models whose measurement laws at the target entry can differ even when their measurement distributions agree in every other column. To do so, we change the probability of observing $x_1$ only when the row factor lies in a specified cube in row latent space and the column factor lies in a specified cube in a column latent space.

Partition each coordinate interval $[-1,1]$ into $K\defeq\lceil2/h_\star\rceil$ equal subintervals of length $h\defeq2/K$. 
Note that since $h_\star\leq1$, we have $K\geq2$ and
\begin{align}
    \frac23h_\star\leq h\leq h_\star.
    \label{eq:full-minimax-grid-scale}
\end{align}
The resulting grid of the subintervals constructs $M=K^r$ cubes $Q_1,\ldots,Q_M$, each of side length $h$ in each latent space $[-1,1]^r$; we use the same grid for the row and column factor spaces. 
Let $Q_a^{-}$ be the central cube of side length $h/2$ inside $Q_a$, and define
\begin{align}
    \phi_a(x)\defeq\min\left\{1,\frac4h\operatorname{dist}_\infty(x,Q_a^c)\right\} \qtext{where} \operatorname{dist}_\infty(x,Q_a^c)\defeq\inf_{y\in Q_a^c}\|x-y\|_\infty.
    \label{eq:full-minimax-bump}
\end{align}
The bump $\phi_a$ vanishes outside and the boundary of $Q_a$, takes value $1$ in $Q_a^{-}$, and is $4/h$-Lipschitz with respect to the Euclidean norm. Write
\begin{align}
    s \defeq \Prob(u_1 \in Q_a) = (h/2)^r \qtext {and} \Prob(u_1 \in Q_a^-) = 2^{-r}s.
    \label{eq:full-minimax-cube-mass}
\end{align}
There are $J=M^2$ cube pairs, each containing the target factors with probability $s^2=M^{-2}$. 

Let the array $\omega\defeq(\omega_{ab})_{a,b=1}^M$, where coordinates $\omega_{ab}\in\{0,1\}$ switch the bump off or on, respectively for the cube $Q_a \times Q_b$. Using the arrays, we define a finite subclass of functions,
\begin{align}
    f_\omega(u,v)
    \defeq\delta\sum_{a,b=1}^M\omega_{ab}\phi_a(u)\phi_b(v) \qtext{where} \omega\in\{0,1\}^{J},\qtext{and} \delta=\frac{Lh}{8}.
    \label{eq:full-minimax-hypercube}
\end{align}
Note the inequality 
\begin{align}
    0\leq f_\omega(u,v)\leq\delta=\frac{Lh}{8}\stackrel{\cref{eq:full-minimax-grid-scale}}{\leq} \frac{Lh_\star}{8} \stackrel{\cref{eq:full-minimax-regime}}{\leq} \frac1{32},
\end{align}
so that $f_\omega \in [-1/4,1/4]$; note that our lower bound construction uses only the nonnegative members of this larger class. 

The next lemma (proven in \cref{subsec:full-minimax-inclusions-proof}) establishes the two inclusions required to reduce our oracle risk to the risk for the finite subclass of functions $f_\omega$.

\begingroup
\renewcommand{\theHmylemma}{\theHsection.\arabic{mylemma}}
\begin{lemma}[\tbf{Inclusion of the constructed models}]
\label{lem:full-minimax-inclusions}
\leavevmode
\begin{enumerate}[label=(\alph*), leftmargin=*]
    \item For every $f\in\mc F_{r,L}$, the model $\mu^f$ in
    \cref{eq:full-minimax-bernoulli-submodel} belongs to $\mc M_{r,L}$.

    \item Suppose $h_\star\leq 1$ and $Lh_\star\leq 1/4$.
    For every $\omega\in\{0,1\}^J$, the function $f_\omega$ in
    \cref{eq:full-minimax-hypercube} belongs to $\mc F_{r,L}$.
\end{enumerate}
\end{lemma}
\endgroup
Set $q_\omega\defeq q_{f_\omega}=1/2+f_\omega$, $\Prob_\omega\defeq\Prob_{f_\omega}$, and $\E_\omega\defeq\E_{f_\omega}$. Thus $\Prob_\omega$ is the joint law induced by $\mu^{f_\omega}$, and $\E_\omega$ is its expectation. Consequently, for every oracle estimator, \cref{eq:full-minimax-submodel-comparison} extends to
\begin{align}
    \sup_{\mu\in\mc M_{r,L}}
    \E_\mu\brackets{\|\what\mu_{1,1}-\mu_{1,1}\|_\kernel^2}\geq\sup_{\omega\in\{0,1\}^J}
    \E_\omega\brackets{\|\what\mu_{1,1}-\mu_{1,1}^{f_\omega}\|_\kernel^2}.
    \label{eq:full-minimax-finite-comparison}
\end{align}
It therefore suffices to lower bound risk over this finite family. 

We next compare models differing in one bump.

\paragraph{Column isolation and conditional risk lower bounds via testing}
Fix and array $\omega$; then let $\omega^{(ab)}$ be the counterpart array obtained by replacing $\omega_{ab}$ with $1-\omega_{ab}$ and leaving all other entries unchanged. This flip changes the Bernoulli probability only on $Q_a\times Q_b$:
\begin{align}
    q_{\omega^{(ab)}}(u,v)-q_\omega(u,v)
    =\delta(1-2\omega_{ab})\phi_a(u)\phi_b(v).
\end{align}

For each cube pair $(a,b)$, we identify an event on which flipping $\omega_{ab}$ changes the probability of observing $x_1$ at the target entry by $\delta$. Conditional on this event, the \textit{oracle data laws} under the two models (by flipping) have total variation distance at most $1/4$. The union of these events has probability bounded below by a positive constant, depending only on $r$ and $H$.

For fixed pair $(a, b)$, define such event:
\begin{align}
    \mc E_{ab}\defeq\{u_1\in Q_a^{-},\ v_1\in Q_b^{-},\ v_t\notin Q_b\text{ for all }t\geq2\}.
    \label{eq:full-minimax-isolation-event}
\end{align}
Let $\Prob_\omega^{ab}\defeq\Prob_\omega(\,\cdot\mid\mc E_{ab})$ denote the law of the entire oracle data $\mc D_{\mrm{or}}$ under model $\omega$, conditional on $\mc E_{ab}$,
so the target Bernoulli parameter on the event  is $q_\omega(u_1,v_1)=\frac12+\delta\omega_{ab}$.
Thus on $\mc E_{ab}$, flipping $\omega_{ab}$ only changes the target probability by $\delta$ in absolute value. For every $t\geq2$, the same event gives $v_t\notin Q_b$, so $\phi_b(v_t)=0$ and the two models agree throughout column $t$. 

We leave the donor rows unrestricted: the calculation in \cref{subsec:full-minimax-isolation-testing-proof} shows that their measurements in column $1$ provide too little information to distinguish the paired models reliably, with the $npN$ scale arising from these measurements.
For probability laws $\Prob,\mbb Q$, write $\mrm{TV}(\Prob,\mbb Q)\defeq\sup_B|\Prob(B)-\mbb Q(B)|$, with the supremum over measurable events.

The next lemma (proven in \cref{subsec:full-minimax-isolation-testing-proof}) bounds the total isolation probability and the conditional total variation.

\begingroup
\renewcommand{\theHmylemma}{\theHsection.\arabic{mylemma}}
\begin{lemma}[\tbf{Column isolation and total variation}]
\label{lem:full-minimax-isolation-testing}
Under \cref{eq:full-minimax-regime},
\begin{enumerate}[label=(\alph*), leftmargin=*]
\item The events $\mc E_{ab}$ have a common probability $\Prob(\mc E_{ab}) = (2^{-r}s)^2 (1 - s)^{T-1} \defeq \pi_E$ for every $a, b \in [M]$ and $\omega\in \{0, 1\}^J$. Hence, the total probabiliy is lower bounded by
\begin{align}
    \sum_{a,b}\Prob_\omega(\mc E_{ab})= M^2\pi_E
    \geq2^{-2r}\exp(-2^{1-r}H). \notag
\end{align}
\item For every $\omega\in\{0,1\}^J$ and $a,b\in[M]$,
\begin{align}
    \mrm{TV}(\Prob_\omega^{ab},\Prob_{\omega^{(ab)}}^{ab})\leq\frac{1}{4}. 
\end{align}
\end{enumerate}
\end{lemma}
\endgroup

\paragraph{From testing to estimation error}

Fix an arbitrary distribution-valued oracle estimator $\what\mu_{1,1}$, recall $\what q$ is its scalar projection used in \cref{eq:full-minimax-scalar-reduction}. 
Define the test
\begin{align}
    \what{\omega}_{ab}\defeq\indic{\what q\geq1/2+\delta/2};
\end{align}
so $\what \omega_{ab} \neq \omega_{ab}$ implies $|\what q - q_\omega(u_1, v_1)|\geq\delta/2$ and
\begin{align}
    (\what q-q_\omega(u_1,v_1))^2
    \geq\frac{\delta^2}{4}\indic{\what{\omega}_{ab}\neq\omega_{ab}}
    \quad\text{on} \quad\mc E_{ab}.
    \label{eq:full-minimax-decoding}
\end{align}
For each pair $(a, b)$, take $\omega$ with $\omega_{ab}=0$, so $\omega^{(ab)}_{ab}=1$. Then, conditioned on the event $\mc E_{ab}$,
\begin{align}
    \Prob_\omega^{ab}(\what{\omega}_{ab}=1)
    +\Prob_{\omega^{(ab)}}^{ab}(\what{\omega}_{ab}=0) &= 1-\big[\Prob_{\omega^{(ab)}}^{ab}(\what{\omega}_{ab}=1)
    -\Prob_\omega^{ab}(\what{\omega}_{ab}=1)\big] \notag\\
    &\geq1-\mrm{TV}(\Prob_\omega^{ab},\Prob_{\omega^{(ab)}}^{ab})
    \stackrel{\text{\cref{lem:full-minimax-isolation-testing} (b)}}{\geq}\frac{3}{4}.
\end{align}
For a fixed coordinate $(a,b)$, there are $2^{J-1}$ combinations of arrays where the inequality of the above display holds. Averaging their error sums therefore yields
\begin{align}
    2^{-J}\sum_{\omega\in\{0,1\}^{J}}
    \Prob_\omega^{ab}(\what{\omega}_{ab}\neq\omega_{ab})\geq\frac38.
    \label{eq:full-minimax-average-testing}
\end{align}
Note that $\mc E_{ab}$ are disjoint across $(a, b)$, so $1 \geq \sum_{a, b \in [M]}\indic{\mc E_{ab}}$. Collecting what we derived, we observe
\begin{align}
    \sup_{\omega \in \{0, 1\}^J} \E_\omega \big[ (\what q - q_\omega(u_1, v_1))^2 \big] &\geq \frac{1}{2^J} \sum_{\omega \in \{0, 1\}^J} \sum_{a, b \in [M]} \E_\omega \big[ (\what q - q_\omega(u_1, v_1))^2 \indic{\mc E_{ab}} \big]\\
    &\stackrel{\cref{eq:full-minimax-decoding}}{\geq} \frac{\delta^2}{4}\sum_{a, b \in [M]} \pi_E \bigg[ \frac{1}{2^J} \sum_{\omega \in \{0, 1\}^J} \Prob_\omega^{ab}(\what \omega_{ab} \neq \omega_{ab}) \bigg]\\
    &\stackrel{\cref{eq:full-minimax-average-testing}}{\geq} \frac{3\delta^2}{32}M^2\pi_E \stackrel{\text{\cref{lem:full-minimax-isolation-testing}} (a)}{\geq} \frac{3\delta^2}{32} 2^{-2r}\exp(-2^{1-r}H)
\end{align}
Take 
\begin{align}
    c(r,H)=\frac{\kappa_\kernel}{1536}\,2^{-2r}\exp\bigl(-2^{1-r}H\bigr).
    \label{eq:minimax-constant}
\end{align}
By \cref{eq:full-minimax-bandwidth,eq:full-minimax-grid-scale,eq:full-minimax-hypercube},
we have $\delta^2=L^2h^2/64\geq a_\star/144$. Combining this with the
preceding display and
\cref{eq:full-minimax-scalar-reduction,eq:full-minimax-finite-comparison}
shows that every oracle estimator has worst-case squared $\mmd$ risk at least
$c(r,H)a_\star$. Taking the infimum over oracle estimators and applying
\cref{eq:full-minimax-oracle-risk} proves \cref{eq:full-minimax-lower-bound}.

\subsection{Proof of \cref{lem:full-minimax-inclusions}}
\label{subsec:full-minimax-inclusions-proof}

We first establish $\mu^f\in\mc M_{r,L}$ for every $f\in\mc F_{r,L}$. Taking inner products with the fixed orthonormal basis elements $\psi_\ell$ and using the reproducing property gives
\begin{align}
    \alpha_\ell^f(u,v)
    &=\langle g_f(u,v),\psi_\ell\rangle_\kernel
    =\psi_\ell(x_0)+q_f(u,v)\langle d_\kernel,\psi_\ell\rangle_\kernel.
\end{align}
Since $q_f=1/2+f$, each coefficient $\alpha_\ell^f$ is $L|\langle d_\kernel,\psi_\ell\rangle_\kernel|$-Lipschitz. By Parseval's identity, these coefficient Lipschitz constants satisfy
\begin{align}
    \sum_{\ell\geq1}\big(L|\langle d_\kernel,\psi_\ell\rangle_\kernel|\big)^2
    =L^2\|d_\kernel\|_\kernel^2
    =L^2\kappa_\kernel\leq L^2.
    \label{eq:full-minimax-submodel-regularity}
\end{align}
Hence $\mu^f\in\mc M_{r,L}$ for every $f\in\mc F_{r,L}$.

For each $(u,v)$, the product $\phi_a(u)\phi_b(v)$ can be nonzero only if $u$ lies in the interior of $Q_a$ and $v$ lies in the interior of $Q_b$. Since the cube interiors are disjoint, at most one term in the sum defining $f_\omega(u,v)$ is nonzero, so $0\leq f_\omega(u,v)\leq\delta$.

Now suppose $(u,v)$ and $(u',v')$ both lie in $Q_a\times Q_b$. Adding and subtracting $\phi_a(u')\phi_b(v)$ gives $\phi_a(u)\phi_b(v)-\phi_a(u')\phi_b(v')= \bigl(\phi_a(u)-\phi_a(u')\bigr)\phi_b(v) + \phi_a(u')\bigl(\phi_b(v)-\phi_b(v')\bigr)$. Then,
\begin{align}
    |f_\omega(u,v)-f_\omega(u',v')|
    &\leq\frac{4\delta}{h}\big(\|u-u'\|_2+\|v-v'\|_2\big) \notag\\
    &\leq\frac{8\delta}{h}\big(\|u-u'\|_2\vee\|v-v'\|_2\big).
\end{align}
For points in different product cubes, split the segment joining them at the cube boundaries, where $f_\omega$ vanishes. Summing the preceding bound over the subsegments gives the same bound for the endpoints, since the subsegment lengths sum to $\|u-u'\|_2\vee\|v-v'\|_2$. Thus $f_\omega$ is $L$-Lipschitz because $8\delta/h=L$. Since $0\leq f_\omega\leq\delta\leq1/32$ by \cref{eq:full-minimax-regime,eq:full-minimax-grid-scale},
we conclude that $f_\omega\in\mc F_{r,L}$.

\subsection{Proof of \cref{lem:full-minimax-isolation-testing}}
\label{subsec:full-minimax-isolation-testing-proof}

Independence of the factors yields
\begin{align}
    \pi_E=\Prob_\omega(\mc E_{ab})
    &=(2^{-r}s)^2(1-s)^{T-1}.
\end{align}
Since $Ms=1$, summing over the $M^2$ cube pairs gives
\begin{align}
    \sum_{a,b}\Prob_\omega(\mc E_{ab}) = M^2\pi_E = 2^{-2r}(1-s)^{T-1} \geq 2^{-2r}\exp(-2Ts)
     \geq 2^{-2r}\exp(-2^{1-r}H).
\end{align}
The first inequality uses $\log(1-s)\geq-2s$, since
$s=(h/2)^r\leq1/2$. The second uses
$Ts=2^{-r}Th^r\leq2^{-r}Th_\star^r\leq2^{-r}H$
by \cref{eq:full-minimax-grid-scale,eq:full-minimax-regime}.

Conditional on $\mc E_{ab}$, the law of $(\mc U,\mc V,\mc A)$ is the same under $\omega$ and $\omega^{(ab)}$. Given these variables, the observed measurements are independent. We introduce the shorthand
\begin{align}
    q_i = q_\omega(u_i, v_1) \qtext{and} q_i' = q_{\omega^{(ab)}}(u_i, v_1).
\end{align}
Recall
\begin{align}
    q_{\omega^{(ab)}}(u_i, v_t) - q_\omega (u_i, v_t) = \delta (1 - 2 \omega_{ab}) \phi_a(u_i)\phi_b(v_t).
\end{align}
So on $\mc E_{ab}$, $q_i' - q_i = \delta(1 - 2\omega_{ab})\phi_a(u_i)$ and zero whenever $t\geq 2$. Since entry $(1, 1)$ is masked, the only observed entries in entry $(i, 1)$ with $u_i \in Q_a$ can provide information about whether $\omega_{ab}$ is zero or one.

Again, conditioned on $\mc E_{ab}$, applying the KL formula for Bernoulli distributions and the inequality $\log x\leq x-1$, we observe
\begin{align}
    \mrm{KL}(\Bern(q_i)\|\Bern(q_i')) \leq \frac{(q_i-q_i')^2}{q_i'(1-q_i')} \stackrel{\text{\cref{lem:full-minimax-inclusions} (b)}}{\leq} \frac{16}{3}\delta^2 \phi_a(u_i)^2
    \label{eq:kl-indi}
\end{align}

Note that $\Prob_{\omega}^{ab}$ and $\Prob_{\omega^{(ab)}}^{ab}$ are the conditional distribution of the entire randomness of the two models characterized; this implies that conditioned on $(\mc U, \mc V, \mc A)$, $\Prob_{\omega}^{ab}$ and $\Prob_{\omega^{(ab)}}^{ab}$ are the randomness of the measurements which we assume to be conditionally independent. On $\mc E_{ab}$, the measurement distributions are identical for all entries with $t\geq 2$. So 
\begin{align}
    \mrm{KL}\big( \Prob_{\omega}^{ab} \| \Prob_{\omega^{(ab)}}^{ab} \big) &= \E_\omega \bigg[\E_\omega \bigg[ \log \bigg( \frac{d\Prob_\omega^{ab}}{d\Prob_{\omega^{(ab)}}^{ab}} \bigg) \Bigm| \mc U, \mc V, \mc A\bigg] \Bigm| \mc E_{ab} \bigg]\\
    &= \E_\omega \bigg[ n \sum_{i = 2}^{N}A_{i, 1} \mrm{KL} (\mrm{Bern}(q_i) \| \mrm{Bern}(q_i')) \Bigm| \mc E_{ab}\bigg]\\
    &\stackrel{\cref{eq:kl-indi}}{\leq} \frac{16}{3}n\delta^2\sum_{i=2}^N\E_\omega \big[ A_{i, 1}\phi_a(u_i)^2 |\mc E_{ab} \big]\\
    &\leq \frac{16}{3}n\delta^2p(N-1)(h/2)^r 
\end{align}
where the last inequality follows from $\E_{\omega}[ A_{i, 1}\phi_a(u_i)^2  |\mc E_{ab} ] \leq \Prob_\omega(A_{i, 1}=1, u_i \in Q_a|\mc E_{ab})=ps$. Lastly, invoke $h \leq h_\star$ and $npNL^2h_\star^{r+2}=1$ and we observe $\mrm{KL}\big( \Prob_{\omega}^{ab} \| \Prob_{\omega^{(ab)}}^{ab} \big) \leq 1/8$. Pinsker's inequality now gives
\begin{align}
    \mrm{TV}(\Prob_\omega^{ab},\Prob_{\omega^{(ab)}}^{ab})
    \leq \sqrt{\tfrac12
    \mrm{KL}(\Prob_\omega^{ab}\|\Prob_{\omega^{(ab)}}^{ab})}
    \leq \tfrac14.
    \label{eq:full-minimax-conditional-tv}
\end{align}

\subsection{Proof of \cref{thm:full-matrix-minimax-optimality}}
\label{subsec:full-minimax-optimality-proof}

By \cref{eq:full-minimax-joint-sample-conditions}, $h_\star\leq1$ and $Lh_\star=\sqrt{a_\star}\leq1/4$. The second inequality of \cref{eq:full-minimax-joint-window} gives $T\leq npNa_\star=h_\star^{-r}$. Thus \cref{thm:full-matrix-oracle-minimax} applies with $H=1$ and yields $\mc R^\star\geq c(r,1)a_\star$.

Now we prove the upper bound of the minimax risk. Fix any model $\mu \in \mc M_{r, L}$ and fix our radius $\eta^\star$ as in \cref{eq:full-minimax-common-radius}. Use \cref{eq:twoevent_conc} and \cref{lem:conditional-concentration} with $\delta = N^{-3}$, then we have $|\rho_{j, 1} - \Delta_{j, 1}|\leq 2\vareps_N$ for all $j\neq1$ with $A_{j,1}=1$ with probability at least $1 - (N-1)\exp{-p^2(T-1)/8} - N^{-2}$. Next apply the argument of \cref{lem:exam_lip} with $\eta' = a_\star$ to observe $|\{ j \neq 1: A_{j, 1}=1, \gio \leq a_\star \}| \geq b_rp(N-1)h_\star^r$ with probability at least $1 - \exp\{-b_rp(N-1)h_\star^r\}$, where $b_r = c_r/8$. On these high probability events, we observe
\begin{align}
    \sum_{j \in \mbf N_{1, \eta^\star}}A_{j, 1} \geq b_r p(N-1)h_\star^r \qtext{and} \gio \leq \eta^\star + 2\vareps_N.
\end{align}
The inequalities in the above display bound the bias and variance term of the mean squared $\mmd$ error by $(a_\star+ \vareps_N)$ up to logarithmic term. Lastly, we observe that \cref{eq:full-minimax-joint-window} and \cref{eq:full-minimax-joint-sample-conditions} is sufficient for $\vareps_N\leq a_\star$ and for the low-probability term of the two events to be bounded by a constant multiple of $a_\star$.

The upper bound argument holds for any $\mu\in \mc M_{r, L}$ for a fixed $\eta^\star$; so collecting what we have derived, there exists a constant $C_r$ depending on $r$ such that
\begin{align}
    c(r, 1)a_\star \leq \mc R^\star \leq \mc R_{\kerNN} \leq C_r a_\star \log n.
\end{align}

\section{Experimental implementation and additional results}
\label{app:sim}

This appendix gives the implementation details for \cref{sec:sim}. 

\subsection{Cross validation and evaluation}
\label{app:experiment-protocol}

Let $E_H=\{\eta_1,\ldots,\eta_H\}$ be a finite set of candidate radii. We use the first $T/2$ columns to compute row distances and the last $T/2$ columns for validation.

The row distance computed from the first $T/2$ columns is
\begin{align}
    \rho^{\rm cv}_{i,j}
    =\frac{\sum_{s=1}^{T/2}A_{i,s}A_{j,s}
    \widehat{\mmd}_{\kernel}^{\,2}(\empdistZ{i,s},\empdistZ{j,s})}
    {\sum_{s=1}^{T/2}A_{i,s}A_{j,s}}.
\end{align}
For each candidate $\eta$, form the neighborhood
$\mbf N^{\rm cv}_{i,\eta}=\{j\ne i:\rho^{\rm cv}_{i,j}\leq\eta\}$. For each observed cell $(i,t)$ with $T/2<t\leq T$, the prediction is
\begin{align}
    \widehat\mu^{\rm cv}_{i,t,\eta}
    =\frac{\sum_{j\in\mbf N^{\rm cv}_{i,\eta}}A_{j,t}\empdistZ{j,t}}
    {\sum_{j\in\mbf N^{\rm cv}_{i,\eta}}A_{j,t}}.
\end{align}
We choose the radius with the smallest average validation loss,
\begin{align}
    \widehat\sigma_\eta
    &=\frac{\sum_{t=T/2+1}^{T}\sum_{i:\,A_{i,t}=1}
      \mmd_{\kernel}^2(\widehat\mu^{\rm cv}_{i,t,\eta},\empdistZ{i,t})}
      {\sum_{t=T/2+1}^{T}\sum_{i=1}^{N} A_{i,t}},
    &\etacv&\in\argmin_{\eta\in E_H}\widehat\sigma_\eta.
    \label{eq:obj-ftn-cv}
\end{align}

\paragraph{Evaluation and uncertainty}
\label{app : choice of distance in simulation}
In the simulations, the measurements in each cell follow Gaussian distribution $\mu_{i,t}$. We compute $\mmd_{\kernel}^2(\widehat\mu_{i,t,\eta},\mu_{i,t})$ analytically. A large independent sample from the target can also be used to approximate this loss. For mean estimation, we report the squared Euclidean distance between the estimated and true means.

We compare methods on the same targets and report medians unless stated otherwise. Confidence intervals are obtained by resampling independent matrices with all their targets.

\subsection{Synthetic designs}
\label{app:synthetic-designs}

Unless stated otherwise, the simulated panels have $N\in\{16,32,\ldots,512\}$ rows and $T=80$ columns, and each observed cell contains $n=30$ independent measurements from its cell distribution $\mu_{i,t}$. Each row has a two-dimensional latent factor $u_i=(u_i(1),u_i(2))\sim U([-1,1]\times[0.2,1])$, drawn independently across rows. Unless stated otherwise, each column has a factor $v_t=(v_t(1),v_t(2))\sim U([0.2,1]\times[0.5,2])$, drawn independently across columns and of the row factors.

\label{app:synthetic-missingness}
We impose two missingness mechanisms on the panel data. Under MCAR, each cell is observed independently with probability $0.5$. Under confounded staggered adoption, the first $N/4$ rows form group~1, the next $N/2$ rows form group~2, and the last $N/4$ rows never adopt. In each period $t\geq T^{\beta_g}$, a unit $i$ of group $g \in \{1, 2\}$ that has not yet adopted adopts with probability
\begin{align}
    p_i=\operatorname{expit}\bigl(\gamma_{g,0}+\gamma_{g,1}u_{i-1}(1)+\gamma_{g,2}u_i(1)+\gamma_{g,3}u_{i+1}(1)\bigr),
\end{align}
with row indices taken cyclically~($u_0=u_N$ and $u_{N+1}=u_1$), $(\beta_1,\beta_2)=(0.5,0.75)$, $\gamma_1=(0,0.5,1,0.5)$, and $\gamma_2=(-2,0.5,1,0.5)$. Writing $\tau_i$ for the adoption time of unit $i$, its control measurements are observed for $t\leq\tau_i$ and missing afterwards, as in \cref{mod:potential-outcome staggered}. The three studies below share the missingness mechanisms and they differ in how the cell distributions are generated.

\paragraph{Alignment of theory and empirical results}
In \cref{fig:sim_mmdbyN}, the cell distribution is $\mu_{i,t}=N(m_{i,t},\Sigma_{i,t})$, where
\begin{align}
    m_{i,t}=u_i(1)v_t(1)\,(-1,1,\ldots,-1,1)\tp
    \qtext{and}
    \Sigma_{i,t}=u_i(2)v_t(2)\operatorname{diag}(1,\tfrac12,\ldots,1,\tfrac12).
    \label{eq:alignment-setting}
\end{align}
We take $d\in\{2,4,6\}$, use both missingness mechanisms, and generate 50 independent matrices for each row size $N$. The target is the single cell $(1,T)$.

\paragraph{Mean estimation}
\label{app:mean-imputation}
\cref{fig:feature-kernel-choice} fixes dimension $d=4$ and replaces the mean and covariance by
\begin{align}
    m_{i,t}=u_i(1)v_t(1)s_t\,(-1,1,-1,1)\tp
    \qtext{and}
    \Sigma_{i,t}=0.75\exp\{u_i(1)z_t\}\operatorname{diag}(1,\tfrac12,1,\tfrac12),
\end{align}
where $z_t\sim U[0.5,1.5]$ is independent of the other factors. The mean and the variance of row $i$ are thus driven by the same factor $u_i(1)$. Panels (a) and (b) of \cref{fig:feature-kernel-choice} take $s_t=1$, and panel (c) takes $s_t=(t/T)^2$ under staggered adoption. 

\kerNN uses the exponential kernel $\kernel(x,y)=\exp\{-\|x-y\|_2^2/(2d)\}$. The radii of \kerNN and Scalar-NN are chosen by cross validation on the target row. SoftImpute is applied to the $N\times Td$ matrix of observed sample means after centering each column, with its penalty chosen on a random $10\%$ holdout of the observed entries.

\paragraph{Dependent columns}
\label{app:dependent-columns}
\cref{fig:dependent-n-sweep} uses the mean and covariance of \cref{eq:alignment-setting} with $d=4$. The $\iid$ panel keeps the column factors as specified above, but in the AR$(k)$ panels, $k\in\{1,2\}$, the column factors are instead generated as
\begin{align}
    v_t(1)&=0.2+0.4\{1+\tanh w_t(1)\},
    \qquad
    v_t(2)=0.5+0.75\{1+\tanh w_t(2)\},\\
    w_t(l)&=\frac{\rho}{k}\sum_{s=1}^{k}w_{t-s}(l)+\sigma_k\epsilon_t(l),
    \qquad l\in\{1,2\},
\end{align}
with $\rho=0.85$, independent $\epsilon_t(l)\sim N(0,1)$, and $\sigma_k$ chosen so that each $w_t(l)$ is stationary with unit variance. This keeps the ranges of the column factors $v_t(1) \in [0.2, 1]$ and $v_t(2)\in [0.5,2]$ but makes them dependent across columns. Every panel uses both missingness mechanisms.

\begin{figure}[!t]
    \centering
    \includegraphics[width=0.95\linewidth]{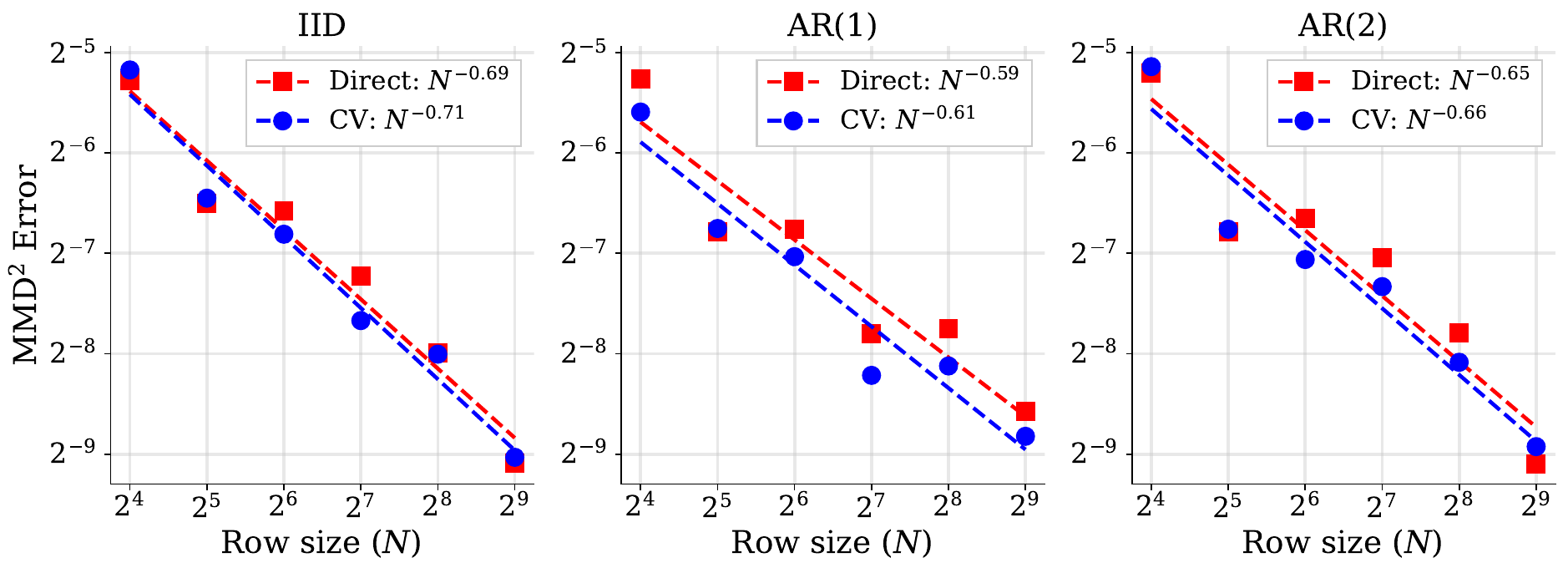}
    \caption{\tbf{CV and Direct under staggered adoption with independent and dependent columns.} Each panel reports the median squared-$\mmd$ error over paired targets as $N$ increases, for $\iid$ column factors or AR$(k)$ column factors with $k=1,2$ and autoregressive parameter $0.85$. The targets are as in \cref{fig:staggered-cv-direct}, and Direct is implemented as in \cref{eq:non-cv-opt}. The IID panel repeats the error panel of \cref{fig:staggered-cv-direct}, and the CV curves coincide with the MNAR curves of \cref{fig:dependent-n-sweep}.}
    \label{fig:dependent-cv-direct}
\end{figure}

\subsection{Computational alternatives for radius selection}
\label{app:computational-alternatives}

In \cref{subsec:selection-computation}, we compare CV with two efficient ways of choosing the radius: Direct, which minimizes \cref{eq:non-cv-opt}, and Entry-KT, which runs CV on the thinned measurments of each cells. Both studies use the model of the alignment study in \cref{app:synthetic-designs} with independent columns, $d=4$, and the exponential kernel of \cref{app:synthetic-designs}. CV is the baseline here and is run separately for each target entry. 

\paragraph{Direct selection}
\label{app:direct-dependence}
\cref{fig:staggered-cv-direct} uses staggered adoption as its missingness pattern.
Direct minimizes the objective in \cref{eq:non-cv-opt}, with $\delta=1/2$ and $\|\kernel\|_\infty=1$. The time reported for each target covers the row distances, the radius selection, and the final estimate. \cref{fig:dependent-cv-direct} repeats this comparison with the AR$(k)$ column factors of \cref{app:synthetic-designs}. Direct attains the same error decay as CV under dependent columns as well.

\paragraph{Entry-wise kernel thinning}
\label{app:entry-thinning}
\cref{fig:entrykt-comparison} uses MCAR with $N\in\{32,256\}$, $n=256$.
Standard thinning keeps a uniformly random subset of $m$ measurements in each observed cell. Entry-KT applies KT-SPLIT and KT-SWAP~\cite{dwivedi2021kernel} with the same kernel, $\delta=1/2$, and $\log_2(n/m)$ halving rounds. We take $m\in\{8,16,32,64,128\}$. 

Let $S\subset[n]$ with $|S|=m$ index the measurements kept in cell $(i,s)$. A kernel thinned cell stands in for all $n$ original measurements, so the term $\frac{1}{n(n-1)}\sum_{\l\neq\l'}\kernel(X_{\l}(i,s),X_{\l'}(i,s))$ in \cref{def : ustat_mmd} is replaced by
\begin{align}
    \frac{1}{n-1}\biggl\{\frac{n}{m^2}\sum_{\l,\l'\in S}\kernel\bigl(X_{\l}(i,s),X_{\l'}(i,s)\bigr)-\frac1m\sum_{\l\in S}\kernel\bigl(X_{\l}(i,s),X_{\l}(i,s)\bigr)\biggr\},
\end{align}
whereas standard thinning uses the usual $m$-sample term. Both methods form the final estimate from all original measurements of the selected neighbors.

\begin{figure}[!htb]
    \centering
    \includegraphics[width=0.95\linewidth]{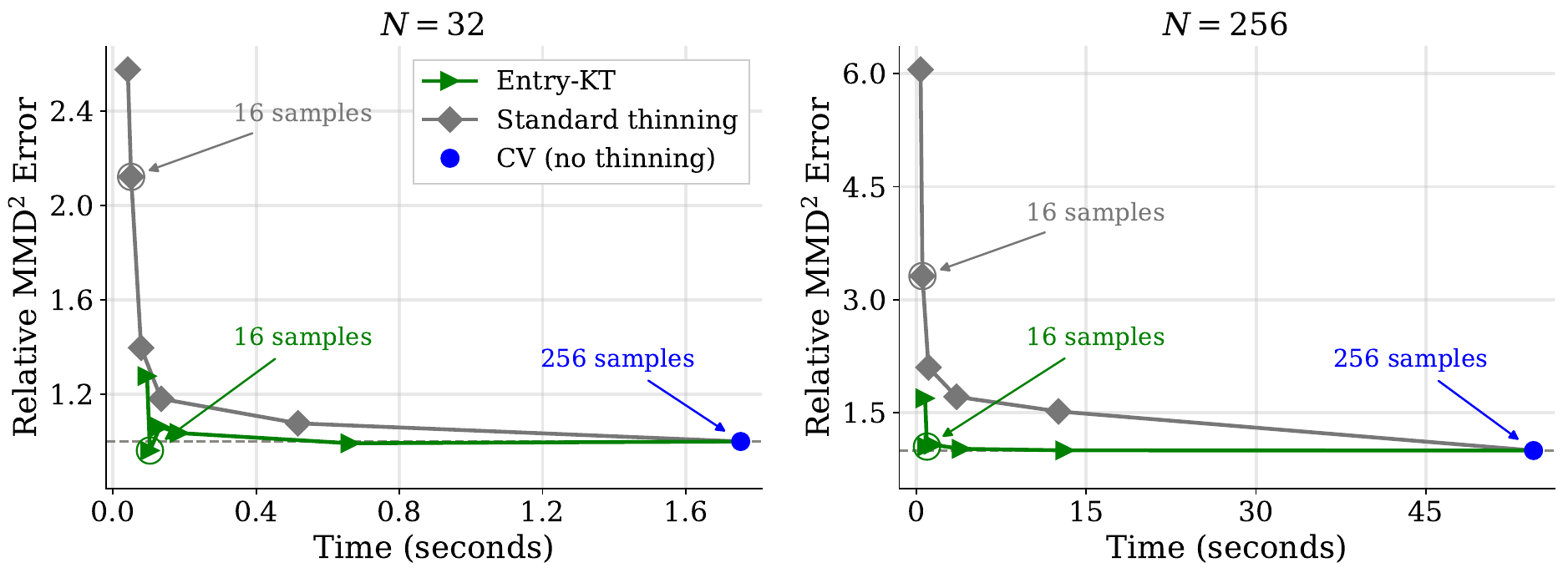}
    \caption{\tbf{Runtime--accuracy tradeoff across retained sample sizes.} Mean per-target squared-$\mmd$ error relative to CV versus median per-target runtime, using the same experiments as \cref{fig:entrykt-comparison}. From left to right, the points on each thinning curve retain $8,16,32,64,128$ measurements per cell; the shared blue endpoint uses all $256$. Annotations highlight the two 16 measurement configurations.}
    \label{fig:entrykt-pareto}
\end{figure}

\cref{fig:entrykt-pareto} shows the tradeoff over all retained sizes. Entry-KT has a lower error than standard thinning at every $m$ for both row sizes.

\subsection{HeartSteps implementation}
\label{app : cv_heartsteps}

At each decision point, the study recorded whether the participant was {available}, meaning that the participant was not driving, not walking or running, and the phone was online~\cite{klasnja2019efficacy}. An entry $(i,t)$ is observed if participant $i$ was available at decision point $t$, a notification was sent, and the study data report a complete step count information the following hour; this hour is split into twelve five-minute counts, which are the measurements of the entry. All other entries are regarded as missing. This gives 3,129 observed entries, shown in dark blue in \cref{fig:heart_missing}. For evaluation, we divide the 200 decision points into five blocks of 40 consecutive decision points~(eight days) and select one block per participant. The observed entries in the selected blocks are the 623 evaluated entries.

Each evaluated entry is fitted in four different ways: with the exponential or the square kernel, and with the CV or the Direct radius. \cref{fig:direct_cv_hs} reports all four fits, and \cref{fig:dist_learning_hs} takes the exponential kernel and the CV radius. Both kernels are applied to the log-transformed step counts of \cref{sec:application}. For a target $(i,t)$, CV follows \cref{app:experiment-protocol} on participant $i$'s observed decision points other than $t$. Direct minimizes \cref{eq:non-cv-opt} with $\delta=1/2$. Since the square kernel is unbounded, $\|\kernel\|_\infty$ is replaced by the largest observed value of $\kernel(x,x)$.

\begin{table}[!b]
    \centering
    \caption{\tbf{Predicting the mean and standard deviation in HeartSteps.} Mean absolute error over the evaluated entries, for the mean and the standard deviation of the twelve log-transformed step counts of an entry.}
    \label{tab:scalar_imp_hs}
    \begin{tabular}{lcc}
        \toprule
        Method & Mean & Standard deviation \\
        \midrule
        $\kerNN$ & 1.065 & 0.734 \\
        Scalar-NN & 1.060 & 0.694 \\
        SoftImpute & 1.045 & 0.658 \\
        USVT & 1.089 & 0.677 \\
        \bottomrule
    \end{tabular}
\end{table}

\paragraph{Step-count summaries}
\cref{fig:dist_learning_hs} considers nine summaries of the twelve step counts of an entry: the mean of $\log(1+\text{steps})$, the proportions of counts equal to zero, above 10, and above 100, and the 10th, 25th, 50th, 75th, and 90th percentiles of $\log(1+\text{steps})$. $\kerNN$ computes every summary from the pooled measurements of the identical neighbors, while Scalar-NN~\cite{li2019nearest} runs nearest neighbors on each summary separately, with its own CV radius. The first four summaries are scored by squared error and the percentiles by pinball loss\footnote{For a level $\alpha\in(0,1)$, the pinball loss of a predicted quantile $q$ at an observation $x$ is $\alpha(x-q)$ if $x\geq q$ and $(1-\alpha)(q-x)$ otherwise. Its expected value over $x$ is smallest when $q$ is the $\alpha$-quantile of the distribution of $x$.}. Each point is the average loss of $\kerNN$ as a percentage of that of Scalar-NN, and the bars are 95\% intervals from resampling participants.

\paragraph{Comparison with matrix completion}
\cref{tab:scalar_imp_hs} compares $\kerNN$ with the square kernel against Scalar-NN, SoftImpute~\cite{mazumder2010softimpute}, and USVT~\cite{chatterjee2015matrix} when predicting the mean and standard deviation of $\log(1+\text{steps})$ on the same 623 entries. The errors for the mean are similar across methods, while $\kerNN$ has the largest error for the standard deviation.

\end{appendix}

\end{document}